\documentclass{article}

\usepackage[preprint]{neurips_2026}

\usepackage[utf8]{inputenc} 
\usepackage[T1]{fontenc}    
\usepackage{hyperref}       
\usepackage{url}            
\usepackage{booktabs}       
\usepackage{amsfonts}       
\usepackage{nicefrac}       
\usepackage{microtype}      
\usepackage{xcolor}         

\usepackage{amsmath}
\usepackage{amssymb}
\usepackage{mathtools}
\usepackage{amsthm}
\usepackage{tikz}
\usetikzlibrary{positioning,arrows.meta,shapes.multipart}
\usepackage{booktabs}
\usepackage{multirow}
\usepackage{subcaption}
\usepackage{algorithm}   
\usepackage{algorithmic} 
\usepackage{subcaption}
\usepackage{cleveref}

\title{QuantKAN: A Unified Quantization Framework \\
for Kolmogorov–Arnold Networks}

\author{%
  Kazi Ahmed Asif Fuad \\
  Department of EECS\\
  Oregon State University\\
  Corvallis, OR 97331 \\
  \texttt{fuadk@oregonstate.edu} \\
  \And
  Lizhong Chen \\
  Department of EECS\\
  Oregon State University\\
  Corvallis, OR 97331 \\
  \texttt{chenliz@oregonstate.edu} \\
}

\begin{document}

\maketitle

\begin{abstract}
  Kolmogorov--Arnold Networks (KANs) replace linear weights with spline-based functions, offering strong expressivity but posing challenges for low-precision deployment due to heterogeneous parameter distributions. We introduce QuantKAN, the first unified framework for quantization-aware training (QAT) and post-training quantization (PTQ) of KANs. The framework employs branch-aware quantizers for base and spline parameters and extends modern QAT and PTQ methods to spline-based layers across EfficientKAN, FastKAN, PyKAN, and KAGN. Experiments on MNIST, CIFAR-10/100, TinyImageNet, and ImageNet provide the first unified QAT/PTQ KAN benchmarks and show that DSQ is the most robust QAT method at aggressive low-bit settings, while GPTQ is the strongest PTQ method at moderate precision. Sensitivity analyses reveal architecture-specific failure modes: spline/basis parameters dominate in FastKAN, while base or scaling parameters dominate in EfficientKAN, GRAM, and PyKAN. Vivado HLS estimates on a Xilinx UltraScale+ device further suggest up to 3.32$\times$ throughput and 7.7$\times$ lower estimated dynamic energy per inference under W4A4, exposing a residual \emph{basis-evaluation tax} that motivates basis-aware microarchitecture.  QuantKAN is available at \url{https://github.com/OSU-STARLAB/QuantKAN/}.
\end{abstract}


\section{Introduction}

Kolmogorov--Arnold Networks (KANs) replace scalar weights with learnable spline- or basis-based functions, offering an interpretable and flexible alternative to conventional MLP and CNN layers \citep{liu2024kan}. Variants of PyKAN such as EfficientKAN, FastKAN, and KAGN demonstrate that these function-based parameterizations can be effective across scientific learning, classification, and convolutional settings \citep{liu2024kan,blealtan_efficientkan_github,li2024kolmogorovarnold,drokin2024kolmogorov,khochawongwat2024gramkan}. However, the same basis expansions that improve expressivity also complicate efficient deployment, especially in low-precision settings.

Quantization is a standard tool for reducing memory and compute cost in modern neural networks, and has been extensively developed for CNNs and Transformers \citep{jacob2018quantization,nagel2020adaround,li2021brecq,frantar2022gptq,lin2023awq,xiao2023smoothquant}. Yet these methods are largely designed for architectures with relatively homogeneous parameter statistics and conventional activation placement. KANs violate these assumptions: their base and spline/polynomial branches play different functional roles and often exhibit different scales, distributions, and sensitivities. As a result, directly applying existing quantizers to KANs is often unreliable, and quantization design for spline-based architectures remains comparatively underexplored.

To address this gap, we introduce \textbf{QuantKAN}, a unified framework for quantizing KANs. QuantKAN supports both quantization-aware training (QAT) and post-training quantization (PTQ), extending representative quantization methods to spline- and polynomial-based layers through branch-aware treatment of base weights, spline/basis coefficients, and activations. Across EfficientKAN, FastKAN, PyKAN (official KAN variant), and KAGN, and across MNIST, CIFAR-10/100, TinyImageNet, and ImageNet, the proposed QuantKAN establishes a unified QAT/PTQ low-bit benchmark and reveals architecture- and parameter-specific quantization behavior in KANs.


Empirically, DSQ is the most robust QAT method, retaining $97.7\%$ on MNIST at W2A2 and $55.2\%$ on ImageNet at W4A4, while GPTQ is the strongest PTQ method overall at moderate precision. Sensitivity analyses show that failure modes are branch- and architecture-specific:
spline/basis components dominate in FastKAN, while base or scaling parameters dominate in EfficientKAN, GRAM, and PyKAN, and branch-aware quantization yields up to $+23.86$ points over a shared quantizer on PyKAN at W2. In Vivado HLS synthesis estimates, W4A4 designs reach up to 3.32$\times$ throughput and 7.7$\times$ lower estimated dynamic energy per inference under a bitwidth-aware activity model, exposing a residual \emph{basis-evaluation tax} that motivates basis-aware microarchitecture. Our contributions are summarized as follows:
\begin{itemize}
    \item \textit{Unified QuantKAN framework:} We introduce a unified framework for QAT and PTQ in KANs, extending representative quantization methods to spline- and polynomial-based layers in an implementation-aligned way.
    \item \textit{Branch-aware quantization:} Modular wrappers for EfficientKAN, FastKAN, PyKAN, and KAGN independently quantize base weights, spline/basis parameters, and activations. 
    \item \textit{Systematic benchmarks, insights, and hardware efficiency:} We establish low-bit benchmarks across multiple KAN families and datasets, provide ablations on method--architecture interactions, parameter sensitivity, and mixed-precision design, and quantify deployment efficiency through FPGA latency, throughput, area, and power analyses that expose the basis-evaluation tax of spline-based architectures.
\end{itemize}

\section{Background and Related Work}

\subsection{Quantization in Neural Networks}
Quantization reduces the memory and compute cost of neural networks by mapping floating-point values to lower-precision representations \citep{gholami2021survey}. Existing approaches are typically divided into quantization-aware training (QAT), which injects fake quantization during training, and post-training quantization (PTQ), which quantizes pretrained models using calibration and/or reconstruction. Representative QAT methods include DoReFa, PACT, QIL, LSQ, LSQ+, and DSQ \citep{zhou2016dorefa,choi2018pact,jung2019qil,esser2019lsq,bhalgat2020lsq+,gong2019dsq}, while representative PTQ methods include AdaRound, BRECQ, GPTQ, HAWQ/HAWQ-V2, AWQ, SmoothQuant, and ZeroQ \citep{nagel2020adaround,li2021brecq,frantar2022gptq,dong2020hawq,yao2021hawqv2,lin2023awq,xiao2023smoothquant,cai2020zeroq}. Although these methods are highly effective for standard architectures, they do not directly address the function-based parameterization and heterogeneous branch structure of KANs.

\subsection{Kolmogorov--Arnold Networks (KANs)}
KANs are motivated by the Kolmogorov--Arnold representation theorem and replace scalar weights with learnable univariate functions, typically parameterized by splines or related bases \citep{liu2024kan}. Compared to MLPs, this shifts nonlinear modeling capacity onto edges, improving interpretability and expressive flexibility. Representative variants include EfficientKAN, FastKAN, PyKAN, and convolutional KAGN \citep{blealtan_efficientkan_github,li2024kolmogorovarnold,liu2024kan,drokin2024kolmogorov,khochawongwat2024gramkan}. At the same time, recent work suggests that KAN advantages can be domain-dependent relative to matched MLP baselines \citep{yu2024kanbefair}.

Recent work on KAN efficiency has progressed along several complementary fronts. Early studies characterized the software and hardware limitations
of KANs~\citep{le2024exploring}, and subsequent hardware-oriented efforts developed LUT- and CIM-based deployment flows such as
KANEL\'E~\citep{hoang2026kanele} and Huang et al.~\citep{huang2025hardware}, which target LUT-native or analog inference on specialized fabrics and face scalability challenges on deeper models. A separate line reduces KAN memory through coefficient clustering or vector-quantized sharing, including MetaCluster~\citep{raffel2025metacluster} and SHARe-KAN~\citep{smith2025share}; these methods compress along an orthogonal axis (the number of distinct coefficient vectors), and the MetaCluster authors explicitly identify bit-width quantization as a complementary technique that can be stacked on top. The closest prior work, KANtize, applies standard uniform quantization with B-spline tabulation for efficient inference~\citep{errabii2026kantize}. These efforts are either hardware-specific or method-specific: none unifies QAT and PTQ, applies branch-aware quantization to KANs' heterogeneous components, or compares systematically across KAN families and datasets. QuantKAN fills this gap, targeting standard integer datapaths and scaling to ImageNet-class KAGN-VGG models; a detailed positioning is provided in Appendix~\ref{app:related-compression}.

\textbf{Our objective is to characterize how KANs respond to quantization, not to adjudicate KAN-versus-MLP performance.} As a relatively new architecture under active development, KANs offer improved interpretability and task-dependent advantages on scientific problems~\citep{liu2024kan,yu2024kanbefair}; understanding their low-bit behavior is therefore a prerequisite for efficient deployment, regardless of how they ultimately compare to MLPs on any given task. A fully matched comparison against quantized MLPs is complementary to this work but lies outside its main scope.

\section{The QuantKAN Framework}
\label{sec:framework}

QuantKAN is a unified framework for studying and applying low-bit quantization to KAN layers across multiple parameterizations. Rather than treating KAN layers as conventional linear modules, QuantKAN captures how quantization affects their base, spline, and basis components. This is important because each KAN layer decomposes into two distinct pathways: a \emph{base branch} that applies learned linear weights to a simple activation, and a \emph{spline/basis branch} that maps inputs through nonlinear basis functions (e.g., B-splines, RBFs, or Gram polynomials) before linearly combining the resulting basis activations with learned coefficients. These branches serve distinct roles and exhibit different parameter statistics and quantization sensitivities.

\begin{figure}[!ht]
\centering
\includegraphics[width=0.65\linewidth]{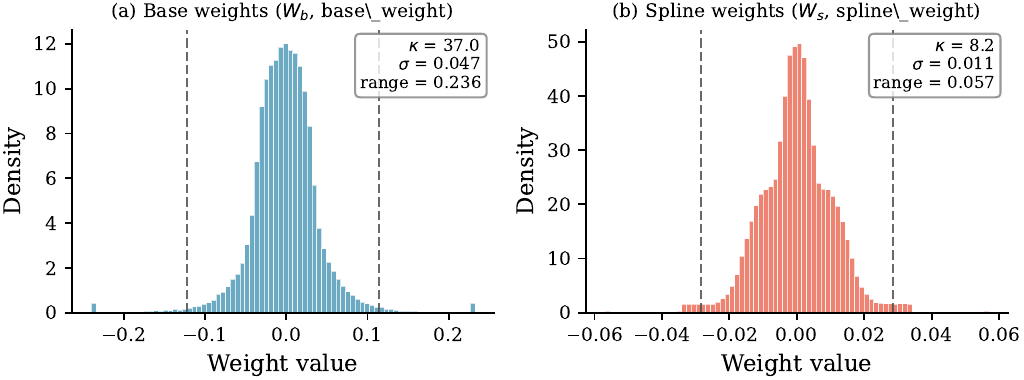}
\caption{Branch-wise weight distributions in a trained EfficientKAN.
Base weights exhibit heavier tails and a wider dynamic range ($\approx 0.236$) than spline/basis weights ($\approx 0.057$), a scale difference of about $4.1\times$. Dashed lines indicate the 1st and 99th percentiles.}
\label{fig:weight_dist}
\end{figure}

We empirically observe in Figure~\ref{fig:weight_dist} that pronounced distributional heterogeneity across KAN branches: base weights often exhibit heavier tails and outliers, while spline/basis parameters are frequently more concentrated, though this pattern can vary by architecture (see Appendix~\ref{app:weight_distributions}). Importantly, distributional width alone does \emph{not} reliably predict quantization sensitivity; our sensitivity and ablation results (Section~\ref{parameter_sensitivity}) show that the most sensitive
branch varies by architecture and is not predictable from distributional
width alone. This combination makes naive ``single-quantizer'' strategies unreliable. 

\textbf{\textit{Branch-aware quantization.}} To avoid coupling statistically distinct branches under a single scale, QuantKAN maintains \emph{independent quantizers} for (i) base weights, (ii) spline/basis weights, and (iii) activations. Let $x$ denote the layer input, $W_b$ the base weights, $W_s$ the spline/basis coefficients, $\sigma(\cdot)$ the base activation (e.g., SiLU), and $\Phi(x)$ the basis expansion. A full-precision KAN layer computes

\begin{equation}
y = f(x) = \underbrace{W_b \cdot \sigma(x)}_{\text{base branch}} \;+\;
\underbrace{W_s \cdot \Phi(x)}_{\text{spline/basis branch}}
\label{eq:kan_forward}
\end{equation}

Under quantization, we apply separate quantizers to each component:

\begin{equation}
\hat{y} = f_q(x) =
Q_b(W_b)\, \sigma\!\big(Q_a(x)\big) \;+\;
Q_s(W_s)\, \Phi\!\big(Q_a(x)\big),
\label{eq:kan_quantized}
\end{equation}

where each quantizer $Q(\cdot;\theta)$ is parameterized by learnable and/or calibrated parameters $\theta$ (e.g., step size, clipping bounds, optional zero-point), optimized \emph{independently} per branch. In our experiments, collapsing branches into a shared quantizer destabilizes scaling and degrades low-bit accuracy, whereas branch-aware quantization consistently improves robustness. Branch-aware quantization improves PyKAN over a shared quantizer by up to $+23.86$ on MNIST and $+17.17$ on CIFAR-10 at W2, while reducing quantization error on sensitive coefficient tensors by $5$--$10\times$ (Appendix~\ref{app:branch_vs_shared}).


\subsection{Quantization-Aware Training for KANs}
\label{sec:qat}

QAT integrates quantizers into the forward pass while approximating gradients through the discrete operations using a straight-through estimator (STE). For KANs, this setup is more delicate than in CNNs/Transformers due to the dual-branch structure and heterogeneous statistics. QuantKAN therefore quantizes three components separately: base weights, spline/basis weights, and activations. Concretely, the quantized forward pass follows Eq.~\eqref{eq:kan_quantized}. Figure~\ref{fig:qat_ptq_kan}(a) illustrates the QAT dataflow where both branches are quantized independently and their outputs are summed.
 
We evaluate and compare QAT methods that represent complementary design axes that are especially relevant for KANs:
(i) \emph{step-size learning} (LSQ/LSQ+) to adapt quantization scales to heterogeneous branch distributions;
(ii) \emph{learned clipping and interval shaping} (PACT, QIL) to control outliers and enforce effective dynamic
ranges; (iii) \emph{non-adaptive baselines} (DoReFa) to quantify the benefit of learning quantizer parameters;
and (iv) \emph{smooth relaxation of rounding} (DSQ) to improve optimization stability when spline/basis pathways
are quantized aggressively. All methods use branch-aware quantization with independent base and spline parameters, as shared quantizers often mismatch KAN statistics and destabilize low-precision training.

\subsection{Post-Training Quantization (PTQ) for KANs}
\label{sec:ptq}

PTQ compresses pretrained models without retraining, using small calibration sets or data-free synthesis. For KANs, PTQ is challenging because base and spline parameters exhibit different dynamic ranges and curvature, causing branch-specific error accumulation. QuantKAN therefore calibrates or reconstructs base and spline branches separately (Figure~\ref{fig:qat_ptq_kan}(b)).

Let $Q(\cdot)$ denote a branch-wise quantizer parameterized by scale(s) and optional zero-point(s).
\begin{equation}
\hat{W}_b = Q(W_b; \alpha_b, z_b),
\qquad
\hat{W}_s = Q(W_s; \alpha_s, z_s),
\end{equation}
and optimize them with a branch-aware reconstruction objective:

\begin{equation}
\begin{aligned}
\mathcal{L}_{\text{PTQ}} ={}& \lambda_b \big\| f_b(x;\hat{W}_b)-f_b(x;W_b)\big\|_2^2 + \lambda_s \big\| f_s(x;\hat{W}_s)-f_s(x;W_s)\big\|_2^2 \\
&+ \big\| f(x;\hat{W}_b,\hat{W}_s)-f(x;W_b,W_s)\big\|_2^2 .
\end{aligned}
\label{eq:ptq_loss}
\end{equation}

where $f_b$ and $f_s$ denote the base and spline/basis branches and $f$ is the full layer mapping.
Different PTQ methods (e.g., AdaRound, BRECQ, GPTQ, AWQ, SmoothQuant, HAWQ-V2, ZeroQ) instantiate this
objective with different calibration and optimization strategies; QuantKAN adapts each method by
preserving branch separation throughout calibration/reconstruction.


We evaluate PTQ methods spanning the major families used in low-bit deployment. Uniform and percentile calibration serve as baselines for scale selection. AdaRound and BRECQ provide reconstruction-based functional-error minimization, which is important when spline/basis parameters are sensitive to small perturbations. AWQ and SmoothQuant apply outlier- and activation-aware scaling, relevant when basis expansions produce channel imbalance. GPTQ uses second-order error compensation, with curvature estimated separately per branch in QuantKAN. ZeroQ provides a data-free fallback when calibration data is limited. 


\begin{figure}[!ht]
\centering

\begin{subfigure}[t]{0.45\columnwidth}
\centering
\resizebox{\linewidth}{!}{%
\begin{tikzpicture}[
  node distance=0.55cm and 0.45cm,
  every node/.style={font=\scriptsize},
  box/.style={
    draw,
    rounded corners=1.2pt,
    minimum height=0.42cm,
    minimum width=1.15cm,
    align=center,
    inner sep=1.2pt
  },
  iobox/.style={box, fill=gray!12},
  quantbox/.style={box, fill=green!10},
  actbox/.style={box, fill=orange!12},
  weightbox/.style={box, fill=blue!10},
  sumnode/.style={
    draw,
    circle,
    inner sep=0pt,
    minimum size=0.28cm,
    font=\scriptsize,
    fill=gray!8
  },
  arr/.style={->, >=stealth, semithick}
]
  \node[iobox] (x) {Input\\$\mathbf{x}$};
  \node[quantbox, right=0.50cm of x] (Qa) {Act.\ Quant.\\$Q_{\!a}(\cdot)$};
  \draw[arr] (x) -- (Qa);

  \node[actbox, above right=0.05cm and 0.50cm of Qa] (base) {Nonlin.\\$\sigma(\cdot)$};
  \node[weightbox, right=0.40cm of base] (Wb) {Weight\\$\hat{\mathbf{W}}_{\!b}$};

  \node[actbox, below right=0.05cm and 0.50cm of Qa] (spline) {Spline\\$\boldsymbol{\Phi}(\cdot)$};
  \node[weightbox, right=0.40cm of spline] (Ws) {Weight\\$\hat{\mathbf{W}}_{\!s}$};

  \draw[arr] (Qa.east) -- ++(0.14cm,0) |- (base.west);
  \draw[arr] (Qa.east) -- ++(0.14cm,0) |- (spline.west);

  \draw[arr] (base) -- (Wb);
  \draw[arr] (spline) -- (Ws);

  \node[sumnode, right=0.45cm of Wb, yshift=-0.65cm] (plus) {$\!+\!$};

  \draw[arr] (Wb.east) -| (plus.north);
  \draw[arr] (Ws.east) -| (plus.south);

  \node[iobox, right=0.30cm of plus] (y) {Output\\$\hat{\mathbf{y}}$};
  \draw[arr] (plus) -- (y);
\end{tikzpicture}%
}
\caption{QAT dataflow.}
\label{fig:qat_kan}
\end{subfigure}
\hfill
\begin{subfigure}[t]{0.45\columnwidth}
\centering
\resizebox{\linewidth}{!}{%
\begin{tikzpicture}[
  node distance=1.0cm and 0.9cm,
  every node/.style={font=\footnotesize},
  box/.style={
    draw,
    rounded corners=1.5pt,
    minimum height=0.55cm,
    minimum width=1.4cm,
    align=center,
    inner sep=2pt
  },
  prebox/.style={box, fill=gray!12},
  quantbox/.style={box, fill=green!10},
  qweightbox/.style={box, fill=blue!10},
  outbox/.style={box, fill=orange!12, minimum width=1.6cm},
  sumnode/.style={
    draw,
    circle,
    inner sep=0pt,
    minimum size=0.32cm,
    font=\footnotesize,
    fill=gray!8
  },
  arr/.style={->, >=stealth, semithick},
  elabel/.style={font=\scriptsize, midway, above}
]
  \node[prebox] (wb) {Pretrained\\$\mathbf{W}_{\!b}$};
  \node[prebox, below=0.75cm of wb] (ws) {Pretrained\\$\mathbf{W}_{\!s}$};

  \node[quantbox, right=1.0cm of wb] (qb) {Quantizer\\$Q_{\!b}$};
  \node[quantbox, right=1.0cm of ws] (qs) {Quantizer\\$Q_{\!s}$};

  \draw[arr] (wb) -- (qb) node[elabel] {calib.};
  \draw[arr] (ws) -- (qs) node[elabel] {calib.};

  \node[qweightbox, right=0.75cm of qb] (wbh) {Quantized\\$\hat{\mathbf{W}}_{\!b}$};
  \node[qweightbox, right=0.75cm of qs] (wsh) {Quantized\\$\hat{\mathbf{W}}_{\!s}$};

  \draw[arr] (qb) -- (wbh);
  \draw[arr] (qs) -- (wsh);

  \node[sumnode, right=0.9cm of wbh, yshift=-0.75cm] (sum) {$\!+\!$};

  \draw[arr] (wbh.east) -| (sum.north);
  \draw[arr] (wsh.east) -| (sum.south);

  \node[outbox, right=0.7cm of sum] (out) {Quantized\\Output};
  \draw[arr] (sum) -- (out);
\end{tikzpicture}%
}
\caption{PTQ flow.}
\label{fig:ptq_kan}
\end{subfigure}

\vspace{-0.2em}
\caption{Comparison of quantization workflows for a KAN layer. (a) Quantization-aware training (QAT): input activations are quantized via $Q_{\!a}$, then processed through parallel base ($\sigma \!\to\! \hat{\mathbf{W}}_{\!b}$) and spline ($\boldsymbol{\Phi} \!\to\! \hat{\mathbf{W}}_{\!s}$) branches. Independent quantizers enable mixed-precision allocation across branches. (b) Post-training quantization (PTQ): base and spline weights are independently quantized via calibration-based quantizers $Q_{\!b}$ and $Q_{\!s}$, then combined to produce the final quantized output.}
\label{fig:qat_ptq_kan}
\vspace{-0.4em}
\end{figure}


\section{Experimental Setup}
\label{sec:setup}

\textit{Models and datasets.} We evaluate multiple Kolmogorov--Arnold Network (KAN) variants, including PyKAN, EfficientKAN, FastKAN, GRAM-KAN, and KAGN-style convolutional KANs, spanning both fully connected and convolutional settings. Experiments are conducted on standard image classification benchmarks: MNIST, CIFAR-10/100, TinyImageNet, and ImageNet.

\textit{Training and evaluation.}All models are first trained in full precision. For quantization-aware training (QAT), we initialize from pretrained models and insert quantizers into the forward pass (Section~\ref{sec:framework}), then fine-tune for 5--20 epochs depending on dataset size. Post-training quantization (PTQ) is applied to the same pretrained models using calibration or reconstruction without task retraining. We report Top-1 test accuracy unless stated otherwise.

\textit{Quantization configuration.} We study weight-only and joint weight--activation quantization at 2--8 bits with independent quantizers for base weights, spline/basis weights, and activations. We denote configurations as $w_x$ for weight-only quantization with full-precision activations ($a{=}32$), and $w_xa_y$ for joint quantization, where $x$ and $y$ denote weight and activation bit-widths. We ablate per-tensor vs.\ per-channel scaling and symmetric vs.\ asymmetric quantization.

For QAT, we evaluate representative step-size, clipping-based, smooth-relaxation, and baseline methods (LSQ/LSQ+, PACT, QIL, DSQ, DoReFa). For PTQ, we compare calibration-based, reconstruction-based, activation-aware, second-order, and data-free approaches (Uniform, AdaRound, BRECQ, AWQ, SmoothQuant, GPTQ, HAWQ-V2, ZeroQ). All experiments are implemented in PyTorch in a unified configuration-driven codebase, with non-quantization settings fixed for fair comparison; full architectural details and hyperparameters are provided in Appendix~\ref{app:hyperparams}.


\section{Results and Analysis}

\subsection{KAN Performance with Quantization-Aware Training (QAT)}
\noindent
Table~\ref{tab:qat_top_3}  reports the top-3 quantization-aware training (QAT) methods for each architecture and dataset, selected based on their robustness under progressively lower bit-precision. To ensure a fair and concise comparison in the main paper, LSQ and LSQ+ are treated as a single method family, and only the stronger variant is retained when both appear among the top performers. The selection emphasizes stability at low-bit regimes, including w3, w2, and joint weight--activation quantization, where performance degradation is most pronounced for spline-based models; full results across all methods are provided in Table~\ref{tab:kan_qat} (Appendix \ref{app:full_qat_res}).

\begin{table}[!ht]
\centering
\scriptsize
\caption{Top-3 QAT methods per architecture, selected based on robustness under low-bit precision. 
When both LSQ and LSQ+ rank among the top performers, only the stronger variant is retained. 
Values $\approx\!1\%$ indicate random-class collapse, i.e., complete failure of the quantized model.}
\resizebox{\columnwidth}{!}{%
\begin{tabular}{lllcccccccc}
\toprule
\textbf{Dataset} & \textbf{Architecture} & \textbf{Method} 
& \textbf{w32a32} & \textbf{w8} & \textbf{w4} & \textbf{w3} & \textbf{w2} 
& \textbf{w4a4} & \textbf{w3a3} & \textbf{w2a2} \\
\midrule

\multirow{9}{*}{MNIST}
& \multirow{3}{*}{KAN FCN}
& DSQ    & 98.00 & 98.43 & 98.24 & 98.13 & 97.71 & 97.71 & 97.82 & 97.71 \\
& & LSQ+   & 98.00 & 98.24 & 98.13 & 98.03 & 97.79 & 97.87 & 97.73 & 96.97 \\
& & PACT   & 98.00 & 98.17 & 98.14 & 98.07 & 97.21 & 97.77 & 97.64 & 96.09 \\
\cmidrule(lr){2-11}
& \multirow{3}{*}{KAN ConvNet}
& DoReFa & 95.70 & 95.41 & 95.67 & 95.17 & 93.55 & 91.09 & 91.08 & 88.52 \\
& & DSQ    & 95.70 & 95.81 & 95.44 & 95.62 & 89.54 & 90.40 & 90.14 & 89.54 \\
& & LSQ+   & 95.70 & 95.70  & 95.56 & 95.17 & 94.46 & 90.95 & 94.11 & 92.04 \\
\cmidrule(lr){2-11}
& \multirow{3}{*}{KAGN Simple}
& DoReFa & 98.67 & 98.70 & 98.71 & 98.58 & 98.33 & 97.93 & 97.96 & 96.22 \\
& & DSQ    & 98.67 & 98.43 & 98.24 & 98.64 & 85.14 & 96.66 & 93.28 & 85.14 \\
& & LSQ    & 98.67 & 98.95 & 98.69 & 98.74 & 98.34 & 86.25 & 78.42 & 76.32 \\
\midrule

\multirow{6}{*}{CIFAR-10}
& \multirow{3}{*}{KAGN Simple}
& DSQ    & 62.24 & 62.77 & 62.88 & 61.31 & 45.44 & 48.60 & 48.33 & 45.44 \\
& & DoReFa & 62.24 & 62.84 & 61.68 & 58.88 & 55.87 & 50.01 & 47.38 & 37.15 \\
& & LSQ    & 62.24 & 62.74 & 61.74 & 61.70 & 56.03 & 28.27 & 30.48 & 29.24 \\
\cmidrule(lr){2-11}
& \multirow{3}{*}{KAGN Simple (8L)}
& DSQ    & 78.10 & 72.22 & 76.86 & 77.36 & 39.44 & 50.64 & 48.05 & 39.44 \\
& & LSQ    & 78.10 & 76.00 & 75.49 & 73.21 & 72.10 & 46.41 & 47.67 & 47.38 \\
& & DoReFa & 78.10 & 76.87 & 70.27 & 64.99 & 52.49 & 49.79 & 38.00 & 30.57 \\
\midrule

\multirow{6}{*}{CIFAR-100}
& \multirow{3}{*}{KAGN Simple (8L V2)}
& DSQ    & 68.06 & 67.12 & 68.03 & 67.54 & 43.10 & 34.01 & 35.45 & 43.10 \\
& & LSQ    & 68.06 & 67.20 & 65.87 & 68.49 & 68.41 & 52.36 & 53.01 & 47.00 \\
& & DoReFa & 68.06 & 67.93 & 66.89 & 53.58 & 23.39 & 29.49 & 9.66  & 8.62 \\
\cmidrule(lr){2-11}
& \multirow{3}{*}{VGG-like KAGN V2}
& DSQ    & 58.77 & 57.82 & 58.58 & 58.23 & 1.00  & 56.81 & 55.86 & 1.00 \\
& & LSQ+   & 58.77 & 58.05 & 56.98 & 1.00  & 30.18 & 1.00  & 1.00  & 1.01 \\
& & DoReFa & 58.77 & 58.50 & 48.96 & 1.00  & 21.61 & 18.37 & 1.00  & 1.08 \\
\bottomrule
\end{tabular}%
}
\label{tab:qat_top_3}
\end{table}

Observation 1: \textbf{\textit{DSQ is frequently the strongest or most stable QAT method}, especially under several joint quantization settings, but LSQ/LSQ+ and DoReFa outperform it in some architecture–dataset–bitwidth combinations.} DSQ either matches or exceeds full-precision accuracy at moderate bit-widths and exhibits a markedly more graceful degradation as precision is reduced. This behavior is especially evident for deeper and more structured KAN variants (e.g., KAGN Simple (8L) and CIFAR-100 models), suggesting that DSQ's smooth, differentiable formulation aligns well with the non-linear nature of spline-based layers.

Observation 2: \textbf{The \textit{LSQ/LSQ+} family remains competitive at moderate precision}, often achieving strong performance at w8 and w4 across architectures (Table~\ref{tab:qat_top_3}). Robustness degrades at very low precision, particularly under joint weight–activation quantization (w4a4 and below), where several models show sharp accuracy drops. LSQ+ is retained only when it improves stability, indicating that its added flexibility is beneficial but not consistently reliable across KAN/KAGN variants. 

Observation 3: \textbf{\textit{DoReFa consistently ranks among the top-3 methods} for simpler or moderately deep architectures, especially on MNIST and CIFAR-10.} While it does not always achieve the highest accuracy at higher precision, DoReFa demonstrates comparatively stable behavior under low-bit weight quantization and remains competitive when other methods experience catastrophic failure. This makes DoReFa a strong and reliable baseline for KAN-based models where simplicity and robustness are prioritized.


Other QAT methods such as PACT and QIL collapse at low precision under joint weight--activation quantization (Table~\ref{tab:kan_qat}), exposing the limits of clipping- and interval-based schemes on spline architectures. \textbf{\textit{Structure-aware and smooth quantization is therefore critical for KANs}}, especially at greater depth and dataset complexity: DSQ's consistent dominance motivates it as the default QAT strategy, with LSQ/LSQ+ and DoReFa as competitive alternatives depending on precision targets.


\textbf{\textit{Large-scale validation}}. Table~\ref{tab:qat_tinyimagenet_imagenet} extends the QAT comparison to TinyImageNet and ImageNet on deep VGG-style KAGN architectures. DSQ remains the most robust method across all precisions, while LSQ+ degrades sharply under joint quantization and DoReFa under quantized activations. These results confirm that smooth, structure-aware quantization persists at scale, reinforcing DSQ as a reliable default for deep KAN deployment.

\begin{table}[!ht]
\centering
\scriptsize
\caption{Top QAT methods under low-bit precision for large datasets. FP denotes full precision.}
\begin{tabular}{l l c c c c c }
\toprule
\textbf{Dataset / Arch.} & \textbf{Method} 
& \textbf{FP} & \textbf{w8} & \textbf{w4} & \textbf{w3} & \textbf{w4a4} \\
\midrule
\multirow{3}{*}{\shortstack[l]{TinyImageNet/\\VGG-KAGN V2}}
& DSQ    & \multirow{3}{*}{42.64} & 44.21 & 44.10 & 44.70 & 43.90  \\
& LSQ+   &       & 43.29 & 43.90 & 43.78 & 30.25  \\
& DoReFa &       & 45.07 & 43.02 & 40.55 & 35.05 \\
\midrule
\multirow{2}{*}{\shortstack[l]{ImageNet/\\VGG-KAGN V4}}
& DSQ    & \multirow{2}{*}{61.15} & 60.93 & 61.39 & 60.47 & 55.19  \\
& LSQ+   &       & 60.01 & 55.13 & 42.98    & 29.28     \\
\bottomrule
\end{tabular}
\label{tab:qat_tinyimagenet_imagenet}
\end{table}


\subsection{KAN Performance with Post-Training Quantization}
\noindent

Table~\ref{tab:ptq_top} summarizes the performance of representative post-training quantization (PTQ) methods across multiple KAN architectures and datasets.  For clarity in the main paper, we focus on the four strongest PTQ approaches: GPTQ, AdaRound, AWQ, and BRECQ, selected based on robustness under reduced bit precision and consistency across architectures.  The complete PTQ results, including additional baselines and all precision settings, are reported in Appendix Table~\ref{tab:ptq_kan_all}.

\begin{table}[!ht]
\centering
\scriptsize
\caption{Top-4 PTQ methods per architecture, selected based on robustness under reduced precision.
Values $\approx\!1\%$ indicate random-class collapse, i.e., complete failure of the quantized model.}
\resizebox{\columnwidth}{!}{%
\begin{tabular}{lllcccccccc}
\toprule
\textbf{Dataset} & \textbf{Architecture} & \textbf{Method}
& \textbf{w32a32} & \textbf{w8} & \textbf{w4} & \textbf{w3} & \textbf{w2}
& \textbf{w4a4} & \textbf{w3a3} & \textbf{w2a2} \\
\midrule

\multirow{12}{*}{MNIST}
& \multirow{4}{*}{KAN FCN}
& GPTQ     & 97.99 & 98.00 & 97.86 & 97.07 & 76.58 & 95.88 & 88.99 & 40.90 \\
& & AdaRound & 97.99 & 97.97 & 97.78 & 96.36 & 42.87 & 95.75 & 88.88 & 32.35 \\
& & AWQ      & 97.99 & 97.98 & 97.80 & 95.53 & 33.65 & 95.55 & 88.73 & 21.02 \\
& & BRECQ    & 97.99 & 73.77 & 72.58 & 73.60 & 50.51 & 70.87 & 86.93 & 52.38 \\
\cmidrule(lr){2-11}
& \multirow{4}{*}{KAN ConvNet}
& GPTQ     & 95.69 & 95.69 & 95.65 & 95.63 & 88.18 & 10.10 & 10.10 & 10.10 \\
& & AdaRound & 95.69 & 95.69 & 95.61 & 95.20 & 80.03 & 9.73  & 10.28 & 10.25 \\
& & AWQ      & 95.69 & 95.68 & 95.62 & 95.01 & 62.70 & 10.10 & 10.10 & 10.32 \\
& & BRECQ    & 95.69 & 36.37 & 37.79 & 30.31 & 35.07 & 75.60 & 93.52 & 64.97 \\
\cmidrule(lr){2-11}
& \multirow{4}{*}{KAGN Simple}
& GPTQ     & 98.66 & 98.65 & 97.96 & 90.11 & 27.39 & 97.96 & 90.11 & 27.39 \\
& & AdaRound & 98.66 & 98.76 & 96.59 & 88.41 & 29.83 & 96.59 & 88.41 & 29.83 \\
& & AWQ      & 98.66 & 98.70 & 97.35 & 57.65 & 13.36 & 97.35 & 57.65 & 13.36 \\
& & BRECQ    & 98.66 & 98.73 & 96.11 & 76.07 & 23.33 & 96.22 & 81.70 & 12.80 \\
\midrule

\multirow{8}{*}{CIFAR-10}
& \multirow{4}{*}{KAGN Simple}
& GPTQ     & 62.23 & 62.36 & 56.52 & 40.85 & 11.68 & 56.52 & 40.85 & 11.68 \\
& & AdaRound & 62.23 & 62.27 & 53.92 & 34.21 & 11.61 & 53.92 & 34.21 & 11.61 \\
& & AWQ      & 62.23 & 62.17 & 54.01 & 22.01 & 9.58  & 54.01 & 22.01 & 9.58  \\
& & BRECQ    & 62.23 & 62.28 & 52.28 & 24.47 & 12.35 & 49.98 & 27.43 & 16.21 \\
\cmidrule(lr){2-11}
& \multirow{4}{*}{KAGN Simple (8L)}
& GPTQ     & 78.08 & 78.02 & 69.13 & 41.59 & 9.92  & 69.13 & 41.59 & 9.92  \\
& & AdaRound & 78.08 & 78.15 & 61.57 & 37.93 & 10.09 & 61.57 & 37.93 & 10.09 \\
& & AWQ      & 78.08 & 77.92 & 60.17 & 28.01 & 12.17 & 60.17 & 28.01 & 12.17 \\
& & BRECQ    & 78.08 & 77.88 & 60.28 & 27.64 & 12.82 & 62.28 & 39.31 & 19.20 \\
\midrule

\multirow{8}{*}{CIFAR-100}
& \multirow{4}{*}{KAGN Simple (8L V2)}
& GPTQ     & 68.09 & 68.05 & 13.47 & 2.65  & 1.14  & 13.47 & 2.65  & 1.14  \\
& & AdaRound & 68.09 & 67.67 & 2.77  & 1.01  & 1.00  & 2.77  & 1.01  & 1.00  \\
& & AWQ      & 68.09 & 67.62 & 3.74  & 1.00  & 1.00  & 3.74  & 1.00  & 1.00  \\
& & BRECQ    & 68.09 & 67.57 & 3.86  & 1.00  & 1.00  & 3.12  & 1.49  & 1.47  \\
\cmidrule(lr){2-11}
& \multirow{4}{*}{VGG-like KAGN V2}
& GPTQ     & 58.70 & 58.23 & 39.73 & 14.23 & 1.98  & 39.57 & 13.26 & 0.99  \\
& & AdaRound & 58.70 & 55.91 & 22.66 & 6.05  & 1.47  & 22.63 & 5.46  & 1.03  \\
& & AWQ      & 58.70 & 58.30 & 29.57 & 13.44 & 1.22  & 29.44 & 12.48 & 1.15  \\
& & BRECQ    & 58.70 & 58.29 & 29.80 & 15.64 & 1.88  & 32.17 & 15.99 & 2.57  \\
\bottomrule
\end{tabular}%
}
\label{tab:ptq_top}
\end{table}


Observation 4: \textbf{GPTQ is the most reliable PTQ method overall, typically preserving near full-precision accuracy at W8 and outperforming other PTQ methods in many W4 settings.} However, W4 performance is architecture-dependent: GPTQ remains strong for MNIST and several CIFAR-10 cases, but degrades sharply on deeper CIFAR-100 KAGN models. This indicates that second-order error compensation is useful for KAN PTQ, but calibration-only PTQ remains insufficient for some deeper or more complex settings.

Observation 5: \textbf{\textit{AdaRound and AWQ form a second tier of competitive PTQ methods}, performing well at moderate precision (w8 and w4) but degrading more rapidly as bit-width decreases.} AdaRound generally outperforms AWQ at lower precisions, suggesting that learned rounding decisions provide greater robustness than activation-aware scaling alone. Nonetheless, both methods struggle under joint weight--activation quantization, especially for deeper KAN models and more complex datasets such as CIFAR-100.

Observation 6: \textbf{\textit{BRECQ exhibits mixed but notable behavior}.} While it underperforms GPTQ and AdaRound on simpler architectures, it remains comparatively robust for convolutional and VGG-like KAN variants, occasionally outperforming other PTQ methods under joint quantization.
This suggests that block-wise reconstruction can partially mitigate quantization error in structured architectures, albeit with limited stability at very low bit-widths.

Other PTQ baselines such as Uniform quantization, HAWQ-V2, and ZeroQ (see Table~\ref{tab:ptq_kan_all}) frequently \textit{collapse} under low-bit or joint quantization, particularly for deeper KAN models. \textit{Overall}, while PTQ is effective for KANs at moderate precision, \textit{its reliability deteriorates rapidly as bit-width decreases}, especially for deeper architectures and joint quantization; without the ability to adapt spline parameters during training, PTQ is less suitable than QAT for aggressive low-bit deployment.

\subsection{Quantization Design Analysis Across KAN Variants}
\label{sec:quant_design_analysis}
Table~\ref{tab:combined_analysis} summarizes the marginal effects of quantization granularity, symmetry, and targets on CIFAR-10 (full results in Appendix Tables~\ref{tab:ablation_full_cifar10_weight_methodcols} and~\ref{tab:ablation_full_cifar10_activation_methodcols}, with granularity heatmaps in Figures~\ref{fig:ablation_heat_weight}--\ref{fig:ablation_heat_act}). For activations, per-channel scaling consistently improves accuracy, notably for PyKAN ($\Delta_{\text{ch-t}}^{(A)}{=}+0.20$). Conversely, per-channel weight scaling is variant-dependent and catastrophic for FastKAN ($\Delta_{\text{ch-t}}^{(W)}{=}-0.74$), indicating poorly conditioned channel-wise scales. Symmetry effects are smaller and variant-dependent (max $|\Delta_{\text{sym-asym}}|{=}0.20$), while FastKAN and GRAM clearly favor weight quantization ($\Delta_{\text{w-a}}{=}+0.17,+0.15$); EfficientKAN and PyKAN show negligible target preference.

\subsection{Parameter Sensitivity and Mixed-Precision Design}
\label{parameter_sensitivity}
We evaluate parameter vulnerability via single-parameter mixed-precision ablations, quantizing one group to $b\in\{2,3,4\}$ bits while keeping others at $8$-bit. DSQ is universally robust (near-zero degradation across datasets and parameters), whereas LSQ exposes strong parameter dependence on CIFAR-10 but not MNIST (Appendix~\ref{sec:appendix_sensitivity}, Table~\ref{tab:appendix_param_sensitivity_full}). Table~\ref{tab:combined_analysis} shows that the most sensitive parameter group varies by architecture under LSQ---basis for FastKAN, base for EfficientKAN and GRAM, scaling for PyKAN---with no single branch dominating. This motivates an architecture-specific policy: allocate $8$ bits to the most sensitive group and aggressively compress the rest, protecting $w_{\text{base}}$ for EfficientKAN and GRAM, $w_{\text{spline}}$ for FastKAN, and $w_{\text{scale,base}}$ for PyKAN.

\begin{table*}[!ht]
\centering
\scriptsize
\caption{Quantization design sensitivity and mixed-precision analysis on CIFAR-10. \emph{Left:} signed marginal effects $\Delta_{\text{ch-t}}^{(\cdot)}$, $\Delta_{\text{sym-asym}}^{(\cdot)}$, $\Delta_{\text{w-a}}$ averaged across bit-widths and methods (positive favors per-channel/symmetric/weight). \emph{Right:} most sensitive parameter under single-parameter mixed precision (LSQ), $\Delta{=}\max(0,\mathrm{Acc}_{\mathrm{FP}}{-}\mathrm{Acc}_q)$, and resulting policy. Policies (A/B/C) map to variant-specific groups: EfficientKAN $(w_{\text{base}}, w_{\text{spline}}, w_{\text{scaler}})$, FastKAN $(w_{\text{base}}, w_{\text{spline}})$, GRAM $(w_{\text{base}}, w_{\text{gram}}, w_{\text{beta}})$, PyKAN $(w_{\text{coef}}, w_{\text{scale,base}}, w_{\text{scale,sp}})$; for PyKAN, A denotes $w_{\text{coef}}$ rather than the base weight.}
\resizebox{\textwidth}{!}{
\begin{tabular}{l|ccccc|lccl}
\toprule
 & \multicolumn{5}{c|}{\textbf{Ablation Marginal Effects}} & \multicolumn{4}{c}{\textbf{Parameter Sensitivity \& Mixed-Precision Policy}} \\
\cmidrule{2-10}
\textbf{Variant} & $\Delta_{\text{ch-t}}^{(W)}$ & $\Delta_{\text{ch-t}}^{(A)}$ & $\Delta_{\text{sym-asym}}^{(W)}$ & $\Delta_{\text{sym-asym}}^{(A)}$ & $\Delta_{\text{w-a}}$ & \textbf{Most Sensitive} & \textbf{Mean $\Delta$} & \textbf{Max $\Delta$} & \textbf{Rec. Bits (A/B/C)} \\
\midrule
EfficientKAN & \phantom{$+$}0.00 & $+$0.07 & $-$0.06 & $+$0.03 & $-$0.01 & A ($w_{\text{base}}$)       & 0.14 & 0.43 & \textbf{A8 / B2 / C2} \\
FastKAN      & $-$0.74           & $+$0.10 & $+$0.12 & $-$0.15 & $+$0.17 & B ($w_{\text{spline}}$)     & 1.37 & 3.07 & \textbf{A2 / B8} \\
GRAM         & $-$0.12           & $+$0.10 & $-$0.20 & $-$0.07 & $+$0.15 & A ($w_{\text{base}}$)       & 0.99 & 2.92 & \textbf{A8 / B4 / C2} \\
PyKAN        & \phantom{$+$}0.00 & $+$0.20 & \phantom{$+$}0.00 & $-$0.02 & $-$0.05 & B ($w_{\text{scale,base}}$) & 0.22 & 0.66 & \textbf{A2 / B8 / C2} \\
\bottomrule
\end{tabular}
}
\label{tab:combined_analysis}
\end{table*}


\section{Hardware Efficiency}
\label{sec:hardware_efficiency}

We evaluate the deployment implications of QuantKAN by synthesizing representative KAN accelerators on a Xilinx Zynq UltraScale+ XCZU7EV FPGA using Vivado HLS. EfficientKAN, PyKAN, and FastKAN are implemented as two-layer MNIST classifiers, while GRAM is evaluated as a reusable GRAMLayer compute cell. To isolate arithmetic efficiency from memory-bandwidth effects, we adopt an \emph{iso-compute} methodology: each FP32 and W4A4 design processes the same logical parallelism per cycle, with low-bit weights packed into fixed-width memory words. The accelerators use AXI-stream weight movement, local basis precomputation, and pipelined accumulation to expose the dominant hardware bottlenecks of each KAN parameterization.

\textbf{Latency and throughput.} Table~\ref{tab:fpga_latency} reports latency (cycle count $\times$ estimated clock period) and throughput ($1/\text{latency}$). QuantKAN improves latency for all variants, but the source of the gain is architecture-dependent. EfficientKAN and PyKAN, both B-spline-based, preserve nearly the same cycle count as FP32 under W4A4 but reduce the clock period from $8.75$ ns to $6.125$ ns, yielding $1.40\times$ and $1.43\times$ throughput respectively; their gains are thus \emph{frequency-driven}, since recursive spline-basis evaluation keeps the cycle count almost unchanged. FastKAN, by contrast, reduces both clock period and cycle count, cutting latency from $5.28$ ms to $1.59$ ms and tripling throughput from $189.5$ to $627.4$ images/s ($3.32\times$). GRAMLayer follows a third pattern: its clock period stays fixed at $6.125$ ns, but W4A4 reduces the cycle count by $36.7\%$ for a $1.58\times$ throughput improvement.

\textbf{Resources, area, and power.} W4A4 reduces BRAM by $55.9$--$65.6\%$, confirming that low-bit packing is effective for learned KAN parameters. LUT and DSP changes are less uniform: spline-based designs still require basis-generation, activation, and casting logic, while GRAMLayer adds polynomial-control overhead. Using HLS resource counts and timing estimates, Appendix~\ref{app:fpga_area_power} reports first-order area and dynamic-power trend estimates under a bitwidth-aware activity model. Under this model, W4A4 reduces estimated area ($0.81$--$0.97\times$), dynamic power ($0.37$--$0.45\times$), and dynamic energy per inference ($0.13$--$0.32\times$) across the four variants, with FastKAN achieving the largest energy reduction because of its latency advantage. A sensitivity sweep over the arithmetic-fraction parameter preserves the qualitative W4A4 advantage across the swept activity assumptions. These results reveal a hardware-specific property of KANs: low-bit quantization substantially reduces learned-parameter storage and MAC cost, but the non-learned basis computation can remain a dominant residual overhead. We refer to this as the \emph{basis-evaluation tax}: hardware-efficient KAN deployment requires both branch-aware quantization and basis-aware microarchitecture.

\begin{table}[t]
\centering
\scriptsize
\caption{Latency and throughput on Xilinx Zynq UltraScale+ XCZU7EV. Cycles are normalized per inference from the HLS-reported maximum latency. Throughput is computed as $1/\text{latency}$.}
\label{tab:fpga_latency}
\resizebox{\columnwidth}{!}{%
\begin{tabular}{lccccc}
\toprule
\textbf{Variant} & \textbf{Clock FP/W4} & \textbf{Cycles FP/W4} & \textbf{Latency FP/W4} & \textbf{Throughput FP/W4} & \textbf{Speedup} \\
\midrule
EfficientKAN & 8.75/6.13 ns & 3.96M/4.03M & 34.62/24.72 ms & 28.9/40.5 img/s & 1.40$\times$ \\
PyKAN        & 8.75/6.13 ns & 3.96M/3.96M & 34.66/24.25 ms & 28.8/41.2 img/s & 1.43$\times$ \\
FastKAN      & 8.75/6.13 ns & 603K/260K   & 5.28/1.59 ms   & 189.5/627.4 img/s & 3.32$\times$ \\
GRAMLayer    & 6.13/6.13 ns & 428K/271K   & 2.62/1.66 ms   & 381.8/603.4 img/s & 1.58$\times$ \\
\bottomrule
\end{tabular}%
}
\end{table}

\textbf{Memory and compute savings.} To quantify deployment benefits beyond accuracy, we evaluate memory and bit-operation (BOP) savings on a representative two-layer EfficientKAN MNIST classifier ($784{\rightarrow}64{\rightarrow}10$, $G{=}5$, $K_s{=}3$), which contains $508{,}160$ parameters and executes $4.57{\times}10^{5}$ linear-branch MACs per inference (full derivation in Appendix~\ref{sec:bop_appendix}). For weight memory, all eight evaluated methods (LSQ, LSQ+, DoReFa, DSQ, GPTQ, AWQ, AdaRound, BRECQ) reduce storage uniformly to $b_w/32$ of the FP32 baseline, yielding $\mathbf{8\times}$ compression ($0.24$\,MB) at W4 and $\mathbf{16\times}$ ($0.12$\,MB) at W2, since memory depends only on the integer codebook width. For compute, the methods diverge: QAT methods reach full $W b_w {\times} A b_a$ INT arithmetic, with $\mathbf{64\times}$ linear-branch BOP reduction at W4A4 and $\mathbf{256\times}$ at W2A2, while weight-only PTQ (GPTQ, AWQ, and standard W$b_w$A16 AdaRound/BRECQ) reaches $16\times$ and $32\times$. Including the non-linear basis evaluation (held at INT16 per standard non-linear-block practice~\citep{kim2021bert,ijcai2022p164}) reduces the QAT ratios to $19.6\times$ and $24.0\times$, quantifying the basis-evaluation tax and motivating QAT for compute-bound deployment with basis-aware microarchitecture as the next lever.

\section{Limitations}
\label{sec:limitations}
Our benchmarks cover four KAN families on five image-classification datasets, the broadest systematic evaluation of KAN quantization to our knowledge. The architecture- and parameter-specific behaviors we report may not transfer unchanged to scientific regression, time-series, graph, or physics-informed settings where KANs are also applied; extending QuantKAN to non-vision domains is left to future work. Our FPGA evaluation (Section~\ref{sec:hardware_efficiency}, Appendix~\ref{app:fpga_latency}) is a controlled microarchitectural probe rather than a fully optimized accelerator, using Vivado HLS estimates instead of post-place-and-route timing or silicon-validated power. The area and power figures (Appendix~\ref{app:fpga_area_power}) are first-order estimates and should be read as relative trends; basis approximation, LUT-based evaluation, operator fusion, and a full \texttt{report\_power} flow could yield larger gains.

\section{Conclusion}
We present \textbf{QuantKAN}, the first unified framework for quantizing Kolmogorov--Arnold Networks under both QAT and PTQ. By explicitly modeling the dual-branch structure of KAN layers, QuantKAN adapts modern quantizers to spline- and polynomial-based components and establishes the first unified QAT/PTQ low-bit benchmark across multiple datasets, KAN variants, and quantization method families. DSQ emerges as the most robust QAT method and GPTQ as the strongest PTQ method at moderate precision, while parameter-wise sensitivity analyses identify architecture-specific
failure modes: spline/basis in FastKAN, base or scaling parameters elsewhere, motivating architecture-specific mixed-precision policies. FPGA synthesis further shows up to $3.32\times$ throughput and $7.7\times$ lower dynamic  energy under W4A4, exposing a residual \emph{basis-evaluation tax} that motivates basis-aware microarchitecture. By reducing the memory and compute footprint of interpretable spline-based models, QuantKAN broadens access to KANs in low-power and edge settings such as embedded and mobile devices, supports more energy-efficient inference, and lowers barriers to research and education on resource-constrained hardware. As a general-purpose compression toolkit that does not introduce new data or deployment contexts, it inherits no risks beyond those common to efficiency research.

\bibliographystyle{plainnat}
\bibliography{quantkan_nips}

\newpage
\appendix

\section*{Appendix Contents}
This appendix provides supporting analyses and implementation details for \textsc{QuantKAN}.
\Cref{app:weight_distributions} characterizes branch-specific weight distributions across major KAN variants and relates them to the sensitivity trends discussed in the main paper.
\Cref{app:branch_vs_shared} validates the central branch-aware design choice of \textsc{QuantKAN} through a controlled accuracy and per-branch quantization-error ablation against a shared-quantizer baseline.
\Cref{app:qat_quantizers,app:ptq} give implementation-aligned formulations and pseudocode for the QAT quantizers and PTQ methods evaluated in our experiments.
\Cref{app:hyperparams} lists the full experimental and quantization hyperparameters used for fair comparisons, together with the complete QAT and PTQ result tables and the granularity, symmetry, and quantization-target ablations.
\Cref{sec:appendix_sensitivity} reports complete parameter-wise sensitivity results that motivate our mixed-precision and branch-aware design choices.
\Cref{app:fpga_latency} details the FPGA latency, throughput, and resource analyses, including approximated area and power estimates that quantify deployment efficiency on a Xilinx Zynq UltraScale+ device.
Finally, \Cref{sec:bop_appendix} derives the memory and bit-operation (BOP) savings reported in the main paper and provides a per-method compute analysis across QAT and PTQ families.

\begin{itemize}
  \item \Cref{app:weight_distributions}: Weight Distribution Analysis Across KAN Variants
  \item \Cref{app:branch_vs_shared}: Branch-Aware Quantization: Mechanism and Ablation
  \item \Cref{app:qat_quantizers}: Quantization-Aware Training (QAT) Quantizers
  \item \Cref{app:ptq}: Post-Training Quantization (PTQ) Methods
  \item \Cref{app:hyperparams}: Experimental Hyperparameters, Full Results, and Ablations
  \item \Cref{sec:appendix_sensitivity}: Additional Parameter Sensitivity Results
  \item \Cref{app:fpga_latency}: FPGA Latency, Throughput, and Resource Analysis
  \item \Cref{sec:bop_appendix}: Memory and BOP Savings: Derivations and Per-Method Analysis
\end{itemize}


\section{Weight Distribution Analysis Across KAN Variants}
\label{app:weight_distributions}

This appendix provides a comprehensive analysis of weight distributions across four major KAN architectures: PyKAN, EfficientKAN, FastKAN, and GRAM based KAN. All models were trained on MNIST with identical hyperparameters (2-layer networks, 64 hidden units). Understanding these distributions, in conjunction with the parameter sensitivity analysis in Section~\ref{parameter_sensitivity}, informed the design of our branch-aware quantization framework.

\subsection{Summary of Distributional Patterns}

Table~\ref{tab:weight_stats_full} presents detailed statistics for base and spline weights
across all variants. Two patterns emerge that are relevant for quantization:

\begin{enumerate} 
\item \textbf{Base weights consistently exhibit heavier tails.} Across all four variants,
base weights show higher kurtosis ($\kappa$) than spline weights, ranging from $\kappa = 4.3$
(GRAM-KAN) to $\kappa = 37.0$ (EfficientKAN). This indicates the presence of outliers that
would dominate min-max calibration.

\item \textbf{Dynamic range relationships vary by architecture.} While base weights have
heavier tails, the relationship between branch ranges differs: EfficientKAN base weights
span $4\times$ the range of spline weights, whereas FastKAN shows the opposite pattern
with spline weights spanning $2\times$ the base range.
\end{enumerate}

These distributional statistics alone do not predict quantization sensitivity. Our parameter sensitivity analysis (Section~\ref{parameter_sensitivity}) reveals that \emph{spline and basis parameters are more sensitive to quantization error} than base weights, despite often having narrower distributions. This counterintuitive finding underscores the importance of empirical sensitivity analysis over distribution-based heuristics.

\begin{table}[!ht]
\centering
\caption{Weight distribution statistics across KAN variants (MNIST, 2-layer networks).
Kurtosis ($\kappa$) measures tail heaviness (Gaussian: $\kappa = 3$); range denotes the
1st--99th percentile span. Note: distributional width does not directly predict quantization
sensitivity---see Section~\ref{parameter_sensitivity} for sensitivity analysis.}
\label{tab:weight_stats_full}
\small
\begin{tabular}{@{}lcccccc@{}}
\toprule
& \multicolumn{3}{c}{\textbf{Base weights} ($W_b$)} & \multicolumn{3}{c}{\textbf{Spline weights} ($W_s$)} \\
\cmidrule(lr){2-4} \cmidrule(lr){5-7}
\textbf{Variant} & $\kappa$ & $\sigma$ & range & $\kappa$ & $\sigma$ & range \\
\midrule
PyKAN        & 36.0 & 0.047 & 0.240 & 2.7  & 0.041 & 0.184 \\
EfficientKAN & 37.0 & 0.047 & 0.236 & 8.2  & 0.011 & 0.057 \\
FastKAN      & 20.6 & 0.038 & 0.200 & 3.0  & 0.096 & 0.447 \\
GRAM-KAN     & 4.3  & 0.050 & 0.167 & 5.2  & 0.026 & 0.097 \\
\bottomrule
\end{tabular}
\end{table}

\subsection{PyKAN}
\label{app:pykan_dist}

PyKAN~\citep{liu2024kan} uses B-spline basis functions with learnable coefficients. The base
branch applies a scaling factor (\texttt{scale\_base}) to a SiLU activation, while the spline
branch combines B-spline coefficients (\texttt{coef}) with a separate scaling term (\texttt{scale\_sp}).

\begin{figure}[t]
\centering
\includegraphics[width=0.95\linewidth]{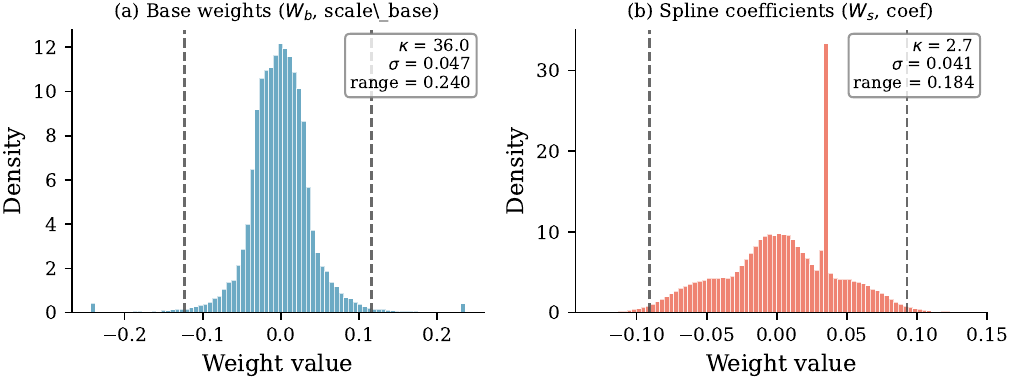}
\caption{Weight distributions in trained PyKAN. \textbf{(a)}~Base weights (\texttt{scale\_base})
show heavy tails ($\kappa = 36.0$) with outliers beyond $\pm 0.2$. \textbf{(b)}~Spline coefficients
(\texttt{coef}) are near-Gaussian ($\kappa = 2.7$) with a sharp peak. Dashed lines: 1st/99th percentiles.}
\label{fig:pykan_weights}
\end{figure}

Figure~\ref{fig:pykan_weights} shows the weight distributions. Base weights exhibit pronounced
heavy tails ($\kappa = 36.0$) with outliers at approximately $\pm 0.25$, while spline coefficients
follow a near-Gaussian distribution ($\kappa = 2.7$) with a characteristic sharp peak near zero.
The spline distribution shows slight positive skew and a distinct mode around 0.05, reflecting
the learned basis function structure.

\textit{Quantization implications:}
\begin{itemize} 
\item Base weights require outlier-robust calibration (99th percentile) due to heavy tails
\item Despite narrower distribution, \texttt{scale\_base} parameters show moderate sensitivity
      under LSQ (mean $\Delta = 0.33$, max $\Delta = 0.66$ on CIFAR-10)
\item DSQ achieves near-zero degradation across all parameter groups, making it the recommended
      QAT method for PyKAN
\end{itemize}

\subsection{EfficientKAN}
\label{app:efficientkan_dist}

EfficientKAN~\citep{blealtan_efficientkan_github} reimplements KANs for computational efficiency using matrix
operations. It maintains explicit \texttt{base\_weight} and \texttt{spline\_weight} tensors
with an optional \texttt{spline\_scaler}.

\begin{figure}[t]
\centering
\includegraphics[width=0.95\linewidth]{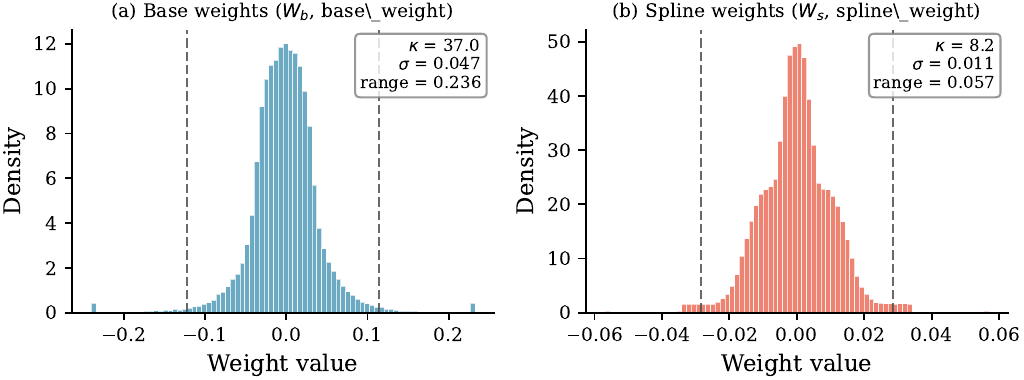}
\caption{Weight distributions in trained EfficientKAN. \textbf{(a)}~Base weights show the
heaviest tails among all variants ($\kappa = 37.0$). \textbf{(b)}~Spline weights are highly
concentrated ($\sigma = 0.011$) with $4\times$ smaller dynamic range.}
\label{fig:efficient_kan_weights}
\end{figure}

Figure~\ref{fig:efficient_kan_weights} reveals the most extreme distributional contrast among
the variants studied. Base weights have the highest kurtosis ($\kappa = 37.0$) with substantial
outliers, while spline weights are highly concentrated ($\kappa = 8.2$, $\sigma = 0.011$) with
a dynamic range only one-quarter that of base weights.

\textit{Quantization implications:}
\begin{itemize} 
\item The $4\times$ range difference makes shared quantization particularly harmful
\item Spline weights' narrow distribution ($\sigma = 0.011$) might suggest easy quantization,
      but \texttt{base\_weight} actually shows the highest sensitivity under LSQ
      (mean $\Delta = 0.14$, max $\Delta = 0.43$ on CIFAR-10)
\item Mixed-precision policy: retain 8-bit for \texttt{base\_weight}, aggressively quantize
      \texttt{spline\_weight} and \texttt{spline\_scaler}
\end{itemize}

\subsection{FastKAN}
\label{app:fastkan_dist}

FastKAN~\citep{li2024kolmogorovarnold} replaces B-splines with radial basis functions (RBFs) for faster
computation. The architecture uses standard linear layers for both branches: \texttt{base\_linear}
and \texttt{spline\_linear}.

\begin{figure}[t]
\centering
\includegraphics[width=0.95\linewidth]{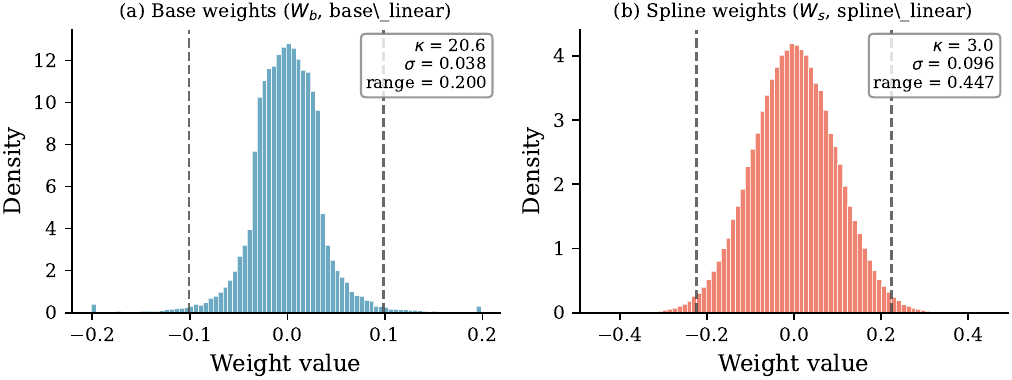}
\caption{Weight distributions in trained FastKAN. \textbf{(a)}~Base weights exhibit moderate
heavy tails ($\kappa = 20.6$). \textbf{(b)}~Spline weights are near-Gaussian ($\kappa = 3.0$)
but span $2\times$ the dynamic range of base weights. Despite wider range, spline weights
show the highest quantization sensitivity among all variants.}
\label{fig:fastkan_weights}
\end{figure}

Figure~\ref{fig:fastkan_weights} shows a pattern distinct from other variants. While base weights
exhibit elevated kurtosis ($\kappa = 20.6$), spline weights are nearly Gaussian ($\kappa = 3.0$)
but with a substantially \emph{wider} dynamic range---approximately $2\times$ that of base weights.

\textit{Quantization implications:}
\begin{itemize} 
\item FastKAN exhibits the highest parameter sensitivity among all variants: \texttt{spline\_linear}
      weights show mean $\Delta = 1.37$, max $\Delta = 3.07$ under LSQ on CIFAR-10
\item \textbf{Critical:} Per-channel weight quantization is harmful for FastKAN
      ($\Delta_{\text{ch-t}}^{(W)} = -0.74$), unlike other variants---use per-tensor weight scaling
\item Mixed-precision policy: retain 8-bit for \texttt{spline\_linear}, quantize \texttt{base\_linear}
      aggressively
\end{itemize}

\subsection{GRAM-KAN}
\label{app:gram_dist}

GRAM-KAN~\citep{drokin2024kolmogorov} uses Gram polynomial basis functions instead of splines.
The base branch uses \texttt{base\_weights}, while the polynomial branch uses
\texttt{grams\_basis\_weights} organized by polynomial degree.

\begin{figure}[t]
\centering
\includegraphics[width=0.95\linewidth]{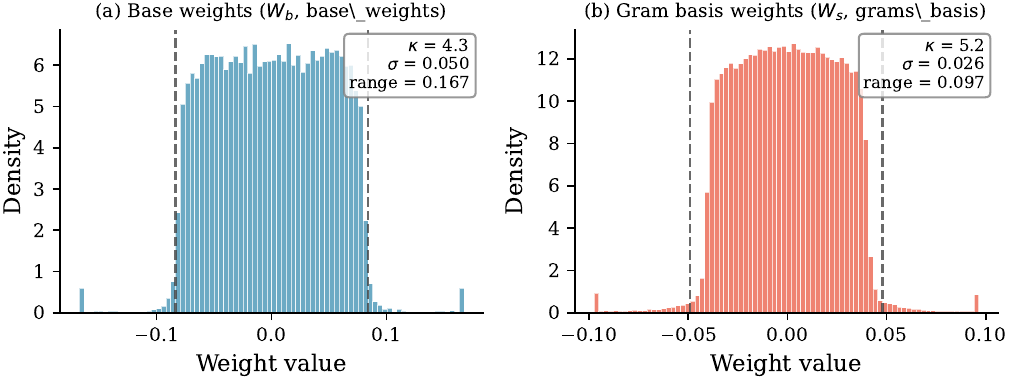}
\caption{Weight distributions in trained GRAM-KAN. \textbf{(a)}~Base weights show near-uniform
distribution with low kurtosis ($\kappa = 4.3$). \textbf{(b)}~Gram basis weights have moderate
kurtosis ($\kappa = 5.2$). Despite similar scales, basis weights are highly sensitive to quantization.}
\label{fig:gram_weights}
\end{figure}

Figure~\ref{fig:gram_weights} shows the most uniform distributions among the variants.
Base weights have the lowest kurtosis ($\kappa = 4.3$), only slightly above Gaussian, with
a notably flat-topped shape. Gram basis weights show moderate kurtosis ($\kappa = 5.2$) with
a more peaked distribution. Both branches have relatively similar scales.

\textit{Quantization implications:}
\begin{itemize}
    \item Lower kurtosis in both branches reduces outlier sensitivity compared to other variants
    \item \texttt{base\_weights} show the highest sensitivity under LSQ on CIFAR-10 (mean $\Delta = 0.99$, max $\Delta = 2.92$), narrowly above \texttt{grams\_basis\_weights} (mean $\Delta = 0.93$, max $\Delta = 2.46$); the two are within rounding
    \item Mixed-precision policy: retain 8-bit for \texttt{base\_weights}, quantize \texttt{grams\_basis\_weights} to 4-bit, \texttt{beta\_weights} to 2-bit
    \item GRAM-KAN benefits most from DSQ, which achieves near-zero degradation across all parameters
\end{itemize}

\subsection{Summary: Distribution vs.\ Sensitivity}
\label{app:dist_vs_sens}

Distributional width alone does not reliably predict quantization sensitivity.
EfficientKAN and GRAM show that the most sensitive group can also be the
widest, while in PyKAN and FastKAN the relationship is inverted.

This finding has important practical implications:

\begin{enumerate}
    \item \textbf{Do not rely solely on distributional statistics} for calibration decisions. Sensitivity analysis (as in Section~5.4) provides more actionable guidance.
    \item \textbf{The most sensitive parameter group is architecture-specific}: spline/basis in FastKAN, base in EfficientKAN and GRAM, scaling in PyKAN. Empirical sensitivity analysis is required per variant.
    \item \textbf{Smooth quantization methods (DSQ) are more robust} across all variants and parameter groups, achieving near-zero degradation where step-size methods (LSQ) show significant sensitivity variation.
\end{enumerate}

\subsection{Detailed Statistics}
\label{app:detailed_stats}

For completeness, Table~\ref{tab:weight_stats_detailed} provides extended statistics including
skewness, interquartile range, and parameter counts.

\begin{table}[!ht]
\centering
\caption{Extended weight statistics across KAN variants.}
\label{tab:weight_stats_detailed}
\small
\begin{tabular}{@{}llcccccc@{}}
\toprule
\textbf{Variant} & \textbf{Branch} & \textbf{Count} & $\mu$ & $\sigma$ & \textbf{Skew} & $\kappa$ & \textbf{IQR} \\
\midrule
\multirow{2}{*}{PyKAN}
  & Base   & 50,816  & $-$0.001 & 0.047 & $-$0.48 & 36.0 & 0.054 \\
  & Spline & 457,344 & 0.004    & 0.041 & $-$0.19 & 2.7  & 0.058 \\
\midrule
\multirow{2}{*}{EfficientKAN}
  & Base   & 50,816  & $-$0.002 & 0.047 & $-$0.37 & 37.0 & 0.053 \\
  & Spline & 457,344 & 0.000    & 0.011 & 0.03    & 8.2  & 0.013 \\
\midrule
\multirow{2}{*}{FastKAN}
  & Base   & 50,890  & 0.000    & 0.038 & $-$0.31 & 20.6 & 0.045 \\
  & Spline & 406,528 & 0.000    & 0.096 & 0.00    & 3.0  & 0.130 \\
\midrule
\multirow{2}{*}{GRAM-KAN}
  & Base   & 50,816  & 0.000    & 0.050 & 0.08    & 4.3  & 0.086 \\
  & Spline & 203,272 & 0.000    & 0.026 & $-$0.01 & 5.2  & 0.042 \\
\bottomrule
\end{tabular}
\end{table}

\subsection{Calibration Recommendations}
\label{app:calibration_reco}

Based on both distributional analysis and sensitivity results, Table~\ref{tab:quant_recommendations}
provides architecture-specific calibration and precision recommendations.

\begin{table}[!ht]
\centering
\caption{Recommended quantization strategies by KAN variant, informed by both distributional
analysis (this appendix) and parameter sensitivity (Section~\ref{parameter_sensitivity}).}
\label{tab:quant_recommendations}
\small
\begin{tabular}{@{}lllc@{}}
\toprule
\textbf{Variant} & \textbf{Recommended QAT} & \textbf{Mixed-precision (base/spline)} & \textbf{Per-ch.\ weights?} \\
\midrule
PyKAN        & DSQ          & 2-bit / 8-bit & Safe \\
EfficientKAN & DSQ or LSQ   & 8-bit / 2-bit & Safe \\
FastKAN      & DSQ          & 2-bit / 8-bit & \textbf{Avoid} \\
GRAM-KAN     & DSQ          & 4-bit / 4-bit & Safe \\
\bottomrule
\end{tabular}
\end{table}

The key takeaway is that \textbf{DSQ is the safest default} across all variants, achieving robust performance without requiring careful per-variant tuning. When using LSQ or other step-size methods, practitioners should (1) prioritize precision for the empirically most sensitive group rather than assuming a universal spline-dominant rule, (2) avoid per-channel weight quantization for FastKAN, and (3) expect higher sensitivity on complex datasets (CIFAR-10/100) compared to simpler benchmarks (MNIST).

\clearpage


\section{Branch-Aware vs.\ Shared Quantization}
\label{app:branch_vs_shared}

A central design choice of \textsc{QuantKAN} is to maintain
\emph{independent} quantizers for the base and spline/basis branches of
each KAN layer (Eq.~\ref{eq:kan_forward} and Eq.~\ref{eq:kan_quantized}). To isolate the contribution
of this choice, we ablate branch-aware quantization against a
\emph{shared-quantizer} baseline in which a single step size is
calibrated on the pooled weights of all branches within a layer and
applied uniformly to each tensor. Both configurations use identical
LSQ-style symmetric, per-tensor initialization ($s_{0}=2\,\mathbb{E}|w|/\sqrt{q_{\max}}$);
weights are quantized post-hoc while activations remain in full
precision, so that the comparison reflects the weight quantizer
assignment alone. We evaluate two representative KAN variants,
EfficientKAN and PyKAN, on four classification benchmarks (MNIST,
CIFAR-10, sklearn digits, and MNIST-1D) at bit-widths
$b\in\{8,6,4,3,2\}$.

\paragraph{Accuracy.}
Table~\ref{tab:branch_vs_shared_acc} reports Top-1 accuracy under both
configurations. The gap widens monotonically as precision is reduced,
and is consistently in favor of branch-aware quantization. For PyKAN
the degradation of the shared baseline is severe already at moderate
precision: on MNIST the shared quantizer drops to $79.30\%$ at W3 and
$60.06\%$ at W2, while branch-aware quantization retains $92.25\%$ and
$83.92\%$, respectively, a gap of $+12.95$ and $+23.86$ points. The
pattern is reproduced on CIFAR-10 ($+12.31$ at W3, $+17.17$ at W2), on
MNIST-1D ($+11.40$ and $+16.60$), and on digits ($+4.17$ and $+15.00$).
EfficientKAN is less extreme because its three branches
(\texttt{base\_weight}, \texttt{spline\_weight}, \texttt{spline\_scaler})
differ in dynamic range by roughly $4\times$ rather than the
${\sim}11\times$ characteristic of PyKAN, but the branch-aware
configuration still dominates at the lowest precisions on CIFAR-10
($+2.71$ at W2) and on MNIST-1D ($+4.10$ at W2). In short, wherever
quantization error becomes consequential, branch-aware assignment is
strictly preferable.

\begin{table}[!ht]
\centering
\small
\caption{Weights-only PTQ accuracy (\%) of branch-aware vs.\ shared
quantization for PyKAN and EfficientKAN. Both configurations use
LSQ-style symmetric per-tensor initialization. $\Delta$ denotes
branch-aware $-$ shared (higher is better); positive values in bold.}
\label{tab:branch_vs_shared_acc}
\begin{tabular}{llcccc}
\toprule
Dataset & Variant & Bits & Branch & Shared & $\Delta$ \\
\midrule
\multirow{6}{*}{MNIST}
& PyKAN        & W4 & 92.78 & 89.39 & \textbf{+3.39} \\
& PyKAN        & W3 & 92.25 & 79.30 & \textbf{+12.95} \\
& PyKAN        & W2 & 83.92 & 60.06 & \textbf{+23.86} \\
& EfficientKAN & W3 & 95.79 & 95.38 & \textbf{+0.41} \\
\midrule
\multirow{5}{*}{CIFAR-10}
& PyKAN        & W6 & 44.48 & 38.15 & \textbf{+6.33} \\
& PyKAN        & W4 & 41.89 & 34.06 & \textbf{+7.83} \\
& PyKAN        & W3 & 33.52 & 21.21 & \textbf{+12.31} \\
& PyKAN        & W2 & 27.99 & 10.82 & \textbf{+17.17} \\
& EfficientKAN & W2 & 40.88 & 38.17 & \textbf{+2.71} \\
\midrule
\multirow{4}{*}{MNIST-1D}
& PyKAN        & W4 & 40.20 & 36.70 & \textbf{+3.50} \\
& PyKAN        & W3 & 39.30 & 27.90 & \textbf{+11.40} \\
& PyKAN        & W2 & 33.70 & 17.10 & \textbf{+16.60} \\
& EfficientKAN & W2 & 36.90 & 32.80 & \textbf{+4.10} \\
\midrule
\multirow{2}{*}{digits}
& PyKAN        & W3 & 93.89 & 89.72 & \textbf{+4.17} \\
& PyKAN        & W2 & 91.94 & 76.94 & \textbf{+15.00} \\
\bottomrule
\end{tabular}
\end{table}

\paragraph{Mechanism: per-branch quantization error.}
Accuracy alone does not reveal \emph{why} a shared quantizer fails.
To expose the mechanism we measure the per-branch normalized
quantization error
\[
\mathrm{NMSE}(w)\;=\;
\frac{\mathbb{E}[(w-\hat w)^{2}]}{\mathbb{E}[w^{2}]},
\]
under both configurations. Since a shared step size must accommodate
the widest branch in the layer, the narrower branches are forced onto
a coarser grid. For PyKAN this is especially pronounced on the
coefficient tensor $w_{\text{coef}}$: across the four datasets the
shared-quantizer NMSE on $w_{\text{coef}}$ exceeds the branch-aware
NMSE by a factor of $5\text{--}10\times$ at W4, and by an order of
magnitude at W3 (Table~\ref{tab:branch_vs_shared_nmse}). For example,
on CIFAR-10 at W4 the coefficient NMSE rises from $0.028$ under
branch-aware to $0.274$ under shared, indicating that over a quarter
of the coefficient energy is replaced by quantization noise; at W3
this inflates to $0.531$ versus $0.073$. EfficientKAN exhibits the
same phenomenon, concentrated on \texttt{base\_weight}: on CIFAR-10
the shared-quantizer NMSE for \texttt{base\_weight} is
$2\text{--}3\times$ larger than branch-aware across all bit-widths
($0.077$ vs.\ $0.029$ at W4; $0.170$ vs.\ $0.074$ at W3). The effect
is \emph{structural}: whichever branch has the largest dynamic range
attracts the shared step size, and every other branch absorbs the
resulting misfit. Branch-aware quantization eliminates this coupling
by fitting each $s_{\text{branch}}$ to its own distribution.

\begin{table}[t]
\centering
\small
\caption{Per-branch normalized quantization MSE (lower is better)
for the branches most penalized by shared quantization. Shared vs.\
branch ratios quantify how severely each branch is corrupted when a
single step size is shared across the layer.}
\label{tab:branch_vs_shared_nmse}
\begin{tabular}{llccccc}
\toprule
Dataset & Variant & Branch & Bits
 & Branch NMSE & Shared NMSE & Ratio \\
\midrule
MNIST   & PyKAN & $w_{\text{coef}}$ & W4 & 0.027 & 0.138 & 5.2$\times$ \\
MNIST   & PyKAN & $w_{\text{coef}}$ & W3 & 0.073 & 0.283 & 3.9$\times$ \\
CIFAR-10 & PyKAN & $w_{\text{coef}}$ & W4 & 0.028 & 0.274 & 9.6$\times$ \\
CIFAR-10 & PyKAN & $w_{\text{coef}}$ & W3 & 0.073 & 0.531 & 7.3$\times$ \\
MNIST-1D & PyKAN & $w_{\text{coef}}$ & W4 & 0.030 & 0.239 & 8.0$\times$ \\
MNIST-1D & PyKAN & $w_{\text{coef}}$ & W3 & 0.072 & 0.499 & 7.0$\times$ \\
digits  & PyKAN & $w_{\text{coef}}$ & W4 & 0.032 & 0.311 & 9.6$\times$ \\
CIFAR-10 & EfficientKAN & $w_{\text{base}}$ & W4 & 0.029 & 0.077 & 2.6$\times$ \\
CIFAR-10 & EfficientKAN & $w_{\text{base}}$ & W3 & 0.074 & 0.170 & 2.3$\times$ \\
CIFAR-10 & EfficientKAN & $w_{\text{base}}$ & W2 & 0.244 & 0.458 & 1.9$\times$ \\
\bottomrule
\end{tabular}
\end{table}

These observations are consistent with the distributional analysis of
Section~\ref{sec:framework} and Appendix~\ref{app:weight_distributions}:
KAN branches are not merely stylistically distinct but exhibit
statistical heterogeneity large enough that a single scaling parameter
cannot simultaneously serve all of them without inducing a
disproportionate error on the narrower branch. Because the
spline/basis pathway is also the one most sensitive to quantization
error (Section~\ref{sec:appendix_sensitivity}), this misallocation of
precision is directly responsible for the accuracy collapse observed
in the shared-quantizer baseline at $b\le 4$. Branch-aware
quantization is therefore not an incremental refinement but a
structural prerequisite for reliable low-bit KAN inference, and its
benefit grows as the precision budget shrinks---precisely the regime
in which quantization is most economically motivated.


\clearpage
\section{Quantization-Aware Training (QAT) Quantizers}
\label{app:qat_quantizers}

This appendix provides a concise and implementation-aligned description of the QAT quantizers used in our framework: LSQ, LSQ+, DoReFa, DSQ, QIL, and PACT. We present each method with (i) core equations, (ii) pseudocode that matches our implementation choices, and (iii) brief guidance on when the method is especially effective for KAN/KAGN-style models.

\subsection{Preliminaries: Fake-Quantization and STE Primitives}
\label{app:qat_prelim}

We adopt the standard \emph{fake-quantization} view: in the forward pass, tensors are quantized to an integer grid and dequantized back to floating point; in the backward pass, we use Straight-Through Estimators (STE) to preserve gradient flow through discrete operations.

Let bit-width be $b$, integer range be $(q_{\min}, q_{\max})$, and step size (scale) be $s>0$. For signed quantization we commonly use $q_{\max}=2^{b-1}-1$; for unsigned we use $q_{\max}=2^b-1$ with $q_{\min}=0$. We use $\mathrm{clip}(x,a,b)$ for clamping.

Our implementation uses compile-friendly STE helpers:
\begin{align}
\mathrm{round\_ste}(x) &= (\lfloor x \rceil - x)_{\text{stopgrad}} + x, \\
\mathrm{sign\_ste}(x)  &= (\mathrm{sign}(x) - x)_{\text{stopgrad}} + x,
\end{align}
and for LSQ-style step-size stabilization, a gradient scaling helper:
\begin{equation}
\mathrm{grad\_scale}(x,g) = (x - xg)_{\text{stopgrad}} + xg.
\end{equation}

KAN/KAGN layers often combine linear mixing with nonlinear function blocks (e.g., basis expansions). This can produce heterogeneous per-channel statistics, skewed intermediate distributions, and occasional outliers. The quantizers below are chosen to address these failure modes via (i) learned step sizes (LSQ), (ii) learned offsets (LSQ+), (iii) bounded transforms (DoReFa), (iv) smoother discretization (DSQ), (v) learned prune/clip intervals (QIL), and (vi) learned activation clipping (PACT).

\subsection{LSQ: Learned Step Size Quantization}
\label{app:lsq}

LSQ learns the step size $s$ by optimizing task loss through a uniform quantizer \citep{esser2019lsq}. For a tensor $x$ (weights or activations),
\begin{equation}
x_q = \mathrm{clip}\!\left(\left\lfloor \frac{x}{s} \right\rceil, q_{\min}, q_{\max}\right),
\qquad
\hat{x} = x_q \cdot s.
\label{eq:lsq_forward}
\end{equation}
The rounding uses STE, and the gradient of $s$ is stabilized via \texttt{grad\_scale}, with a scale factor typically proportional to $1/\sqrt{N\cdot q_{\max}}$ (per-tensor or per-channel), as in LSQ.

Learning $s$ using task loss tends to outperform fixed calibration at low bit-widths, and per-channel $s$ can accommodate heterogeneous channel scales.

\begin{algorithm}[!ht]
\caption{LSQ fake-quantization (implementation-aligned)}
\label{alg:lsq}
\begin{algorithmic}[1]
\REQUIRE Tensor $x$, bits $b$, flags: \texttt{all\_positive}, \texttt{symmetric}, \texttt{per\_channel}, axis \texttt{ch\_axis}
\STATE Set $(q_{\min},q_{\max})$ from flags (unsigned or signed; strict symmetric optional).
\STATE Initialize learnable step $s$ (per-tensor or per-channel), e.g. $s_0 \propto \mathbb{E}|x|/\sqrt{q_{\max}}$.
\STATE Compute gradient scale $g \leftarrow 1/\sqrt{N_{\text{per-scale}}\,q_{\max}}$.
\STATE $s \leftarrow \mathrm{grad\_scale}(\max(s,\varepsilon), g)$
\STATE $u \leftarrow x/s$
\STATE $u \leftarrow \mathrm{clip}(u, q_{\min}, q_{\max})$
\STATE $x_q \leftarrow \mathrm{round\_ste}(u)$
\RETURN $\hat{x} \leftarrow x_q \cdot s$
\end{algorithmic}
\end{algorithm}

LSQ is a strong default for weight and activation QAT when uniform hardware-friendly levels are desired but ranges vary across layers/channels (common in KAN coefficient tensors).

\subsection{LSQ+: Learned Step Size and Offset (Asymmetric Quantization)}
\label{app:lsqplus}

LSQ+ extends LSQ by learning an offset (zero-point-like) term to better fit skewed distributions \citep{bhalgat2020lsq+}. In its general form,
\begin{equation}
x_q = \mathrm{clip}\!\left(\left\lfloor \frac{x-\beta}{s} \right\rceil, n, p\right),
\qquad
\hat{x} = x_q\cdot s + \beta.
\label{eq:lsqplus_general}
\end{equation}
In our implementation, we store a learnable offset $z$ in the integer domain and use an affine form in scaled coordinates. A practical detail in our code is to avoid trivial gradient cancellation through the de-offset step by detaching the dequant offset term in the final reconstruction (this preserves stable learning of $z$ in practice).

\begin{algorithm}[!ht]
\caption{LSQ+ fake-quantization (implementation-aligned)}
\label{alg:lsqplus}
\begin{algorithmic}[1]
\REQUIRE Tensor $x$, bits $b$, learnable $(s,z)$, integer range $(q_{\min},q_{\max})$
\STATE $s \leftarrow \mathrm{grad\_scale}(\max(s,\varepsilon), g_s)$;\ \ $z \leftarrow \mathrm{grad\_scale}(z, g_z)$
\STATE $z_{\text{eff}} \leftarrow \mathrm{clip}(z, q_{\min}, q_{\max})$ \COMMENT{forward clamp}
\STATE $q_{\text{pre}} \leftarrow x/s + z_{\text{eff}}$
\STATE $q_{\text{int}} \leftarrow \mathrm{round\_ste}(\mathrm{clip}(q_{\text{pre}}, q_{\min}, q_{\max}))$
\STATE $z_{\text{ng}} \leftarrow \mathrm{stopgrad}(z_{\text{eff}})$ \COMMENT{prevents gradient cancellation}
\RETURN $\hat{x} \leftarrow (q_{\text{int}} - z_{\text{ng}})\cdot s$
\end{algorithmic}
\end{algorithm}

LSQ+ is especially helpful for activations that are skewed or shifted. This often arises in nonlinear function blocks where the output distribution is not centered around zero.

\subsection{DoReFa: Bounded Transforms + Uniform Discretization}
\label{app:dorefa}

DoReFa-Net proposes low-bit quantization via simple bounded transforms followed by uniform discretization \citep{zhou2016dorefa}. Our implementation is QAT-ready and back-compatible, supporting: (i) $b=32$ identity, (ii) special $b=1$ behavior, (iii) per-channel weight stats, and (iv) optional stochastic gradient quantization hooks.

\paragraph{Activation quantization.}
We assume/clamp activations to $[0,1]$. For $b\ge 2$,
\begin{equation}
\hat{a}=\mathrm{clip}\!\left(\frac{\mathrm{round\_ste}(a\cdot n)}{n},\,0,\,1\right),
\qquad n=2^b-1,
\end{equation}
and for $b=1$ we use a $0.5$ threshold with STE.

\paragraph{Weight quantization (our exact behavior).}
Let $w_t=\tanh(w)$.
\begin{itemize}
\item \textbf{$b=1$}: binary weights with learned magnitude $E=\mathbb{E}|w_t|$ (per-channel or per-tensor, detached),
\[
\hat{w}=\mathrm{sign\_ste}(w_t)\cdot E.
\]
\item \textbf{$b\ge 2$}: compute $m=\max|w_t|$ (per-channel or per-tensor, detached), map to $[0,1]$ via
$
w_{01}=\mathrm{clip}(w_t/(2m)+1/2,0,1),
$
quantize $w_{01}$ uniformly, then map back:
\[
\hat{w}=(2\hat{w}_{01}-1)\cdot m.
\]
\end{itemize}

\begin{algorithm}[!ht]
\caption{DoReFa fake-quantization (matches provided implementation)}
\label{alg:dorefa}
\begin{algorithmic}[1]
\REQUIRE Tensor $x$, bits $b$, mode $\in\{\textbf{weight},\textbf{activation}\}$, flags: \texttt{per\_channel}, \texttt{ch\_axis}
\IF{$b=32$} \RETURN $x$ \ENDIF
\IF{mode=\textsf{weight}}
  \STATE $x_t \leftarrow \tanh(x)$
  \IF{$b=1$}
    \STATE $E \leftarrow \mathbb{E}(|x_t|)$ (detached; per-channel if enabled)
    \RETURN $\mathrm{sign\_ste}(x_t)\cdot E$
  \ELSE
    \STATE $m \leftarrow \max(|x_t|)$ (detached; per-channel if enabled)
    \STATE $w_{01} \leftarrow \mathrm{clip}(x_t/(2m)+0.5,\ 0,\ 1)$
    \STATE $n \leftarrow 2^b-1$
    \STATE $\hat{w}_{01} \leftarrow \mathrm{clip}(\mathrm{round\_ste}(w_{01}\cdot n)/n,\ 0,\ 1)$
    \RETURN $\hat{x} \leftarrow (2\hat{w}_{01}-1)\cdot m$
  \ENDIF
\ELSE
  \STATE $a_{01} \leftarrow \mathrm{clip}(x,\ 0,\ 1)$
  \IF{$b=1$}
    \STATE $y \leftarrow \mathbb{I}[a_{01}\ge 0.5]$
    \RETURN $(y-a_{01})_{\text{stopgrad}} + a_{01}$
  \ELSE
    \STATE $n \leftarrow 2^b-1$
    \RETURN $\mathrm{clip}(\mathrm{round\_ste}(a_{01}\cdot n)/n,\ 0,\ 1)$
  \ENDIF
\ENDIF
\end{algorithmic}
\end{algorithm}

DoReFa can be a strong stability anchor at low bit-width because $\tanh(\cdot)$ bounds weights and scaling uses detached per-channel statistics. This can help prevent runaway scales in deep stacks.

\subsection{DSQ: Differentiable Soft Quantization}
\label{app:dsq}

DSQ replaces hard rounding with a differentiable soft staircase controlled by a sharpness parameter, bridging full precision and low-bit training \citep{gong2019dsq}. The sharpness is typically annealed so that soft rounding becomes closer to hard rounding later in training.

\begin{algorithm}[!ht]
\caption{DSQ fake-quantization (implementation-aligned structure)}
\label{alg:dsq}
\begin{algorithmic}[1]
\REQUIRE Tensor $x$, bits $b$, bounds (unsigned or symmetric), sharpness $\alpha$
\STATE Clamp $x$ to a bounded interval (unsigned or symmetric depending on mode)
\STATE Map to quant units: $t \leftarrow (x-\ell)/s$ (unsigned) or $t\leftarrow x/s$ (symmetric)
\STATE Soft-round: $t \leftarrow \mathrm{SoftRound}(t;\alpha)$
\STATE Clamp $t$ to $[q_{\min},q_{\max}]$
\RETURN Dequantize: $\hat{x}\leftarrow t\cdot s + \ell$ (unsigned) or $\hat{x}\leftarrow t\cdot s$ (symmetric)
\end{algorithmic}
\end{algorithm}

DSQ is useful when hard rounding + STE causes brittle optimization at very low bit-width (e.g., 2--3 bits), particularly when quantization noise is amplified by nonlinear function blocks.

\subsection{QIL: Quantization Interval Learning (Pruning + Clipping + Discretization)}
\label{app:qil}

QIL learns a quantization interval and reshapes values into a bounded range before discretization \citep{jung2019qil}. It can explicitly \emph{prune} small magnitudes (map to $0$) and \emph{clip} large magnitudes (map to $\pm 1$), while learning an internal mapping controlled by $(c,d,\gamma)$:
\begin{equation}
v(w)=
\begin{cases}
0, & |w| < c-d,\\
\mathrm{sign}(w), & |w| > c+d,\\
(\mathrm{clip}(\alpha |w|+\beta,0,1))^{\gamma}\,\mathrm{sign}(w), & \text{otherwise},
\end{cases}
\quad
\alpha=\frac{1}{2d},\ \beta=-\frac{c}{2d}+\frac{1}{2}.
\end{equation}
We then discretize on a uniform grid. Our implementation uses a safe minimum of one positive level for weights at very low bits, i.e., $q=\max(2^{b-1}-1,1)$.

\begin{algorithm}[!ht]
\caption{QIL fake-quantization (implementation-aligned)}
\label{alg:qil}
\begin{algorithmic}[1]
\REQUIRE Tensor $x$, bits $b$, mode $\in\{\textsf{weight},\textsf{activation}\}$, learned $(c,d)$, optional $\gamma$
\STATE Parameterize $d>0$ and $c\ge d$ (e.g., softplus); compute $\alpha\leftarrow 1/(2d)$, $\beta\leftarrow -c/(2d)+1/2$
\IF{mode=\textsf{weight}}
  \STATE $a\leftarrow |x|$, $s\leftarrow \mathrm{sign}(x)$
  \STATE $m \leftarrow \mathrm{clip}(\alpha a+\beta,0,1)$; optionally $m\leftarrow m^\gamma$
  \STATE $v \leftarrow 0$ if $a<c-d$; $v\leftarrow s$ if $a>c+d$; else $v\leftarrow m\cdot s$
  \STATE $q\leftarrow \max(2^{b-1}-1,1)$
  \RETURN $\hat{x}\leftarrow (\mathrm{round\_ste}(|v|q)/q)\cdot \mathrm{sign}(v)$
\ELSE
  \STATE $u \leftarrow 0$ if $x<c-d$; $u\leftarrow 1$ if $x>c+d$; else $u\leftarrow \mathrm{clip}(\alpha x+\beta,0,1)$
  \STATE $q\leftarrow 2^b-1$
  \RETURN $\hat{x}\leftarrow \mathrm{round\_ste}(u q)/q$
\ENDIF
\end{algorithmic}
\end{algorithm}

QIL can be effective when tensors have many small values plus a few large outliers, because the learned interval can prune noise and clip extremes. However, at very low bits, it can be higher-variance than LSQ/PACT unless interval learning is well regularized.

\subsection{PACT: Parameterized Clipping Activation}
\label{app:pact}
PACT learns a clipping threshold $\alpha$ to reduce activation range and improve effective resolution \citep{choi2018pact}. After clipping, it applies uniform quantization:
\begin{equation}
x_c=\mathrm{clip}(x,0,\alpha),\quad
\Delta=\alpha/(2^b-1),\quad
\hat{x}=\Delta\cdot \mathrm{clip}\!\left(\left\lfloor x_c/\Delta\right\rceil,0,2^b-1\right).
\end{equation}
Our implementation also supports a symmetric extension $x_c=\mathrm{clip}(x,-\alpha,\alpha)$ for signed tensors, with $\Delta=\alpha/(2^{b-1}-1)$.

\begin{algorithm}[!ht]
\caption{PACT fake-quantization (implementation-aligned)}
\label{alg:pact}
\begin{algorithmic}[1]
\REQUIRE Tensor $x$, bits $b$, learned clip $\alpha$, flag \texttt{symmetric}
\STATE Ensure $\alpha>0$ (e.g., softplus + $\varepsilon$ or clamp)
\IF{symmetric}
  \STATE $q_{\max}\leftarrow 2^{b-1}-1$, $q_{\min}\leftarrow -2^{b-1}$ (or strict variant)
  \STATE $x_c\leftarrow \mathrm{clip}(x,-\alpha,\alpha)$; $\Delta\leftarrow \alpha/q_{\max}$
\ELSE
  \STATE $q_{\max}\leftarrow 2^b-1$, $q_{\min}\leftarrow 0$
  \STATE $x_c\leftarrow \mathrm{clip}(x,0,\alpha)$; $\Delta\leftarrow \alpha/q_{\max}$
\ENDIF
\STATE $q \leftarrow \mathrm{round\_ste}(x_c/\Delta)$
\STATE $q \leftarrow \mathrm{clip}(q,q_{\min},q_{\max})$
\RETURN $\hat{x}\leftarrow q\cdot \Delta$
\end{algorithmic}
\end{algorithm}

PACT is a strong choice for activation quantization when rare outliers otherwise force a large range and poor resolution. Learning $\alpha$ shrinks the range and often stabilizes low-bit activation quantization in nonlinear blocks.

\subsection{Summary: Which Quantizer When (Practical, KAN/KAGN-Oriented)}
\label{app:qat_summary}

A practical, implementation-consistent rule-set for KAN/KAGN is:
\begin{itemize}
\item \textbf{Default}: LSQ for weights (per-channel, symmetric), LSQ or LSQ+ for activations depending on skew.
\item \textbf{Skewed activations}: prefer LSQ+ (learned offset) to better use available levels.
\item \textbf{Outlier-dominated activations}: prefer PACT (learned clipping).
\item \textbf{Unstable low-bit training (2--3 bits)}: consider DSQ for smoother optimization.
\item \textbf{Heavy-tailed / many small coefficients}: consider QIL (learned interval with pruning+clipping), but expect higher variance without regularization.
\item \textbf{Stable, minimal baseline}: DoReFa (bounded $\tanh$ weights + detached per-channel scaling) can remain stable when learned-parameter quantizers are fragile.
\end{itemize}

\clearpage
\section{Post-Training Quantization (PTQ) Methods}
\label{app:ptq}

This appendix presents the post-training quantization (PTQ) methods used in this
work. Each method is described using (i) a mathematical formulation that characterizes
the quantization objective, and (ii) an implementation-aligned pseudo-code block.
Unlike quantization-aware training (QAT), PTQ methods do not update model
parameters via backpropagation; instead, they rely on calibration data and
analytical or optimization-based criteria to reduce quantization error.

\subsection{Preliminaries}
\label{app:ptq_prelim}

Let $W \in \mathbb{R}^{d_{\text{out}}\times d_{\text{in}}}$ denote a floating-point
weight tensor and $X \in \mathbb{R}^{d_{\text{in}}\times n}$ denote calibration inputs.
Uniform affine quantization maps $W$ to
\begin{equation}
\hat{W} = s \cdot \mathrm{clip}\!\left(
\left\lfloor \frac{W}{s} \right\rceil,
q_{\min}, q_{\max}
\right),
\label{eq:uniform_ptq}
\end{equation}
where $s$ is a scale and $[q_{\min}, q_{\max}]$ defines the integer range.
All PTQ methods discussed below modify either the rounding operator,
the scaling strategy, or the objective used to select $\hat{W}$.

KAN models introduce additional challenges for PTQ due to heterogeneous parameter
groups (base weights, spline coefficients, polynomial kernels) and non-uniform
activation distributions, motivating structure-aware and sensitivity-aware PTQ.

\subsection{Uniform PTQ}
\label{app:ptq_uniform}

Uniform PTQ applies Eq.~\eqref{eq:uniform_ptq} directly using min--max calibrated
scales. Each weight element is quantized independently, without considering the
effect of quantization error on the layer output. For symmetric quantization,
the scale is computed as
\begin{equation}
s = \frac{\max |W|}{2^{b-1}-1}.
\end{equation}

\begin{algorithm}[!ht]
\caption{Uniform PTQ (implementation-aligned)}
\label{alg:ptq_uniform}
\begin{algorithmic}[1]
\REQUIRE Weight tensor $W$, bits $b$, flag \texttt{symmetric}
\IF{symmetric}
  \STATE $q_{\max}\leftarrow 2^{b-1}-1$, $q_{\min}\leftarrow -2^{b-1}$
  \STATE $s \leftarrow \max |W| / q_{\max}$
\ELSE
  \STATE $q_{\max}\leftarrow 2^b-1$, $q_{\min}\leftarrow 0$
  \STATE $s \leftarrow (\max W - \min W) / q_{\max}$
\ENDIF
\STATE $q \leftarrow \mathrm{round}(W / s)$
\STATE $q \leftarrow \mathrm{clip}(q, q_{\min}, q_{\max})$
\RETURN $\hat{W} \leftarrow q \cdot s$
\end{algorithmic}
\end{algorithm}

Uniform PTQ is computationally efficient and stable for KAN base weights, but it
often produces large functional errors for spline or polynomial parameters with
heavy-tailed distributions.

\subsection{AdaRound}
\label{app:ptq_adaround}

AdaRound improves upon uniform rounding by optimizing rounding decisions using
calibration data. Instead of immediately applying $\lfloor\cdot\rceil$,
AdaRound introduces learnable rounding offsets $\alpha$ and defines
\begin{equation}
\hat{W} = s \cdot \left(
\left\lfloor \frac{W}{s} \right\rfloor + \sigma(\alpha)
\right),
\end{equation}
where $\sigma(\cdot)$ is a sigmoid function.
The offsets $\alpha$ are optimized to minimize output reconstruction error
\begin{equation}
\min_{\alpha} \; \|WX - \hat{W}X\|_2^2 .
\end{equation}

\begin{algorithm}[!ht]
\caption{AdaRound PTQ (layer-wise reconstruction)}
\label{alg:ptq_adaround}
\begin{algorithmic}[1]
\REQUIRE Weight tensor $W$, calibration inputs $X$, bits $b$, iterations $T$
\STATE Initialize scale $s$ and rounding parameters $\alpha$
\FOR{$t = 1$ to $T$}
  \STATE $\hat{W} \leftarrow s \cdot \left(\lfloor W/s \rfloor + \sigma(\alpha)\right)$
  \STATE Compute $\|WX - \hat{W}X\|_2^2$
  \STATE Update $\alpha$
\ENDFOR
\STATE Finalize rounding decisions
\RETURN $\hat{W}$
\end{algorithmic}
\end{algorithm}

AdaRound is particularly effective for spline coefficients and polynomial kernels
in KANs, where naive rounding can distort learned functional relationships.

\subsection{GPTQ and GPTQ-Strict}
\label{app:ptq_gptq}

GPTQ minimizes a second-order approximation of quantization error by incorporating
Hessian information estimated from calibration data. The quantization objective is
\begin{equation}
\mathcal{L}_{\text{GPTQ}} =
(W - \hat{W})^\top H (W - \hat{W}),
\end{equation}
where $H$ is a Hessian or Hessian proxy of the loss with respect to $W$.
Weights are quantized sequentially, and the error introduced by early decisions
is compensated by updating remaining weights.

\begin{algorithm}[!ht]
\caption{GPTQ PTQ (block-wise Hessian-aware)}
\label{alg:ptq_gptq}
\begin{algorithmic}[1]
\REQUIRE Weight matrix $W$, calibration inputs $X$, block size $B$, damping $\lambda$
\STATE Estimate Hessian proxy $H \approx 2XX^\top + \lambda I$
\STATE Compute Cholesky factorization of $H^{-1}$
\FOR{each block of columns}
  \FOR{each column $j$ in block}
    \STATE Quantize $W_{:,j}$ to $\hat{W}_{:,j}$
    \STATE Propagate error using $H^{-1}$
  \ENDFOR
\ENDFOR
\RETURN $\hat{W}$
\end{algorithmic}
\end{algorithm}

The GPTQ-Strict variant enforces a fixed triangular update order, improving
numerical stability and determinism. GPTQ is especially effective for large
KAN mixing matrices and spline tensors.

\subsection{BRECQ}
\label{app:ptq_brecq}

BRECQ reconstructs quantized blocks by minimizing output reconstruction error on
calibration data. For a block with inputs $X^{(i)}$ and outputs $Y^{(i)}$, BRECQ
optimizes
\begin{equation}
\min_{\hat{W}^{(i)}} \; \|Y^{(i)} - \hat{W}^{(i)} X^{(i)}\|_2^2 .
\end{equation}

\begin{algorithm}[!ht]
\caption{BRECQ PTQ (block reconstruction)}
\label{alg:ptq_brecq}
\begin{algorithmic}[1]
\REQUIRE Block weights $W^{(i)}$, cached inputs $X^{(i)}$, outputs $Y^{(i)}$, iterations $T$
\FOR{$t = 1$ to $T$}
  \STATE Quantize block weights $\hat{W}^{(i)}$
  \STATE Compute $\|Y^{(i)} - \hat{W}^{(i)} X^{(i)}\|_2^2$
  \STATE Update quantization parameters
\ENDFOR
\RETURN $\hat{W}^{(i)}$
\end{algorithmic}
\end{algorithm}

BRECQ is well-suited for structured KAN blocks when targeting extremely low
precision (e.g., INT2/INT3).

\subsection{AWQ}
\label{app:ptq_awq}

AWQ reduces quantization error by rescaling weights using activation statistics.
Given calibration activations $A$, AWQ computes a per-channel scaling factor
\begin{equation}
\gamma = \mathbb{E}[|A|],
\end{equation}
and rescales weights as $\tilde{W} = W / \gamma$ prior to quantization.

\begin{algorithm}[!ht]
\caption{AWQ PTQ (activation-aware scaling)}
\label{alg:ptq_awq}
\begin{algorithmic}[1]
\REQUIRE Weights $W$, calibration activations $A$, bits $b$
\STATE $\gamma \leftarrow \mathbb{E}[|A|]$
\STATE $\tilde{W} \leftarrow W / \gamma$
\STATE Apply uniform PTQ to $\tilde{W}$
\RETURN $\hat{W}$
\end{algorithmic}
\end{algorithm}

AWQ is particularly effective for spline-based KAN layers, where a small number of
channels dominate activation magnitude.

\subsection{HAWQ-V2}
\label{app:ptq_hawq}

HAWQ-V2 assigns mixed precision using layer-wise sensitivity estimates based on the
Hessian trace. The trace is approximated using Hutchinson’s estimator:
\begin{equation}
\mathrm{Tr}(H) \approx \mathbb{E}_{r}[r^\top H r],
\end{equation}
where $r$ is a random Rademacher vector.

\begin{algorithm}[!ht]
\caption{HAWQ-V2 mixed-precision PTQ}
\label{alg:ptq_hawq}
\begin{algorithmic}[1]
\REQUIRE Model layers $\{\ell\}$, calibration data
\FOR{each layer $\ell$}
  \STATE Estimate $\mathrm{Tr}(H_\ell)$
\ENDFOR
\STATE Allocate bit-widths based on sensitivity
\RETURN Mixed-precision configuration
\end{algorithmic}
\end{algorithm}

KAN layers exhibit heterogeneous sensitivity due to nonlinear basis functions,
making Hessian-aware bit allocation particularly effective.

\subsection{SmoothQuant}
\label{app:ptq_smoothquant}

SmoothQuant reduces activation outliers by redistributing scale between weights
and activations. Given a smoothing factor $\alpha$, it applies
\begin{equation}
W' = W \cdot \alpha, \qquad A' = A / \alpha,
\end{equation}
followed by quantization of $W'$ and $A'$.

\begin{algorithm}[!ht]
\caption{SmoothQuant PTQ (scale redistribution)}
\label{alg:ptq_smoothquant}
\begin{algorithmic}[1]
\REQUIRE Weights $W$, activations $A$, smoothing factor $\alpha$
\STATE $W' \leftarrow W \cdot \alpha$
\STATE $A' \leftarrow A / \alpha$
\STATE Quantize $W'$ and $A'$
\RETURN Quantized model
\end{algorithmic}
\end{algorithm}

This approach is especially beneficial when spline activations exhibit heavy-tailed
distributions.

\subsection{ZeroQ}
\label{app:ptq_zeroq}

ZeroQ enables data-free PTQ by synthesizing calibration inputs that match internal
statistics. Synthetic inputs $X_{\text{syn}}$ are optimized to minimize
\begin{equation}
\mathcal{L}_{\text{syn}} =
\|\mu(X_{\text{syn}}) - \mu_{\text{ref}}\|_2^2 +
\|\sigma(X_{\text{syn}}) - \sigma_{\text{ref}}\|_2^2 .
\end{equation}

\begin{algorithm}[!ht]
\caption{ZeroQ data-free PTQ}
\label{alg:ptq_zeroq}
\begin{algorithmic}[1]
\REQUIRE Model, reference statistics
\STATE Initialize synthetic inputs
\STATE Optimize inputs to match statistics
\STATE Apply PTQ using synthetic data
\RETURN Quantized model
\end{algorithmic}
\end{algorithm}

\clearpage

\section{Experimental Hyperparameters}
\label{app:hyperparams}

This appendix summarizes the hyperparameters used across all experiments.
All runs are configuration-driven, and unless otherwise stated, hyperparameters are
held fixed across quantization methods to ensure fair comparison.

\subsection{Datasets and Preprocessing}

\paragraph{MNIST.}
Images are normalized using mean $0.1307$ and standard deviation $0.3081$.
During training, we apply random rotation (up to $10^\circ$); evaluation uses
deterministic normalization only. Input resolution is $28\times28$ (single channel).

\paragraph{CIFAR-10 and CIFAR-100.}
We use standard data augmentation including random horizontal flip, AutoAugment
(CIFAR10 / ImageNet / SVHN policies), and TrivialAugmentWide.
Inputs are normalized to mean $0.5$ and standard deviation $0.5$ per channel.
Evaluation uses deterministic normalization without augmentation.

\paragraph{TinyImageNet.}
Images are resized and cropped to $64\times64$ resolution.
Training uses random crop with padding $4$, random horizontal flip, and
AutoAugment (ImageNet policy). Evaluation uses resize followed by center crop.
ImageNet normalization statistics are applied.

\paragraph{ImageNet.}
Training uses random resized crop ($224\times224$), random horizontal flip,
and AutoAugment (ImageNet policy). Evaluation uses resize to $256$ followed by
center crop to $224$. Standard ImageNet normalization is applied.
Class indices are remapped to preserve the canonical $0$--$999$ label space.




\subsection{Model Architectures}
\label{sec:architectures}

We evaluate KAN backbones that appear in Table~\ref{tab:kan_qat}, spanning both fully connected and convolutional designs:
\begin{itemize}
  \item \textbf{Fully connected KAN}: \textbf{KAN FCN}, implemented with variant-specific spline layers (EfficientKAN-style KANLinear, FastKANLayer, GRAMLayer, and PyKANLayer) in a two-layer MLP for MNIST/CIFAR-scale inputs.
  \item \textbf{Convolutional KAN}: \textbf{KAN ConvNet}, a shallow CNN that replaces standard convolutions and classifiers with KAN convolution blocks (KAN\_Convolutional\_Layer) and KANLinear heads; we use dataset-specific instantiations (e.g., \texttt{KANConvNet\_MNIST}, \texttt{KANConvNet\_CIFAR10}).
  \item \textbf{Convolutional KAGN}: \textbf{KAGN Simple} and \textbf{KAGN Simple 8 Layers}, implemented as \texttt{SimpleConvKAGN} and \texttt{EightSimpleConvKAGN} with stacked \texttt{KAGNConv2DLayer} blocks followed by global pooling and a KAGN/linear classifier head. For CIFAR-10 we evaluate both the 4-block ``Simple'' configuration and the 8-layer configuration.
  \item \textbf{VGG-like KAGN}: \textbf{VGG Like KAGN V2}, implemented via \texttt{vggkagn} (VGG11v2-style configuration) with \texttt{KAGNConv2DLayer} features, adaptive pooling, and a linear classifier head. We use this VGG-like KAGN only for CIFAR-100 as reported in Table~\ref{tab:kan_qat}. These VGG-like KAGN models are adopted directly from ~\cite{drokin2024kolmogorov}.
\end{itemize}

Unless stated otherwise, the fully connected \textbf{KAN FCN} models use two spline-based layers (input $\rightarrow$ hidden $\rightarrow$ output), with hidden width set per dataset (e.g., 64 for MNIST and up to 256 for CIFAR-scale inputs).
The \textbf{KAN ConvNet} models are shallow backbones with KAN convolution blocks, max-pooling between stages, and a KANLinear classifier.
\textbf{KAGN Simple} uses four KAGN convolution blocks with downsampling and global pooling, while \textbf{KAGN Simple 8 Layers} deepens this design to eight KAGN convolution blocks.
The \textbf{VGG Like KAGN V2} backbone follows a VGG11-style stage pattern with KAGN convolution blocks and max-pooling, ending with adaptive pooling and a linear classifier.

\subsection{Optimization and Training}

\paragraph{Optimizer.}
All models are trained using Adam or AdamW with default $\beta$ parameters
($\beta_1 = 0.9$, $\beta_2 = 0.999$). Weight decay is set to $10^{-4}$ for
CIFAR-scale and larger datasets, and disabled for MNIST unless otherwise noted.

\paragraph{Learning rate.}
The base learning rate is selected per dataset scale:
\begin{itemize}
\item MNIST: $1\times10^{-3}$
\item CIFAR-10/100: $1\times10^{-3}$ to $3\times10^{-4}$
\item TinyImageNet / ImageNet: $3\times10^{-4}$
\end{itemize}
Learning rate schedules follow cosine decay or step decay as specified in the
configuration files.

\paragraph{Batch size.}
Batch size is dataset-dependent:
\begin{itemize}
\item MNIST: 128--256
\item CIFAR-10/100: 128
\item TinyImageNet: 128
\item ImageNet: 64
\end{itemize}

\paragraph{Epochs.}
Models are trained for:
\begin{itemize}
\item MNIST: 20--40 epochs
\item CIFAR-10/100: 200 epochs
\item TinyImageNet: 100 epochs
\item ImageNet: 90 epochs
\end{itemize}
Early stopping is applied only for selected MNIST runs.

\subsection{Quantization-Aware Training (QAT)}

\paragraph{Bit-widths.}
We evaluate weight-only and joint weight--activation quantization at
$\{8, 4, 3, 2\}$ bits.

\paragraph{Quantizer configuration.}
Unless otherwise stated:
\begin{itemize}
\item Base weights, spline/basis weights, and activations use \textbf{independent quantizers}
\item Activations default to asymmetric quantization
\item Weights default to symmetric quantization
\item Per-tensor scaling is used by default; per-channel scaling is evaluated in ablations
\end{itemize}

\paragraph{Method-specific settings.}
\begin{itemize}
\item \textbf{LSQ / LSQ+}: step sizes initialized from running statistics; LSQ+ uses learnable
zero-points
\item \textbf{PACT}: clipping thresholds initialized uniformly and learned per branch
\item \textbf{QIL}: interval centers and widths initialized from min--max statistics
\item \textbf{DSQ}: temperature parameter initialized to a low value and annealed during training
\item \textbf{DoReFa}: fixed $\tanh$-based normalization (no learnable parameters)
\end{itemize}

\subsection{Post-Training Quantization (PTQ)}

\paragraph{Calibration.}
Unless stated otherwise, PTQ uses 512--2048 calibration samples drawn from the training set.
For data-free methods, synthetic inputs are generated following the original method
specifications.

\paragraph{Method-specific settings.}
\begin{itemize}
\item \textbf{AdaRound}: soft-to-hard rounding with cosine annealing
\item \textbf{BRECQ}: block-wise reconstruction with joint branch calibration
\item \textbf{AWQ}: per-channel activation-aware scaling applied independently per branch
\item \textbf{SmoothQuant}: activation scaling folded into weights, applied separately to
base and spline/basis pathways
\item \textbf{GPTQ}: damped least-squares optimization with branch-specific curvature estimation
\item \textbf{HAWQ-V2}: Hessian trace estimation with sub-weight-level bit allocation
\item \textbf{ZeroQ}: synthetic calibration using TV+$\ell_2$ regularization
\end{itemize}

\subsection{Reproducibility}

All experiments use deterministic data loading when enabled, fixed random seeds, and identical pretrained checkpoints across quantization methods.
The full set of configuration files used in this work is included with the codebase to enable exact reproduction of all reported results.

\clearpage
\subsection{Full QAT Results}
\label{app:full_qat_res}
\begin{table*}[!ht]
\centering
\small
\caption{Accuracy (\%) of QAT methods on different datasets under various bit-precision settings. (\textbf{w4} means \textbf{w4a32}, likewise \textbf{w3}=\textbf{w3a32}, \textbf{w2}=\textbf{w2a32}).}
\vspace{0.5em}
\begin{tabular}{lccccccccc}
\toprule
\textbf{Architecture} & \textbf{Method} & \textbf{w32a32} & \textbf{w8} & \textbf{w4} & \textbf{w3} & \textbf{w2} & \textbf{w4a4} & \textbf{w3a3} & \textbf{w2a2} \\
\midrule
\multicolumn{10}{c}{MNIST dataset} \\
\midrule

\multirow{6}{*}{KAN FCN}
 & LSQ    & \multirow{6}{*}{98.00} & 98.22 & 98.11 & 98.00 & 97.55 & 97.94 & 97.70 & 96.76 \\
 & LSQ+   &                       & 98.24 & 98.13 & 98.03 & 97.79 & 97.87 & 97.73 & 96.97 \\
 & DoReFa &                       & 98.19 & 98.14 & 97.96 & 97.63 & 95.82 & 95.62 & 94.95 \\
  & DSQ    &                       & 98.43 & 98.24 & 98.13 & 97.71 & 97.71 & 97.82 & 97.71 \\
 & PACT   &                       & 98.17 & 98.14 & 98.07 & 97.21 & 97.77 & 97.64 & 96.09 \\
 & QIL    &                       & 98.30 & 98.04 & 97.37 & 93.15 & 97.61 & 95.95 & 22.17 \\ 
\midrule

\multirow{6}{*}{KAN ConvNet}
 & LSQ    & \multirow{6}{*}{95.70} & 18.53 & 94.00 & 94.40 & 94.95 & 92.44 & 91.35 & 90.78 \\
 & LSQ+   &                       & 11.35 & 95.56 & 95.17 & 94.46 & 90.95 & 94.11 & 92.04 \\
 & DoReFa &                       & 95.41 & 95.67 & 95.17 & 93.55 & 91.09 & 91.08 & 88.52 \\
  & DSQ    &                       & 95.81 & 95.44 & 95.62 & 89.54 & 90.40 & 90.14 & 89.54 \\
 & PACT   &                       & 95.12 & 94.28 & 91.07 & 11.35 & 94.39 & 90.25 & 11.35 \\
 & QIL    &                       & 94.34 & 93.03 & 87.86 & 11.35 & 93.17 & 91.74 & 11.35 \\ 
\midrule

\multirow{6}{*}{KAGN Simple}
 & LSQ    & \multirow{6}{*}{98.67} & 98.95 & 98.69 & 98.74 & 98.34 & 86.25 & 78.42 & 76.32 \\
 & LSQ+   &                       & 97.15 & 98.49 & 98.50 & 98.35 & 89.03 & 88.33 & 91.93 \\
 & DoReFa &                       & 98.70 & 98.71 & 98.58 & 98.33 & 97.93 & 97.96 & 96.22 \\
  & DSQ    &                       & 98.43 & 98.24 & 98.64 & 85.14 & 96.66 & 93.28 & 85.14 \\
 & PACT   &                       & 98.24 & 98.61 & 97.63 & 11.35 & 98.36 & 94.93 & 11.35 \\
 & QIL    &                       & 98.80 & 98.76 & 97.76 & 75.85 & 85.56 & 28.29 & 11.35 \\ 
\midrule

\multicolumn{10}{c}{CIFAR-10 dataset} \\
\midrule

\multirow{6}{*}{KAGN Simple}
 & LSQ    & \multirow{6}{*}{62.24} & 62.74 & 61.74 & 61.70 & 56.03 & 28.27 & 30.48 & 29.24 \\
 & LSQ+   &                       & 62.03 & 61.38 & 60.90 & 58.26 & 29.23 & 29.30 & 29.14 \\
 & DoReFa &                       & 62.84 & 61.68 & 58.88 & 55.87 & 50.01 & 47.38 & 37.15 \\
  & DSQ    &                       & 62.77 & 62.88 & 61.31 & 45.44 & 48.60 & 48.33 & 45.44 \\
 & PACT   &                       & 59.58 & 54.15 & 40.94 & 10.00 & 37.64 & 26.14 & 10.00 \\
 & QIL    &                       & 61.57 & 49.76 & 35.57 & 10.00 & 28.23 & 14.35 & 10.00 \\
\midrule

\multirow{6}{*}{\shortstack{KAGN Simple\\8 Layers}}
 & LSQ    & \multirow{6}{*}{78.10} & 76.00 & 75.49 & 73.21 & 72.10 & 46.41 & 47.67 & 47.38 \\
 & LSQ+   &                       & 70.27 & 74.71 & 75.10 & 72.24 & 47.47 & 46.15 & 39.65 \\
 & DoReFa &                       & 76.87 & 70.27 & 64.99 & 52.49 & 49.79 & 38.00 & 30.57 \\
  & DSQ    &                       & 72.22 & 76.86 & 77.36 & 39.44 & 50.64 & 48.05 & 39.44 \\
 & PACT   &                       & 76.60 & 70.22 & 10.00 & 10.00 & 50.39 & 10.00 & 10.00 \\
 & QIL    &                       & 77.31 & 71.93 & 52.95 & 10.00 & 39.72 & 12.54 & 10.00 \\
\midrule

\multicolumn{10}{c}{CIFAR-100 dataset} \\
\midrule

\multirow{6}{*}{\shortstack{KAGN Simple\\8 Layers V2}}
 & LSQ    & \multirow{6}{*}{68.06} & 67.20 & 65.87 & 68.49 & 68.41 & 52.36 & 53.01 & 47.00 \\
 & LSQ+   &                       & 66.50 & 67.70 & 68.12 & 64.35 & 58.27 & 59.64 & 1.30 \\
 & DoReFa &                       & 67.93 & 66.89 & 53.58 & 23.39 & 29.49 & 9.66  & 8.62 \\
  & DSQ    &                       & 67.12 & 68.03 & 67.54 & 43.10 & 34.01 & 35.45 & 43.10 \\
 & PACT   &                       & 68.20 & 66.94 & 59.39 & 1.00  & 47.92 & 20.43 & 1.00 \\
 & QIL    &                       & 68.11 & 61.14 & 1.02  & 1.00  & 2.19  & 1.01  & 1.00 \\
\midrule

\multirow{6}{*}{\shortstack{VGG Like KAGN V2}}
 & LSQ    & \multirow{6}{*}{58.77} & 46.15 & 58.51 & 56.31 & 52.99 & \textbf{1.00}  & \textbf{1.00}  & 1.00 \\
 & LSQ+   &                       & 58.05 & 56.98 & \textbf{1.00}  & 30.18 & \textbf{1.00}  & \textbf{1.00}  & 1.01 \\
 & DoReFa &                       & 58.50 & 48.96 & \textbf{1.00}  & 21.61 & 18.37 & 1.00  & 1.08 \\
  & DSQ    &                       & 57.82 & 58.58 & 58.23 & 1.00  & 56.81 & 55.86 & 1.00  \\
 & PACT   &                       & 58.64 & 30.97 & 25.09 & 1.00  & 7.13  & 1.16  & 1.00 \\
 & QIL    &                       & 59.15 & 18.86 & 1.15  & 1.00  & 12.25 & 1.13  & 1.00 \\

\bottomrule
\end{tabular}
\vspace{-0.5em}
\label{tab:kan_qat}
\end{table*}

\clearpage
\subsection{Full PTQ Results}

\begin{table*}[!ht]
\centering
\small
\caption{Accuracy (\%) of PTQ methods on different datasets under various bit-precision settings. (\textbf{w4} means \textbf{w4a32}, likewise \textbf{w3}=\textbf{w3a32}, \textbf{w2}=\textbf{w2a32}).}
\vspace{0.5em}
\begin{tabular}{lccccccccc}
\toprule
\textbf{Architecture} & \textbf{Method} & \textbf{w32a32} & \textbf{w8} & \textbf{w4} & \textbf{w3} & \textbf{w2} & \textbf{w4a4} & \textbf{w3a3} & \textbf{w2a2} \\
\midrule
\multicolumn{10}{c}{MNIST dataset} \\
\midrule

\multirow{7}{*}{KAN FCN}
  & GPTQ     & \multirow{7}{*}{97.99} & 98.00 & 97.86 & 97.07 & 76.58 & 95.88 & 88.99 & 40.90 \\
  & Adaround &                        & 97.97 & 97.78 & 96.36 & 42.87 & 95.75 & 88.88 & 32.35 \\
  & AWQ      &                        & 97.98 & 97.80 & 95.53 & 33.65 & 95.55 & 88.73 & 21.02 \\
  & BRECQ    &                        & 73.77 & 72.58 & 73.60 & 50.51 & 70.87 & 86.93 & 52.38 \\
  & Uniform  &                        & 97.97 & 97.81 & 96.15 & 39.48 & 97.81 & 96.15 & 39.48 \\
  & HAWQ-V2  &                        & 97.97 & 97.97 & 97.81 & 96.14 & --    & --    & --    \\
  & ZeroQ    &                        & 98.00 & 97.75 & 96.86 & 48.89 & 95.19 & 90.81 & 29.06 \\
\midrule

\multirow{7}{*}{KAN ConvNet}
  & GPTQ     & \multirow{7}{*}{95.69} & 95.69 & 95.65 & 95.63 & 88.18 & 10.10 & 10.10 & 10.10 \\
  & Adaround &                        & 95.69 & 95.61 & 95.20 & 80.03 & 9.73  & 10.28 & 10.25 \\
  & AWQ      &                        & 95.68 & 95.62 & 95.01 & 62.70 & 10.10 & 10.10 & 10.32 \\
  & BRECQ    &                        & 36.37 & 37.79 & 30.31 & 35.07 & 75.60 & 93.52 & 64.97 \\
  & Uniform  &                        & 95.68 & 95.51 & 94.51 & 68.63 & 95.51 & 94.51 & 68.63 \\
  & HAWQ-V2  &                        & 95.68 & 95.69 & 95.49 & 94.37 & --    & --    & --    \\
  & ZeroQ    &                        & 95.70 & 95.67 & 95.52 & 84.32 & 10.10 & 10.10 & 8.92  \\
\midrule

\multirow{7}{*}{KAGN Simple}
  & GPTQ     & \multirow{7}{*}{98.66} & 98.65 & 97.96 & 90.11 & 27.39 & 97.96 & 90.11 & 27.39 \\
  & Adaround &                        & 98.76 & 96.59 & 88.41 & 29.83 & 96.59 & 88.41 & 29.83 \\
  & AWQ      &                        & 98.70 & 97.35 & 57.65 & 13.36 & 97.35 & 57.65 & 13.36 \\
  & BRECQ    &                        & 98.73 & 96.11 & 76.07 & 23.33 & 96.22 & 81.70 & 12.80 \\
  & Uniform  &                        & 98.73 & 95.83 & 74.84 & 20.36 & 95.83 & 74.84 & 20.36 \\
  & HAWQ-V2  &                        & 98.73 & 34.74 & 35.27 & 27.30 & --    & --    & --    \\
  & ZeroQ    &                        & 98.66 & 97.90 & 88.91 & 12.13 & 97.90 & 88.91 & 12.13 \\
\midrule

\multicolumn{10}{c}{CIFAR-10 dataset} \\
\midrule

\multirow{7}{*}{KAGN Simple}
  & GPTQ     & \multirow{7}{*}{62.23} & 62.36 & 56.52 & 40.85 & 11.68 & 56.52 & 40.85 & 11.68 \\
  & Adaround &                        & 62.27 & 53.92 & 34.21 & 11.61 & 53.92 & 34.21 & 11.61 \\
  & AWQ      &                        & 62.17 & 54.01 & 22.01 & 9.58  & 54.01 & 22.01 & 9.58  \\
  & BRECQ    &                        & 62.28 & 52.28 & 24.47 & 12.35 & 49.98 & 27.43 & 16.21 \\
  & Uniform  &                        & 62.19 & 51.67 & 23.95 & 15.01 & 51.67 & 23.95 & 15.01 \\
  & HAWQ-V2  &                        & 62.29 & 31.56 & 32.05 & 20.31 & --    & --    & --    \\
  & ZeroQ    &                        & 62.30 & 56.01 & 44.71 & 16.79 & 56.01 & 44.71 & 16.79 \\
\midrule

\multirow{7}{*}{\shortstack{KAGN Simple\\8 Layers}}
  & GPTQ     & \multirow{7}{*}{78.08} & 78.02 & 69.13 & 41.59 & 9.92  & 69.13 & 41.59 & 9.92  \\
  & Adaround &                        & 78.15 & 61.57 & 37.93 & 10.09 & 61.57 & 37.93 & 10.09 \\
  & AWQ      &                        & 77.92 & 60.17 & 28.01 & 12.17 & 60.17 & 28.01 & 12.17 \\
  & BRECQ    &                        & 77.88 & 60.28 & 27.64 & 12.82 & 62.28 & 39.31 & 19.20 \\
  & Uniform  &                        & 77.91 & 59.47 & 30.17 & 9.11  & 59.47 & 30.17 & 9.11  \\
  & HAWQ-V2  &                        & 77.89 & 43.13 & 28.29 & 24.29 & --    & --    & --    \\
  & ZeroQ    &                        & 77.94 & 70.41 & 39.71 & 13.31 & 70.41 & 39.71 & 13.31 \\
\midrule

\multicolumn{10}{c}{CIFAR-100 dataset} \\
\midrule

\multirow{7}{*}{\shortstack{KAGN Simple\\8 Layers V2}}
  & GPTQ     & \multirow{7}{*}{68.09} & 68.05 & 13.47 & 2.65  & 1.14  & 13.47 & 2.65  & 1.14  \\
  & Adaround &                        & 67.67 & 2.77  & 1.01  & 1.00  & 2.77  & 1.01  & 1.00  \\
  & AWQ      &                        & 67.62 & 3.74  & 1.00  & 1.00  & 3.74  & 1.00  & 1.00  \\
  & BRECQ    &                        & 67.57 & 3.86  & 1.00  & 1.00  & 3.12  & 1.49  & 1.47  \\
  & Uniform  &                        & 67.66 & 3.74  & 1.00  & 1.00  & 3.74  & 1.00  & 1.00  \\
  & HAWQ-V2  &                        & 67.57 & 1.62  & 1.08  & 1.03  & --    & --    & --    \\
  & ZeroQ    &                        & 68.07 & 13.06 & 2.24  & 1.46  & 13.06 & 2.24  & 1.46  \\
\midrule

\multirow{7}{*}{\shortstack{VGG Like\\KAGN V2}}
  & GPTQ     & \multirow{7}{*}{58.70} & 58.23 & 39.73 & 14.23 & 1.98  & 39.57 & 13.26 & 0.99  \\
  & Adaround &                        & 55.91 & 22.66 & 6.05  & 1.47  & 22.63 & 5.46  & 1.03  \\
  & AWQ      &                        & 58.30 & 29.57 & 13.44 & 1.22  & 29.44 & 12.48 & 1.15  \\
  & BRECQ    &                        & 58.29 & 29.80 & 15.64 & 1.88  & 32.17 & 15.99 & 2.57  \\
  & Uniform  &                        & 55.88 & 21.08 & 3.65  & 1.15  & 21.08 & 3.65  & 1.15  \\
  & HAWQ-V2  &                        & 58.31 & 12.06 & 7.99  & 5.56  & --    & --    & --    \\
  & ZeroQ    &                        & 58.17 & 39.68 & 14.29 & 1.55  & 39.47 & 13.85 & 1.15  \\
\bottomrule
\end{tabular}
\vspace{-0.5em}
\label{tab:ptq_kan_all}
\end{table*}

\subsection{Granularity \& Symmetry Ablations}
\begin{figure}[!ht]
  \centering
  \includegraphics[width=0.7\linewidth]{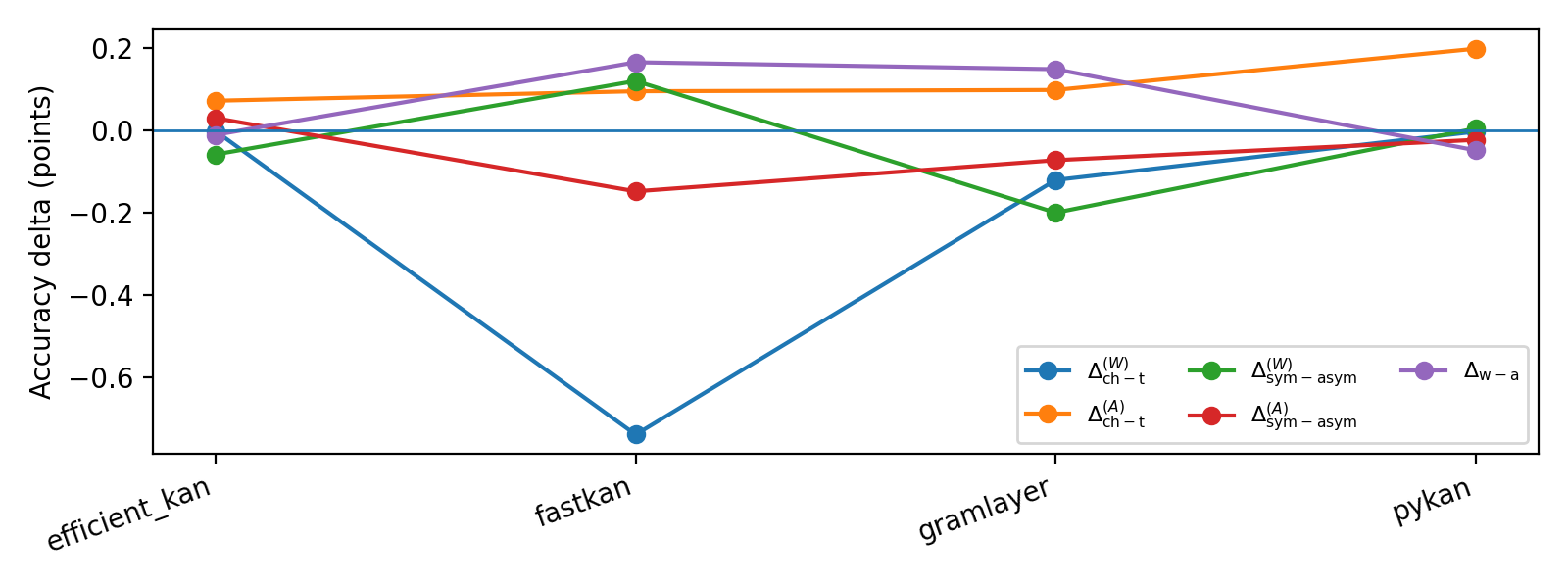}
  \caption{CIFAR-10 ablation summary across KAN variants. We plot average marginal effects aggregated over bit-widths and quantizers:
  $\Delta_{\text{ch-t}}^{(W/A)}$ compares per-channel vs.\ per-tensor scaling for weights/activations,
  $\Delta_{\text{sym-asym}}^{(W/A)}$ compares symmetric vs.\ asymmetric quantization,
  and $\Delta_{\text{w-a}}$ compares weight vs.\ activation sensitivity. Positive values indicate higher accuracy.}
  \label{fig:ablation_effects}
\end{figure}

Figure~\ref{fig:ablation_effects}, we summarize average marginal effects of (i) granularity (per-channel vs.\ per-tensor), (ii) symmetry (symmetric vs.\ asymmetric), and (iii) quantization target (weights vs.\ activations), aggregated over bit-widths and quantizers on CIFAR-10. Tables~\ref{tab:ablation_full_cifar10_weight_methodcols} and
\ref{tab:ablation_full_cifar10_activation_methodcols} report exhaustive CIFAR-10 results.
Rows index (variant, bit-width), and columns enumerate symmetry--granularity settings for each quantizer.

\begin{table*}[!ht]
\centering
\scriptsize
\setlength{\tabcolsep}{2.4pt}
\renewcommand{\arraystretch}{0.88}
\caption{Full ablation results on CIFAR10 with quantization applied to \textbf{weights}. Each cell reports Top-1 accuracy (\%).}
\vspace{0.2em}
\begin{tabular}{llcccccccccccc}
\toprule
\textbf{Variant} & \textbf{Bits} & \multicolumn{4}{c}{\textbf{DoReFa}} & \multicolumn{4}{c}{\textbf{DSQ}} & \multicolumn{4}{c}{\textbf{LSQ}} \\
\cmidrule(lr){1-2}\cmidrule(lr){3-6}\cmidrule(lr){7-10}\cmidrule(lr){11-14}
 &  & \textbf{Asym-Ch} & \textbf{Asym-T} & \textbf{Sym-Ch} & \textbf{Sym-T} & \textbf{Asym-Ch} & \textbf{Asym-T} & \textbf{Sym-Ch} & \textbf{Sym-T} & \textbf{Asym-Ch} & \textbf{Asym-T} & \textbf{Sym-Ch} & \textbf{Sym-T} \\
\midrule
eff kan & w8a32 & 49.67 & 49.60 & 49.78 & 49.77 & 49.87 & 49.93 & 49.82 & 49.41 & 49.78 & 49.84 & 49.78 & 49.75 \\
 & w4a32 & 49.34 & 49.74 & 49.34 & 49.59 & 49.39 & 49.61 & 49.53 & 49.65 & 49.64 & 49.76 & 49.73 & 49.69 \\
 & w3a32 & 49.56 & 48.67 & 49.36 & 49.31 & 49.71 & 49.63 & 49.67 & 49.57 & 49.39 & 49.51 & 49.26 & 49.17 \\
 & w2a32 & 47.87 & 48.04 & 48.12 & 48.62 & 49.92 & 49.66 & 49.54 & 49.33 & 49.78 & 48.65 & 48.57 & 48.22 \\
 & w4a4 & 44.72 & 44.67 & 44.75 & 44.68 & 49.72 & 50.16 & 49.63 & 49.76 & 32.68 & 33.65 & 33.08 & 33.86 \\
 & w3a3 & 44.43 & 44.30 & 44.25 & 44.42 & 49.87 & 50.02 & 49.84 & 49.88 & 33.80 & 33.26 & 33.19 & 33.39 \\
 & w2a2 & 41.81 & 42.10 & 42.04 & 41.87 & 49.85 & 49.97 & 50.42 & 50.17 & 32.75 & 32.70 & 32.52 & 32.25 \\
\midrule
fastkan & w8a32 & 39.93 & 39.86 & 39.74 & 40.26 & 39.97 & 39.76 & 40.11 & 39.94 & 39.70 & 41.57 & 40.12 & 39.76 \\
 & w4a32 & 39.28 & 39.01 & 39.38 & 39.13 & 39.75 & 40.47 & 40.17 & 40.18 & 39.40 & 39.35 & 39.75 & 39.11 \\
 & w3a32 & 37.17 & 37.33 & 37.41 & 37.86 & 40.21 & 40.14 & 40.06 & 40.37 & 38.27 & 40.40 & 39.56 & 40.91 \\
 & w2a32 & 30.97 & 31.07 & 31.16 & 30.42 & 40.22 & 40.95 & 40.40 & 40.93 & 40.32 & 41.44 & 40.07 & 41.44 \\
 & w4a4 & 30.12 & 29.99 & 29.58 & 29.84 & 35.58 & 10.00 & 35.69 & 36.57 & 28.34 & 28.01 & 27.24 & 28.21 \\
 & w3a3 & 28.07 & 28.34 & 29.06 & 29.26 & 35.86 & 10.00 & 34.48 & 35.73 & 27.15 & 31.50 & 27.42 & 31.22 \\
 & w2a2 & 24.81 & 25.15 & 24.84 & 25.24 & 34.76 & 10.00 & 34.55 & 10.00 & 25.03 & 30.52 & 28.88 & 30.73 \\
\midrule
gramlayer & w8a32 & 49.89 & 49.79 & 49.71 & 49.85 & 49.76 & 49.44 & 49.52 & 49.68 & 46.57 & 47.92 & 48.49 & 49.68 \\
 & w4a32 & 50.08 & 49.82 & 49.74 & 50.06 & 50.03 & 49.48 & 49.67 & 49.91 & 48.67 & 49.39 & 48.54 & 49.76 \\
 & w3a32 & 49.54 & 49.46 & 49.82 & 49.85 & 49.67 & 49.52 & 49.84 & 49.82 & 49.63 & 49.50 & 49.06 & 49.60 \\
 & w2a32 & 49.28 & 49.23 & 49.54 & 49.34 & 49.72 & 49.45 & 49.76 & 50.20 & 48.56 & 48.79 & 48.22 & 48.54 \\
 & w4a4 & 44.06 & 44.49 & 44.40 & 43.66 & 48.89 & 49.29 & 48.95 & 48.87 & 41.60 & 41.37 & 40.83 & 41.62 \\
 & w3a3 & 44.03 & 43.53 & 43.86 & 43.42 & 49.02 & 49.19 & 49.12 & 48.77 & 40.95 & 40.62 & 41.14 & 40.90 \\
 & w2a2 & 41.16 & 41.07 & 40.97 & 41.44 & 48.63 & 48.91 & 48.80 & 49.00 & 37.63 & 37.74 & 31.29 & 31.73 \\
\midrule
pykan & w8a32 & 50.84 & 50.69 & 50.92 & 50.33 & 50.53 & 50.42 & 50.58 & 50.23 & 50.50 & 50.35 & 50.74 & 50.97 \\
 & w4a32 & 50.62 & 50.51 & 50.27 & 50.57 & 50.47 & 50.78 & 50.41 & 50.88 & 50.21 & 50.41 & 50.97 & 50.15 \\
 & w3a32 & 50.01 & 50.27 & 50.51 & 49.98 & 50.76 & 50.98 & 50.75 & 50.65 & 50.24 & 50.26 & 50.32 & 50.51 \\
 & w2a32 & 49.16 & 49.79 & 49.66 & 49.61 & 50.21 & 50.69 & 50.38 & 50.60 & 50.41 & 49.59 & 49.98 & 49.05 \\
 & w4a4 & 44.82 & 44.84 & 45.14 & 44.89 & 50.72 & 50.76 & 50.67 & 50.77 & 31.76 & 31.96 & 31.63 & 32.48 \\
 & w3a3 & 44.49 & 44.38 & 44.52 & 44.46 & 51.02 & 50.53 & 51.10 & 51.28 & 32.16 & 31.88 & 31.75 & 32.26 \\
 & w2a2 & 41.85 & 42.01 & 41.76 & 41.78 & 51.16 & 50.96 & 51.05 & 50.87 & 31.54 & 31.85 & 30.85 & 31.32 \\
\bottomrule
\end{tabular}
\label{tab:ablation_full_cifar10_weight_methodcols}
\end{table*}

\begin{table*}[!ht]
\centering
\scriptsize
\setlength{\tabcolsep}{2.4pt}
\renewcommand{\arraystretch}{0.88}
\caption{Full ablation results on CIFAR10 with quantization applied to \textbf{activations}. Each cell reports Top-1 accuracy (\%).}
\vspace{0.2em}
\begin{tabular}{llcccccccccccc}
\toprule
\textbf{Variant} & \textbf{Bits} & \multicolumn{4}{c}{\textbf{DoReFa}} & \multicolumn{4}{c}{\textbf{DSQ}} & \multicolumn{4}{c}{\textbf{LSQ}} \\
\cmidrule(lr){1-2}\cmidrule(lr){3-6}\cmidrule(lr){7-10}\cmidrule(lr){11-14}
 &  & \textbf{Asym-Ch} & \textbf{Asym-T} & \textbf{Sym-Ch} & \textbf{Sym-T} & \textbf{Asym-Ch} & \textbf{Asym-T} & \textbf{Sym-Ch} & \textbf{Sym-T} & \textbf{Asym-Ch} & \textbf{Asym-T} & \textbf{Sym-Ch} & \textbf{Sym-T} \\
\midrule
eff kan & w8a32 & 49.56 & 49.86 & 49.60 & 49.74 & 49.75 & 49.65 & 49.56 & 49.62 & 49.47 & 49.38 & 49.65 & 49.76 \\
 & w4a32 & 49.33 & 49.69 & 49.75 & 49.51 & 49.69 & 49.46 & 49.55 & 49.78 & 49.83 & 49.45 & 49.54 & 49.53 \\
 & w3a32& 49.15 & 48.98 & 49.24 & 49.30 & 49.78 & 49.90 & 49.75 & 49.63 & 49.26 & 48.90 & 49.52 & 49.44 \\
 & w2a32& 48.32 & 48.06 & 47.91 & 48.63 & 49.44 & 49.79 & 49.56 & 49.61 & 48.80 & 49.08 & 48.36 & 48.80 \\
 & w4a4 & 44.54 & 44.71 & 44.55 & 44.76 & 49.75 & 49.60 & 49.74 & 49.84 & 33.51 & 32.95 & 33.85 & 33.31 \\
 & w3a3 & 44.55 & 44.52 & 44.78 & 44.35 & 49.94 & 49.94 & 49.74 & 49.99 & 34.62 & 33.06 & 34.18 & 33.19 \\
 & w2a2 & 42.19 & 41.88 & 41.64 & 41.92 & 49.75 & 50.16 & 49.94 & 50.17 & 32.92 & 32.43 & 33.20 & 32.38 \\
\midrule
fastkan & w8a32 & 39.90 & 39.97 & 39.95 & 40.08 & 40.33 & 39.85 & 40.08 & 39.71 & 39.38 & 39.48 & 38.87 & 39.20 \\
 & w4a32 & 39.35 & 39.36 & 39.08 & 39.40 & 40.16 & 40.05 & 40.23 & 40.03 & 38.19 & 39.30 & 39.52 & 39.06 \\
 & w3a32& 37.53 & 37.37 & 37.90 & 37.78 & 40.04 & 39.87 & 39.84 & 40.17 & 38.66 & 39.22 & 38.48 & 39.15 \\
 & w2a32 & 31.12 & 30.83 & 31.31 & 31.18 & 39.96 & 40.11 & 40.01 & 40.09 & 40.80 & 40.22 & 40.82 & 40.27 \\
 & w4a4 & 30.78 & 29.67 & 30.22 & 30.10 & 37.31 & 36.16 & 36.22 & 36.83 & 28.39 & 28.23 & 28.49 & 28.20 \\
 & w3a3 & 28.51 & 28.29 & 28.50 & 28.75 & 36.12 & 36.60 & 36.22 & 35.96 & 29.35 & 27.97 & 28.71 & 27.24 \\
 & w2a2 & 25.25 & 25.13 & 25.10 & 25.12 & 32.38 & 34.95 & 33.29 & 33.62 & 30.20 & 29.06 & 27.75 & 26.65 \\
\midrule
gramlayer & w8a32 & 50.04 & 49.46 & 49.65 & 49.71 & 50.04 & 49.56 & 49.56 & 50.19 & 50.12 & 49.65 & 47.85 & 48.47 \\
 & w4a32 & 49.51 & 49.73 & 49.88 & 49.88 & 49.89 & 49.84 & 49.51 & 49.69 & 49.58 & 49.16 & 49.61 & 49.72 \\
 & w3a32 & 49.55 & 49.65 & 49.68 & 49.58 & 49.74 & 49.78 & 50.35 & 49.67 & 49.55 & 49.35 & 48.87 & 49.44 \\
 & w2a32 & 49.13 & 48.88 & 48.76 & 48.89 & 50.14 & 49.78 & 50.18 & 50.04 & 48.42 & 48.70 & 48.92 & 48.26 \\
 & w4a4 & 43.89 & 44.35 & 44.35 & 43.75 & 48.84 & 49.12 & 48.86 & 48.74 & 41.68 & 41.40 & 41.12 & 40.63 \\
 & w3a3 & 43.22 & 43.86 & 43.68 & 43.37 & 48.47 & 48.55 & 48.66 & 49.02 & 41.00 & 40.25 & 41.21 & 41.10 \\
 & w2a2 & 41.01 & 40.99 & 41.26 & 40.48 & 48.29 & 48.90 & 48.99 & 48.95 & 31.81 & 31.53 & 32.10 & 30.75 \\
\midrule
pykan & w8a32 & 50.78 & 50.64 & 50.79 & 50.42 & 50.52 & 50.52 & 50.67 & 50.37 & 50.43 & 50.55 & 50.56 & 50.69 \\
 & w4a32 & 50.57 & 50.44 & 50.44 & 50.45 & 50.61 & 50.34 & 50.53 & 50.52 & 50.15 & 50.65 & 50.39 & 50.41 \\
 & w3a32 & 50.43 & 50.34 & 50.30 & 50.20 & 50.71 & 50.54 & 50.91 & 50.34 & 49.96 & 50.08 & 49.85 & 49.77 \\
 & w2a32 & 50.64 & 49.50 & 49.13 & 50.28 & 50.47 & 50.41 & 50.53 & 50.57 & 50.17 & 49.87 & 49.71 & 50.29 \\
 & w4a4 & 44.90 & 44.64 & 44.77 & 44.82 & 50.68 & 50.87 & 50.69 & 50.66 & 32.06 & 32.13 & 32.73 & 31.93 \\
 & w3a3 & 44.37 & 44.68 & 44.72 & 44.32 & 50.68 & 51.01 & 50.78 & 50.61 & 33.06 & 31.70 & 33.25 & 31.86 \\
 & w2a2 & 42.16 & 41.61 & 41.77 & 41.68 & 50.74 & 50.84 & 51.09 & 50.69 & 33.20 & 31.34 & 32.80 & 31.75 \\
\bottomrule
\end{tabular}
\label{tab:ablation_full_cifar10_activation_methodcols}
\end{table*}

\begin{figure*}[!ht]
  \centering
  \includegraphics[width=0.7\linewidth]{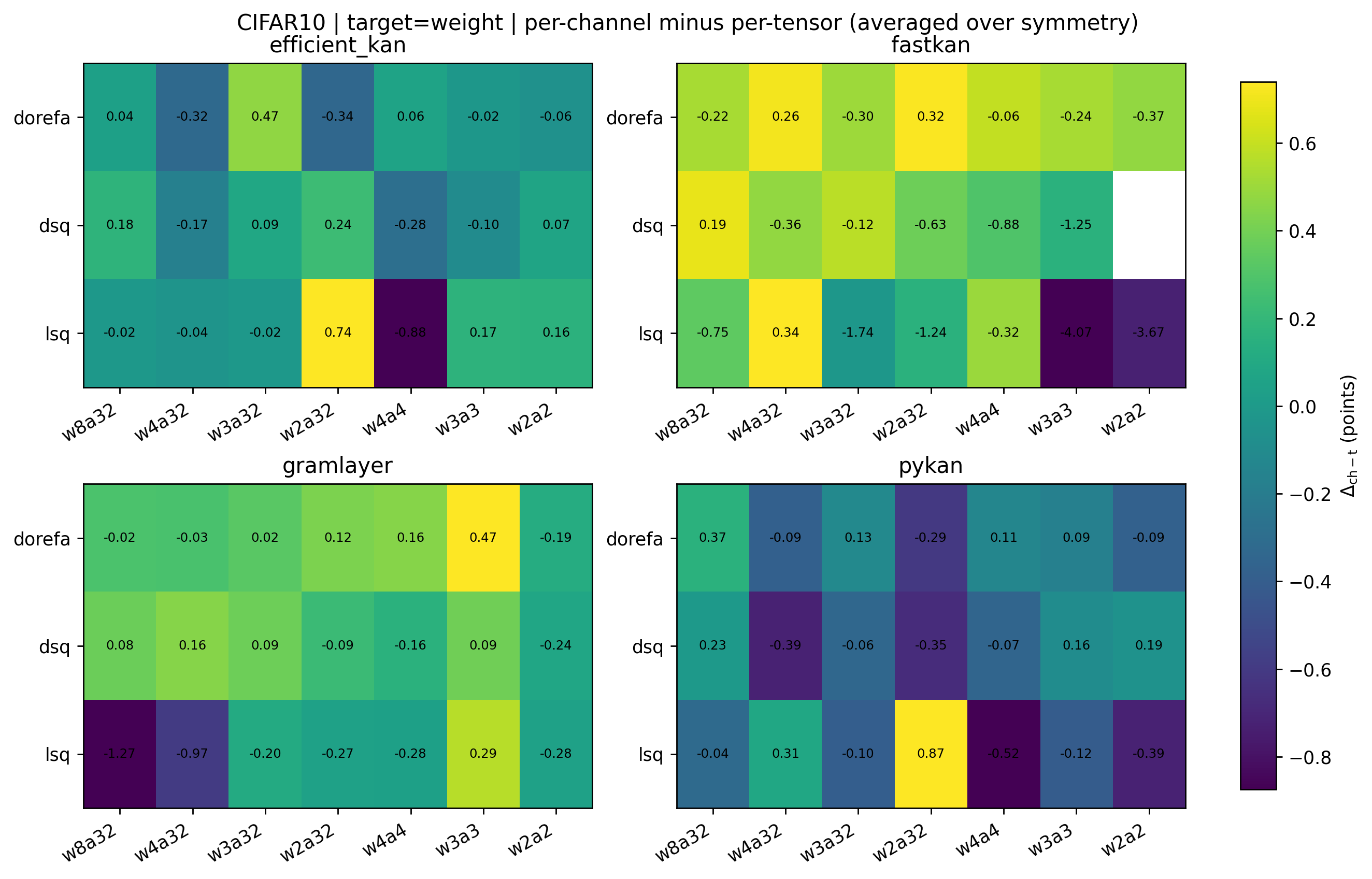}
  \caption{Granularity sensitivity for \textbf{weight} quantization on CIFAR-10.
  Each panel corresponds to a KAN variant; cells report $\Delta_{\mathrm{ch-t}}$ (per-channel minus per-tensor accuracy, averaged over symmetric/asymmetric ranges) across quantizers and bit-widths.}
  \label{fig:ablation_heat_weight}
\end{figure*}

\begin{figure*}[!ht]
  \centering
  \includegraphics[width=0.7\linewidth]{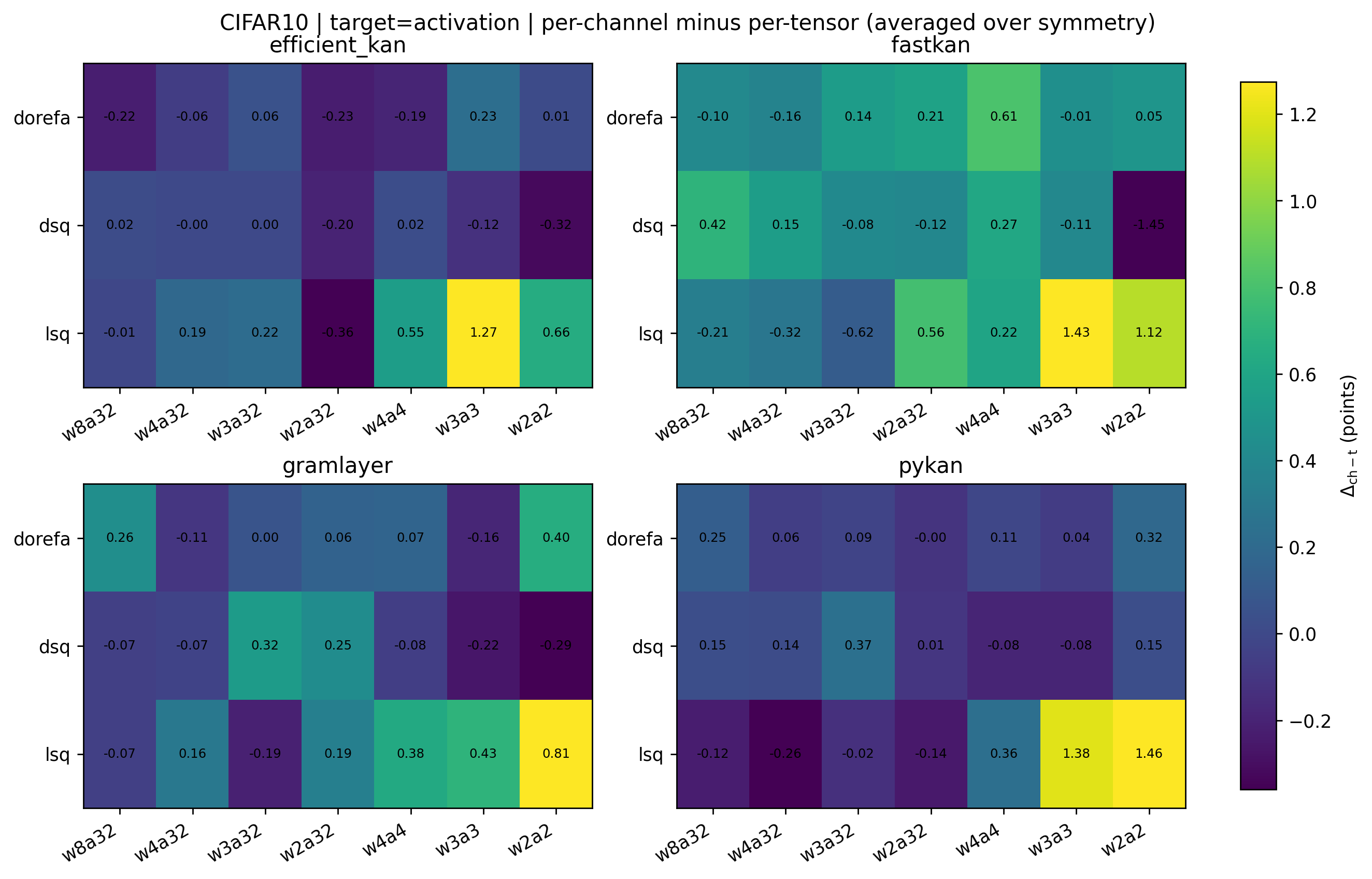}
  \caption{Granularity sensitivity for \textbf{activation} quantization on CIFAR-10.
  Each panel corresponds to a KAN variant; cells report $\Delta_{\mathrm{ch-t}}$ (per-channel minus per-tensor accuracy, averaged over symmetric/asymmetric ranges) across quantizers and bit-widths.}
  \label{fig:ablation_heat_act}
\end{figure*}

\paragraph{Granularity heatmaps.}
Figures~\ref{fig:ablation_heat_weight}--\ref{fig:ablation_heat_act} report $\Delta_{\mathrm{ch-t}}$,
the accuracy change from per-tensor to per-channel scaling (positive favors per-channel), averaged over symmetric/asymmetric ranges.
This isolates granularity sensitivity across quantizers and bit-widths for each KAN variant.

\subsection{Detailed Positioning Against Other KAN Compression Strategies}
\label{app:related-compression}

This appendix expands on the brief discussion in the main paper and
positions QuantKAN with respect to three recent KAN compression and
acceleration approaches: vector-clustering compression
(MetaCluster~\citep{raffel2025metacluster}), LUT-based FPGA inference
(KANEL\'E~\citep{hoang2026kanele}), and CIM-based edge inference for KANs
(Huang et al.~\citep{huang2025hardware}). Our goal is to clarify why these
approaches address \emph{different and largely complementary} aspects of
KAN deployment, and therefore why a head-to-head accuracy--efficiency
comparison would conflate non-substitutable techniques.

\subsection{Compression Axes}

The four approaches operate along distinct compression axes:

\begin{itemize}
\item \textbf{MetaCluster} reduces the \emph{number of distinct coefficient
vectors} per layer via meta-learned manifold shaping followed by $K$-means
clustering. Each per-edge vector is replaced by a centroid index, yielding
up to $80\times$ parameter-storage reduction on image classification and
$124.1\times$ on equation modeling, with centroids stored at full precision.
\item \textbf{KANEL\'E} replaces each learned spline activation with a
precomputed lookup table mapped directly onto FPGA LUT fabric, eliminating
basis evaluation and MAC arithmetic entirely. The resulting design is
LUT-native, sidesteps DSP and BRAM usage, and reports up to $2700\times$
latency improvement over prior KAN-on-FPGA work on small benchmarks.
\item \textbf{Huang et al.} accelerate KAN inference on RRAM-based analog
compute-in-memory hardware using an 8-bit quantization scheme co-designed
with the CIM substrate, targeting energy-efficient lightweight edge
inference.
\item \textbf{QuantKAN (ours)} reduces the \emph{bit-width} of base weights,
spline/basis weights, and activations via branch-aware QAT and PTQ, enabling
integer arithmetic and BOP reduction on standard datapaths (GPU integer
units, INT tensor cores, generic FPGA arithmetic).
\end{itemize}

These axes are not interchangeable. Clustering reduces the count of unique
vectors but leaves each centroid in floating point. LUT and CIM mappings
eliminate arithmetic by replacing it with table lookups or analog
conductance reads, but require co-designed hardware. Bit-width quantization
reduces both storage and arithmetic cost on commodity integer hardware.
A complete deployment stack can in principle combine multiple axes
simultaneously.

\subsection{Orthogonality and Complementarity}

\paragraph{MetaCluster.} The orthogonality of bit-width quantization and
vector clustering is acknowledged explicitly by the MetaCluster authors,
who state that ``bit-width quantization is an alternative avenue, which is
largely orthogonal and complementary to the proposed approach. Applying
quantization on top of MetaCluster (e.g., quantizing the codebook of
centroids) can further reduce the memory footprint''~\citep[Sec.~5.3]{raffel2025metacluster}.
A direct head-to-head comparison would therefore not measure the relative
merit of the two techniques but rather their effect on different parts of
the parameter representation. The natural integration is a stacked design
in which MetaCluster shapes and clusters per-edge vectors and QuantKAN
quantizes the resulting centroid codebook and activations; we leave a
quantitative study of this combination to future work.

\paragraph{KANEL\'E and CIM-based KAN inference.} LUT-based and CIM-based
KAN accelerators target deployment regimes that are categorically
different from QuantKAN's. KANEL\'E maps each learned univariate activation
into FPGA LUTs after quantization and pruning; the resulting model is
hardware-bound to LUT-native inference and is co-designed with the FPGA
fabric. Huang et al.\ similarly co-design KAN inference with RRAM-based
analog CIM. Neither approach produces a model that runs on standard
integer arithmetic, and both have so far been demonstrated on
small-to-moderate benchmarks (e.g., MNIST, JSC, Moons, Wine, Dry Bean,
ToyADMOS).

A recent companion study of low-bit KAN quantization and tabulation
characterizes the scalability of full spline tabulation explicitly, noting
that this approach ``remains viable for models of limited dimensions\ldots
however, it does not scale well for larger models, such as CNN4 and
ResKAN18, exceeding the available resources by orders of magnitude on a
high-end FPGA''~\citep{errabii2026kantize}. By contrast,
QuantKAN targets bit-width quantization for general-purpose integer
datapaths and is evaluated on KAGN architectures up to ImageNet scale,
including a deep VGG-style KAGN backbone (Tables~\ref{tab:qat_tinyimagenet_imagenet} and
related). The two paradigms therefore occupy different points in the
deployment design space.

\subsection{Positioning Table}

Table~\ref{tab:compression-positioning} summarizes the comparison. All
numbers are taken directly from the cited papers; we use ``--'' where the
paper does not report the corresponding metric.

\begin{table}[h]
\centering
\scriptsize
\caption{Positioning of QuantKAN against recent KAN compression and
acceleration strategies. Compression axes are not directly substitutable;
see text. Numbers are reported as in the original papers.}
\label{tab:compression-positioning}
\begin{tabular}{lllll}
\toprule
\textbf{Approach} & \textbf{Axis} & \textbf{Storage / compute gain} &
\textbf{Hardware target} & \textbf{Largest dataset} \\
\midrule
MetaCluster & Vector clustering &
up to $80\times$ params &
General &
CIFAR-100 \\
KANEL\'E & LUT mapping (FPGA) &
LUT-native FPGA mapping &
FPGA (LUT-native) &
MNIST / JSC / ToyADMOS \\
Huang et al. & 8-bit + CIM &
Edge-class energy/latency &
RRAM CIM (analog) &
Lightweight edge \\
\textbf{QuantKAN} & Bit-width (QAT+PTQ) &
up to $16\times$ mem.,~$256\times$ BOP &
General + FPGA &
ImageNet \\
\bottomrule
\end{tabular}
\end{table}

\subsection{Implications for Comparison}

A head-to-head accuracy--efficiency comparison between QuantKAN and these
approaches would not isolate a meaningful design axis: each method achieves
its gains through a different mechanism (clustering versus tabulation
versus analog inference versus low-bit arithmetic), and reporting them on
a single Pareto plot would mix non-substitutable quantities (number of
unique vectors, LUT count, RRAM cells, BOPs) under a single ``efficiency''
label. The more informative question, which we leave to future work, is
whether stacking these axes (e.g., MetaCluster centroid quantization, or
LUT mapping of QuantKAN-quantized splines) yields multiplicative gains.
QuantKAN provides the bit-width foundation on which such stacks can be
built.


\clearpage
\section{Additional Parameter Sensitivity Results}
\label{sec:appendix_sensitivity}

This appendix provides the full parameter-wise sensitivity results that underlie the summary and design recommendations presented in the main paper.
Specifically, we report the degradation incurred when quantizing a single parameter group to low precision while keeping all remaining parameters at 8-bit.

\begin{figure}[!ht]
  \centering
  \includegraphics[width=0.6\columnwidth]{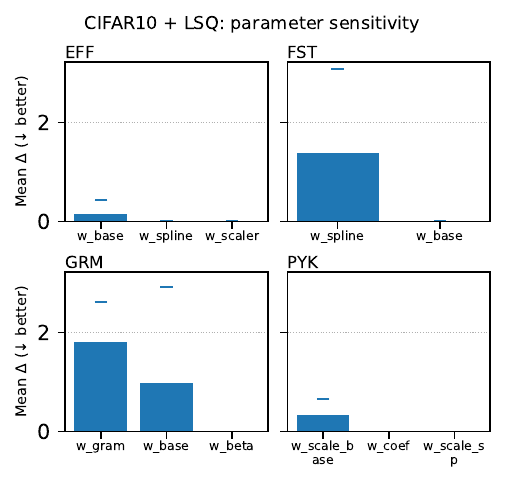}
  \caption{\textbf{CIFAR-10 + LSQ parameter-wise sensitivity under single-parameter mixed precision.}
Bars show mean degradation $\Delta=\max(0,\mathrm{Acc}_{\mathrm{FP}}-\mathrm{Acc})$ averaged over
$b\in\{2,3,4\}$; markers indicate worst-case degradation. Lower is better.
For GRM, $w_{\text{base}}$ ($\Delta\!\approx\!0.99$) and $w_{\text{gram}}$ ($\Delta\!\approx\!0.93$)
are within rounding; see Table~\ref{tab:appendix_param_sensitivity_full} for exact values.}
  \label{fig:param_sensitivity_cifar10_lsq}
  \vspace{-0.8em}
\end{figure}

Table~\ref{tab:appendix_param_sensitivity_full} reports mean degradation
$\Delta=\max(0,\mathrm{Acc}_{\mathrm{FP}}-\mathrm{Acc})$
for each parameter group and bit-width $b\in\{2,3,4\}$ on CIFAR-10 and MNIST, respectively.
Lower values indicate greater robustness to quantization.
These detailed results support the main paper’s conclusions regarding parameter sensitivity and the resulting mixed-precision design strategy.

\begin{table*}[!ht]
\centering
\caption{Full parameter-wise sensitivity under single-parameter mixed precision.
We report mean degradation $\Delta=\max(0,\mathrm{Acc}_{\mathrm{FP}}-\mathrm{Acc})$ when quantizing a single parameter group to bit-width $b\in\{2,3,4\}$ while keeping all others at 8-bit (lower is better).}
\vspace{-0.4em}
\scriptsize
\setlength{\tabcolsep}{4pt}
\renewcommand{\arraystretch}{1.1}

\begin{tabular}{l l c c c c c c}
\toprule
& & \multicolumn{3}{c}{\textbf{CIFAR-10}} & \multicolumn{3}{c}{\textbf{MNIST}} \\
\cmidrule(lr){3-5} \cmidrule(lr){6-8}
\textbf{Variant} & \textbf{Parameter}
& $b{=}2$ & $b{=}3$ & $b{=}4$
& $b{=}2$ & $b{=}3$ & $b{=}4$ \\
\midrule

EFF--DSQ
& $w_{\text{base}}$   & 0.00 & 0.00 & 0.00 & 0.00 & 0.00 & 0.00 \\
& $w_{\text{scaler}}$ & 0.00 & 0.00 & 0.00 & 0.00 & 0.00 & 0.00 \\
& $w_{\text{spline}}$ & 0.00 & 0.00 & 0.00 & 0.00 & 0.00 & 0.00 \\
\midrule

EFF--LSQ
& $w_{\text{base}}$   & 0.43 & 0.00 & 0.00 & \textbf{0.74} & 0.06 & 0.00 \\
& $w_{\text{scaler}}$ & 0.00 & 0.00 & 0.00 & 0.00 & 0.00 & 0.00 \\
& $w_{\text{spline}}$ & 0.00 & 0.00 & 0.00 & 0.00 & 0.00 & 0.00 \\
\midrule

FAST--DSQ
& $w_{\text{base}}$   & 0.00 & 0.00 & 0.00 & 0.00 & 0.00 & 0.00 \\
& $w_{\text{spline}}$ & 0.00 & 0.00 & 0.00 & 0.00 & 0.00 & 0.00 \\
\midrule

FAST--LSQ
& $w_{\text{base}}$   & 0.00 & 0.00 & 0.00 & \textbf{0.14} & 0.00 & 0.00 \\
& $w_{\text{spline}}$ & \textbf{3.07} & 1.05 & 0.00 & 0.00 & 0.00 & 0.00 \\
\midrule

GRAM--DSQ
& $w_{\text{base}}$ & 0.00 & 0.14 & 0.36 & 0.00 & 0.00 & 0.00 \\
& $w_{\text{beta}}$ & 0.00 & 0.07 & 0.12 & 0.00 & 0.00 & 0.00 \\
& $w_{\text{gram}}$ & 0.00 & 0.00 & 0.05 & 0.00 & 0.00 & 0.00 \\
\midrule

GRAM--LSQ
& $w_{\text{base}}$ & \textbf{2.92} & 0.04 & 0.00 & 0.31 & 0.06 & \textbf{0.48} \\
& $w_{\text{beta}}$ & 0.00 & 0.00 & 0.00 & 0.00 & 0.00 & 0.00 \\
& $w_{\text{gram}}$ & \textbf{2.62} & 0.33 & 2.46 & \textbf{0.46} & 0.00 & 0.11 \\
\midrule

PyKAN--DSQ
& $w_{\text{coef}}$       & 0.00 & 0.00 & 0.00 & 0.00 & 0.00 & 0.00 \\
& $w_{\text{scale,base}}$ & 0.00 & 0.00 & 0.00 & 0.00 & 0.00 & 0.00 \\
& $w_{\text{scale,sp}}$   & 0.00 & 0.00 & 0.00 & 0.00 & 0.00 & 0.00 \\
\midrule

PyKAN--LSQ
& $w_{\text{coef}}$       & -- & 0.00 & 0.00 & 0.00 & 0.00 & 0.00 \\
& $w_{\text{scale,base}}$ & \textbf{0.66} & -- & 0.00 & \textbf{0.30} & 0.02 & 0.00 \\
& $w_{\text{scale,sp}}$   & 0.00 & 0.00 & 0.00 & 0.00 & 0.00 & 0.00 \\
\bottomrule
\end{tabular}
\label{tab:appendix_param_sensitivity_full}
\vspace{-0.8em}
\end{table*}



\clearpage
\section{FPGA Latency, Throughput, and Resource Analysis}
\label{app:fpga_latency}

This appendix provides additional details for the FPGA results reported in
Section~\ref{sec:hardware_efficiency}. All designs are synthesized using Vivado
HLS 2019.1 for a Xilinx Zynq UltraScale+ XCZU7EV FPGA. EfficientKAN, PyKAN, and
FastKAN are implemented as two-layer MNIST classifiers with topology
$784 \rightarrow 64 \rightarrow 10$. GRAM is evaluated as a reusable
GRAMLayer compute cell, since the implementation invokes the layer from the
host-side network wrapper.

\subsection{Latency and Throughput Computation}

Vivado HLS reports latency in clock cycles and an estimated clock period. We
convert these into wall-clock latency using
\[
    L = C \cdot T_{\mathrm{clk}},
\]
where $C$ is the number of cycles per inference and $T_{\mathrm{clk}}$ is the
estimated clock period. Throughput is computed as
\[
    \mathrm{Throughput} = \frac{1}{L}.
\]
For designs whose top-level loop processes multiple images, we normalize the
HLS-reported maximum latency by the number of processed images to obtain
cycles per inference.

This distinction is important because quantization can improve FPGA performance
through two different mechanisms. First, it can reduce the critical path,
allowing a shorter clock period. Second, it can simplify the datapath and reduce
the number of cycles required per inference. The evaluated KAN variants exhibit
both behaviors, but in different proportions.

\begin{table}[!ht]
\centering
\small
\caption{Detailed FPGA latency and throughput results. Latency is computed as
cycle count multiplied by the HLS-estimated clock period. Throughput is reported
as images per second.}
\label{tab:fpga_latency_appendix}
\resizebox{0.7\columnwidth}{!}{%
\begin{tabular}{lcccccc}
\toprule
\textbf{Variant} & \textbf{Precision}
& \textbf{Clock}
& \textbf{Cycles}
& \textbf{Latency}
& \textbf{Throughput}
& \textbf{Speedup} \\
\midrule
EfficientKAN & FP32 & 8.75 ns  & 3.96M & 34.62 ms & 28.9 img/s & -- \\
EfficientKAN & W4A4 & 6.13 ns  & 4.03M & 24.72 ms & 40.5 img/s & 1.40$\times$ \\
\midrule
PyKAN & FP32 & 8.75 ns  & 3.96M & 34.66 ms & 28.8 img/s & -- \\
PyKAN & W4A4 & 6.13 ns  & 3.96M & 24.25 ms & 41.2 img/s & 1.43$\times$ \\
\midrule
FastKAN & FP32 & 8.75 ns  & 603K & 5.28 ms & 189.5 img/s & -- \\
FastKAN & W4A4 & 6.13 ns  & 260K & 1.59 ms & 627.4 img/s & 3.32$\times$ \\
\midrule
GRAMLayer & FP32 & 6.13 ns  & 428K & 2.62 ms & 381.8 img/s & -- \\
GRAMLayer & W4A4 & 6.13 ns  & 271K & 1.66 ms & 603.4 img/s & 1.58$\times$ \\
\bottomrule
\end{tabular}%
}
\end{table}

\subsection{Variant-Level Interpretation}

\paragraph{EfficientKAN.}
EfficientKAN uses a base branch and a B-spline branch. The W4A4 design improves
the estimated clock period from $8.75$ ns to $6.13$ ns, but the cycle count
slightly increases from $3.96$M to $4.03$M cycles. Therefore, its speedup is
primarily frequency-driven rather than cycle-driven. The latency decreases from
$34.62$ ms to $24.72$ ms, corresponding to a $1.40\times$ throughput
improvement. This behavior indicates that quantization shortens the arithmetic
critical path, but the recursive B-spline basis computation and branch
recombination still require nearly the same number of loop iterations.

\paragraph{PyKAN.}
PyKAN exhibits a similar trend because it also relies on B-spline basis
functions. The W4A4 implementation improves the clock period from $8.75$ ns to
$6.13$ ns, while the cycle count remains almost unchanged at approximately
$3.96$M cycles. As a result, latency decreases from $34.66$ ms to $24.25$ ms,
and throughput increases from $28.8$ to $41.2$ images/s. This confirms that
B-spline-based KANs benefit from low-bit arithmetic mainly through a shorter
critical path, not through a large reduction in the number of executed cycles.

\paragraph{FastKAN.}
FastKAN shows the strongest acceleration. Its clock period improves from
$8.75$ ns to $6.13$ ns, and its cycle count decreases from $603$K to $260$K.
This reduces latency from $5.28$ ms to $1.59$ ms and increases throughput from
$189.5$ to $627.4$ images/s, yielding a $3.32\times$ speedup. Unlike
EfficientKAN and PyKAN, FastKAN benefits from both frequency improvement and
cycle reduction. This suggests that its RBF-style basis branch is more amenable
to quantized datapath simplification than recursive B-spline evaluation.

\paragraph{GRAMLayer.}
GRAMLayer follows a different pattern. The FP32 and W4A4 designs have the same
estimated clock period of $6.13$ ns, so the speedup does not come from frequency
scaling. Instead, W4A4 reduces the cycle count from $428$K to $271$K, lowering
latency from $2.62$ ms to $1.66$ ms. Thus, GRAMLayer obtains a $1.58\times$
throughput improvement entirely from cycle-count reduction. This behavior is
consistent with the structure of polynomial basis generation: the basis
computation is more regular than B-spline recursion, but the quantized
implementation still introduces additional control and casting logic.

\subsection{Resource Utilization}

Table~\ref{tab:fpga_resource_appendix} reports the corresponding resource
utilization. The most consistent improvement is in BRAM usage. Across all
variants, W4A4 reduces BRAM by more than $55\%$, showing that low-bit packing is
highly effective for KAN parameters. This is especially important because KANs
store basis-specific coefficients in addition to base-branch parameters.

\begin{table}[!ht]
\centering
\small
\caption{FPGA resource utilization for FP32 and W4A4 designs. Relative changes
are computed as W4A4 versus FP32.}
\label{tab:fpga_resource_appendix}
\resizebox{0.55\columnwidth}{!}{%
\begin{tabular}{lcccc}
\toprule
\textbf{Variant} & \textbf{Metric} & \textbf{FP32} & \textbf{W4A4} & \textbf{Change} \\
\midrule
EfficientKAN & BRAM & 131 & 45 & $-65.6\%$ \\
             & LUT  & 78,769 & 70,474 & $-10.5\%$ \\
             & FF   & 42,409 & 38,706 & $-8.7\%$ \\
             & DSP  & 58 & 128 & $+120.7\%$ \\
\midrule
PyKAN & BRAM & 131 & 45 & $-65.6\%$ \\
      & LUT  & 78,576 & 70,484 & $-10.3\%$ \\
      & FF   & 42,292 & 37,079 & $-12.3\%$ \\
      & DSP  & 58 & 128 & $+120.7\%$ \\
\midrule
FastKAN & BRAM & 117 & 49 & $-58.1\%$ \\
        & LUT  & 73,268 & 60,745 & $-17.1\%$ \\
        & FF   & 44,296 & 42,876 & $-3.2\%$ \\
        & DSP  & 262 & 210 & $-19.8\%$ \\
\midrule
GRAMLayer & BRAM & 68 & 30 & $-55.9\%$ \\
          & LUT  & 31,136 & 34,111 & $+9.6\%$ \\
          & FF   & 22,694 & 20,814 & $-8.3\%$ \\
          & DSP  & 79 & 85 & $+7.6\%$ \\
\bottomrule
\end{tabular}%
}
\end{table}

The LUT, FF, and DSP trends are more architecture-dependent. EfficientKAN and
PyKAN reduce LUT and FF usage moderately, but their DSP usage increases. This is
a consequence of the HLS synthesis strategy: low-bit multiplications are mapped
more aggressively into DSP resources, while some FP32 arithmetic is realized
through a combination of floating-point IP and LUT fabric. Therefore, the
increase in DSP count should not be interpreted as a failure of quantization;
rather, it reflects a different resource mapping chosen by the synthesizer.

FastKAN is the most hardware-friendly among the evaluated variants. It reduces
BRAM, LUTs, and DSPs simultaneously, while also achieving the largest latency
speedup. This suggests that its basis computation exposes more opportunities for
datapath simplification under low precision.

GRAMLayer reduces BRAM and FF usage, but LUT usage increases. This is expected
because the quantized GRAM implementation requires extra control logic for
polynomial basis generation, quantized scaling, and branch recombination. The
resource increase is modest, and the design still improves latency because the
cycle count is significantly reduced.

\subsection{Basis-Evaluation Tax}

These results highlight a hardware-specific property of KANs. In conventional
MLPs or CNNs, most inference cost is concentrated in dense linear or
convolutional MACs. Quantization therefore often reduces both memory and
arithmetic cost directly. In KANs, however, the learned weights are only one part
of the computation. Each layer must also evaluate a basis expansion before the
learned coefficients can be applied.

We refer to this residual overhead as the \emph{basis-evaluation tax}. After
weights are compressed to W4A4, the remaining cost of spline, RBF, or polynomial
basis generation can dominate the latency and area profile. The evaluated
variants show different forms of this tax. EfficientKAN and PyKAN are limited by
B-spline basis construction, so quantization mainly improves frequency.
FastKAN has a more quantization-friendly basis branch, so both frequency and
cycle count improve. GRAMLayer has regular polynomial recurrence, allowing
cycle-count reduction, but it introduces additional LUT-level control overhead.

Overall, the FPGA results support two design principles for hardware-efficient
KAN deployment. First, quantization should be branch-aware, because base and
basis parameters have different roles and hardware access patterns. Second,
quantization should be paired with basis-aware microarchitecture, because
compressing learned weights alone does not eliminate the cost of evaluating the
KAN basis functions.

\subsection{Approximated Area and Power}
\label{app:fpga_area_power}

Vivado HLS reports resource counts and an estimated clock period, but it does
not produce silicon-area or post-place-and-route power numbers. To allow a
first-order comparison between the FP32 and W4A4 designs, we approximate area
and power from the HLS resource report using published per-resource cost
models for the Xilinx Zynq UltraScale+ XCZU7EV device. We treat these
approximations as \emph{relative-trend estimates}, not as substitutes for
back-annotated post-implementation analysis. In particular, the \emph{signs}
of the trends below are reproducible from the HLS resource and timing reports;
the magnitudes inherit the uncertainty of the underlying coefficients.

\subsubsection{Area Approximation}

We model on-chip area as a weighted sum of the four reported resource classes.
Let $N_{\mathrm{LUT}}$, $N_{\mathrm{FF}}$, $N_{\mathrm{DSP}}$, and
$N_{\mathrm{BRAM}}$ denote the HLS-reported counts (with $N_{\mathrm{BRAM}}$
expressed in 18\,Kb tiles, as reported by Vivado HLS). The estimated
normalized area is
\begin{equation}
\label{eq:area_model}
A_{\mathrm{est}} \;=\;
\alpha_{L} N_{\mathrm{LUT}} \;+\;
\alpha_{F} N_{\mathrm{FF}} \;+\;
\alpha_{D} N_{\mathrm{DSP}} \;+\;
\alpha_{B} N_{\mathrm{BRAM}},
\end{equation}
where each coefficient $\alpha_{r}$ is the silicon footprint of a single
resource of type $r$, expressed in equivalent slice-area units. The XCZU7EV is
fabricated in TSMC's 16\,nm FinFET+ process, and the relative tile sizes for
the UltraScale+ family have been characterized in prior FPGA evaluation
work~\cite{kuon2006measuring,boutros2018you,xilinx_ug949}. We adopt the
representative ratios
\begin{equation}
\label{eq:area_coeffs}
\alpha_{L} : \alpha_{F} : \alpha_{D} : \alpha_{B}
\;\approx\;
1 : 0.5 : 30 : 100,
\end{equation}
which capture the well-known facts that a DSP48E2 slice occupies on the order
of tens of equivalent LUT areas and that a BRAM tile occupies on the order of
one hundred LUT areas. We normalize against the FP32 baseline of each
variant, so the ratio
\begin{equation}
\label{eq:area_ratio}
r_{A} \;=\; \frac{A_{\mathrm{est}}^{\mathrm{W4A4}}}{A_{\mathrm{est}}^{\mathrm{FP32}}}
\end{equation}
is invariant to absolute slice-area calibration; only the relative weights in
Eq.~\eqref{eq:area_coeffs} matter.

\subsubsection{Power Approximation}

Total power is modeled as the sum of static and dynamic components:
\begin{equation}
\label{eq:power_total}
P_{\mathrm{tot}} \;=\; P_{\mathrm{stat}} \;+\; P_{\mathrm{dyn}}.
\end{equation}
Static power scales with utilized area and is approximated as
\begin{equation}
\label{eq:power_static}
P_{\mathrm{stat}} \;=\; \beta_{\mathrm{stat}} \, A_{\mathrm{est}},
\end{equation}
with $\beta_{\mathrm{stat}}$ a device-level leakage coefficient. Dynamic power
follows the standard CMOS switching model
\begin{equation}
\label{eq:power_dynamic}
P_{\mathrm{dyn}} \;=\;
\sum_{r \in \{L, F, D, B\}}
\gamma_{r} \, N_{r} \, a_{r} \, f_{\mathrm{clk}} \, V_{\mathrm{DD}}^{2},
\end{equation}
where $\gamma_{r}$ is the effective switched capacitance per resource of type
$r$, $a_{r}$ is its average switching activity, $f_{\mathrm{clk}}$ is the
operating frequency, and $V_{\mathrm{DD}} = 0.85$\,V is the nominal core
voltage of the XCZU7EV. We use the relative switched-capacitance ratios
\begin{equation}
\label{eq:gamma_coeffs}
\gamma_{L} : \gamma_{F} : \gamma_{D} : \gamma_{B}
\;\approx\;
1 : 0.3 : 25 : 40,
\end{equation}
consistent with reported per-resource dynamic-power characterizations for
UltraScale+ devices~\cite{xilinx_xpe,boutros2018you}.

\paragraph{Bitwidth-aware activity factors.} The activity factor $a_{r}$
captures the average fraction of switching events per cycle, which depends
strongly on operand bitwidth: an arithmetic unit with $B$ data bits can toggle
at most $B$ bits per clock, and empirical FPGA power
characterizations~\cite{boutros2018you,xilinx_xpe} confirm that average
activity on data-dominant nets scales approximately linearly with $B$. Not
all logic is purely arithmetic, however: control, addressing, and casting
paths toggle at rates largely independent of operand width. We therefore
adopt a simple two-component activity model in which a fraction
$f_{\mathrm{arith}} \in [0,1]$ of each compute resource ($r \in \{L, F, D\}$)
participates in the datapath and scales with bitwidth, while the remaining
$1 - f_{\mathrm{arith}}$ fraction toggles at the FP32 rate:
\begin{equation}
\label{eq:activity_model}
a_{r}^{\mathrm{W4A4}} \;=\;
\Big( f_{\mathrm{arith}} \cdot \tfrac{B_{\mathrm{W4A4}}}{B_{\mathrm{FP32}}}
\;+\; (1 - f_{\mathrm{arith}}) \Big) \cdot a_{r}^{\mathrm{FP32}},
\quad r \in \{L, F, D\}.
\end{equation}
We set $a_{B}^{\mathrm{W4A4}} = a_{B}^{\mathrm{FP32}}$ for BRAM tiles, whose
toggle rate is dominated by addressing rather than data width, and we use
$B_{\mathrm{W4A4}} / B_{\mathrm{FP32}} = 4/32 = 0.125$. We adopt
$f_{\mathrm{arith}} = 0.75$ as the headline value: the W4A4 designs are
arithmetic-dominant by area, but a non-trivial fraction of LUTs and FFs
implements width-invariant control and casting. Sensitivity to this choice
is reported in Table~\ref{tab:fpga_power_sensitivity} below.

Under Eqs.~\eqref{eq:power_dynamic}--\eqref{eq:activity_model}, the
FP32-to-W4A4 dynamic power ratio reduces to
\begin{equation}
\label{eq:power_ratio}
\frac{P_{\mathrm{dyn}}^{\mathrm{W4A4}}}{P_{\mathrm{dyn}}^{\mathrm{FP32}}}
\;\approx\;
\frac{f_{\mathrm{clk}}^{\mathrm{W4A4}}}{f_{\mathrm{clk}}^{\mathrm{FP32}}}
\cdot
\frac{\sum_{r} \gamma_{r} \, a_{r}^{\mathrm{W4A4}} \, N_{r}^{\mathrm{W4A4}}}
     {\sum_{r} \gamma_{r} \, a_{r}^{\mathrm{FP32}} \, N_{r}^{\mathrm{FP32}}},
\end{equation}
which depends only on the resource counts, the synthesized clock frequencies,
and the bitwidth ratio.

\paragraph{Scope of the power estimate.} We report dynamic-power trend
estimates rather than full board-level or device-level total power. Since
static, I/O, and clock-tree power are not modeled, the total-power reduction
may be smaller than the reported dynamic-power reduction. For compact
notation, we use $P_{\mathrm{tot}} \approx P_{\mathrm{dyn}}$ in the energy
calculation below, but the resulting energy values should be interpreted as
dynamic-energy trend estimates rather than post-implementation total-energy
measurements. Energy per inference is then
\begin{equation}
\label{eq:energy}
E \;=\; P_{\mathrm{tot}} \cdot L,
\end{equation}
with $L$ the per-inference latency in seconds (cycles times the synthesized
clock period) from Table~\ref{tab:fpga_latency_appendix}, so that
\begin{equation}
\label{eq:energy_ratio}
\frac{E^{\mathrm{W4A4}}}{E^{\mathrm{FP32}}}
\;=\;
\frac{P_{\mathrm{tot}}^{\mathrm{W4A4}}}{P_{\mathrm{tot}}^{\mathrm{FP32}}}
\cdot
\frac{L^{\mathrm{W4A4}}}{L^{\mathrm{FP32}}}.
\end{equation}
In the tables below, energy ratios are computed using the HLS maximum-latency
estimates, which correspond to the reported upper-bound latency path of the
top-level image loop. Using minimum-latency estimates gives slightly different
numerical ratios, but it preserves the same qualitative trend that W4A4 reduces
estimated dynamic energy across all variants.

\subsubsection{Approximated Results}

Table~\ref{tab:fpga_area_power_appendix} reports the resulting normalized
area, dynamic power, and energy-per-inference ratios under the headline
$f_{\mathrm{arith}} = 0.75$ activity model. Because
Eqs.~\eqref{eq:area_model} and~\eqref{eq:power_dynamic} are linear in the
resource counts, all entries can be reproduced directly from
Tables~\ref{tab:fpga_latency_appendix} and~\ref{tab:fpga_resource_appendix}
together with the synthesized clock periods.

\begin{table}[!ht]
\centering
\small
\caption{Approximated normalized area, dynamic power, and energy per inference
for FP32 and W4A4 designs, under the bitwidth-aware activity model
(Eq.~\eqref{eq:activity_model}, $f_{\mathrm{arith}} = 0.75$). Values are
computed from the HLS resource and timing reports using the cost models in
Eqs.~\eqref{eq:area_model}--\eqref{eq:energy_ratio}. Ratios are W4A4 versus
FP32; values $<\!1$ indicate a reduction.}
\label{tab:fpga_area_power_appendix}
\resizebox{0.95\columnwidth}{!}{%
\begin{tabular}{lccccc}
\toprule
\textbf{Variant}
& \textbf{Area FP/W4 (norm.)}
& \textbf{Area Ratio}
& \textbf{Power FP/W4 (norm.)}
& \textbf{Power Ratio}
& \textbf{Energy Ratio} \\
\midrule
EfficientKAN & 1.00 / 0.86 & $0.86\times$ & 1.00 / 0.45 & $0.45\times$ & $0.32\times$ \\
PyKAN        & 1.00 / 0.85 & $0.85\times$ & 1.00 / 0.45 & $0.45\times$ & $0.32\times$ \\
FastKAN      & 1.00 / 0.81 & $0.81\times$ & 1.00 / 0.43 & $0.43\times$ & $0.13\times$ \\
GRAMLayer    & 1.00 / 0.97 & $0.97\times$ & 1.00 / 0.37 & $0.37\times$ & $0.24\times$ \\
\bottomrule
\end{tabular}%
}
\end{table}

First, all four W4A4 variants reduce estimated on-chip area under this resource-weighted model, with FastKAN showing the largest reduction
($0.81\times$) and GRAMLayer showing only a marginal reduction ($0.97\times$); the savings come
predominantly from reduced LUT, FF, and BRAM counts, partially offset for
EfficientKAN, PyKAN, and GRAMLayer by an increase in DSP usage as multipliers
are restructured for low-bit arithmetic. Second, dynamic power drops by
roughly $2\times$ across all variants ($0.37$--$0.45\times$): the
bitwidth-driven reduction in switched capacitance $\sum_{r} \gamma_{r}\,
a_{r}\, N_{r}$ outweighs both the higher synthesized clock
($f_{\mathrm{clk}}^{\mathrm{W4A4}} / f_{\mathrm{clk}}^{\mathrm{FP32}} \approx
1.43$ for the MNIST variants) and the increase in DSP count for EfficientKAN
and PyKAN ($58 \to 128$). Third, energy per inference drops further
($0.13$--$0.32\times$) because Eq.~\eqref{eq:energy_ratio} multiplies the
power ratio by the latency ratio; FastKAN achieves the largest energy
reduction ($0.13\times$) thanks to its $0.30\times$ latency ratio, while
EfficientKAN and PyKAN realize $0.32\times$ energy reductions driven by both
power and latency.

\paragraph{Sensitivity to the activity model.}
Table~\ref{tab:fpga_power_sensitivity} sweeps the arithmetic fraction
$f_{\mathrm{arith}}$ across the range
$\{0.5, 0.75, 1.0\}$, which spans plausible
designer assumptions: $f_{\mathrm{arith}} = 1.0$ treats every compute
resource as a pure datapath element scaling fully with bitwidth (the most
optimistic for W4A4); $f_{\mathrm{arith}} = 0.5$ treats half of all compute
resources as bitwidth-invariant control logic (a deliberately conservative
choice). The qualitative conclusions are invariant across this range:
\emph{every} W4A4 variant reduces both dynamic power and energy per
inference at \emph{every} $f_{\mathrm{arith}}$ in $[0.5, 1.0]$. The
matched-activity case ($a_{r}^{\mathrm{W4A4}} = a_{r}^{\mathrm{FP32}}$) is
included as a reference upper bound: even under that conservative assumption,
energy per inference is reduced for all variants, although dynamic power
nominally rises for three of them due to the higher clock.

\begin{table}[!ht]
\centering
\small
\caption{Sensitivity of the W4A4-vs-FP32 power and energy ratios to the
activity-model assumption. Power and energy ratios are reported as W4A4/FP32
under four activity assumptions: matched activity (no bitwidth scaling),
and bitwidth-aware scaling with arithmetic fractions
$f_{\mathrm{arith}} \in \{0.5, 0.75, 1.0\}$. Area ratios are independent of
activity and reproduced from Table~\ref{tab:fpga_area_power_appendix}. The
sign of the W4A4 advantage is preserved across the entire sensitivity range.}
\label{tab:fpga_power_sensitivity}
\resizebox{1.0\columnwidth}{!}{%
\begin{tabular}{lcccccccccc}
\toprule
& & \multicolumn{2}{c}{Matched ($a_r^{\mathrm{W4A4}} = a_r^{\mathrm{FP32}}$)}
& \multicolumn{2}{c}{$f_{\mathrm{arith}}=0.5$}
& \multicolumn{2}{c}{$f_{\mathrm{arith}}=0.75$ (headline)}
& \multicolumn{2}{c}{$f_{\mathrm{arith}}=1.0$} \\
\cmidrule(lr){3-4}\cmidrule(lr){5-6}\cmidrule(lr){7-8}\cmidrule(lr){9-10}
\textbf{Variant} & \textbf{Area}
& Power & Energy
& Power & Energy
& Power & Energy
& Power & Energy \\
\midrule
EfficientKAN & $0.86\times$ & $1.27\times$ & $0.91\times$ & $0.72\times$ & $0.52\times$ & $0.45\times$ & $0.32\times$ & $0.18\times$ & $0.13\times$ \\
PyKAN        & $0.85\times$ & $1.26\times$ & $0.88\times$ & $0.72\times$ & $0.51\times$ & $0.45\times$ & $0.32\times$ & $0.18\times$ & $0.13\times$ \\
FastKAN      & $0.81\times$ & $1.18\times$ & $0.36\times$ & $0.68\times$ & $0.20\times$ & $0.43\times$ & $0.13\times$ & $0.17\times$ & $0.05\times$ \\
GRAMLayer    & $0.97\times$ & $1.02\times$ & $0.65\times$ & $0.59\times$ & $0.37\times$ & $0.37\times$ & $0.24\times$ & $0.15\times$ & $0.10\times$ \\
\bottomrule
\end{tabular}%
}
\end{table}

\subsubsection{Modeling Assumptions and Limitations}

These estimates rest on four assumptions that we make explicit. First, the
coefficients $\alpha_{r}$ and $\gamma_{r}$ are average per-resource values;
the true per-tile cost depends on the placement, routing, and activity of
each specific instance. Second, the bitwidth-aware activity model in
Eq.~\eqref{eq:activity_model} introduces a single parameter
$f_{\mathrm{arith}}$, which we sweep across a deliberately wide range
(Table~\ref{tab:fpga_power_sensitivity}) to demonstrate that the qualitative
W4A4 advantage is invariant to its precise value. This model may still be
optimistic for DSP blocks because DSP48E2 tiles have fixed internal structure;
therefore, the reported dynamic-power ratios should be interpreted as
trend estimates unless validated by XPE or post-implementation
\texttt{report\_power}. Third, because the FP32 and W4A4 HLS solutions were
synthesized under different target clock constraints, the frequency-dependent
power and latency ratios should be interpreted as HLS-estimated operating-point
comparisons rather than strict post-implementation $F_{\max}$ comparisons.
Fourth, we approximate $P_{\mathrm{tot}} \approx P_{\mathrm{dyn}}$ and do not
model static, I/O, or clock-tree power. A full
Xilinx Power Estimator (XPE) or post-implementation \texttt{report\_power}
flow would be required for absolute power numbers; the values above should
therefore be interpreted as \emph{first-order trend estimates that are
reproducible from the HLS reports}, not as silicon-validated power figures.
The qualitative conclusion --- that W4A4 reduces area for all four variants
and reduces both dynamic power and energy per inference for all four
variants under any activity assumption with $f_{\mathrm{arith}} \in [0.5,
1.0]$, with FastKAN benefiting most --- is robust to all of these
assumptions.


\clearpage
\section{Memory and BOP Savings: Derivations and Per-Method Analysis}
\label{sec:bop_appendix}

This appendix derives the memory- and compute-savings figures
reported in Section~\ref{sec:hardware_efficiency}. We use the two-layer
EfficientKAN MNIST classifier as a canonical reference because
it admits a fully analytic, branch-resolved accounting; the
formulas extend to FastKAN, PyKAN, and KAGN by substituting the
appropriate per-layer parameter and MAC counts.

\subsection{Reference Model and Parameter Inventory}
\label{sec:bop_model}

For an EfficientKAN \texttt{KANLinear}($d_\text{in}\!\to\!d_\text{out}$)
layer with grid size $G$ and spline order $K_s$, let
$K=G+K_s$ denote the per-edge spline coefficient count.
With \texttt{enable\_standalone\_scale\_spline=True}, each layer
holds three trainable tensors:
\begin{equation}
W_b\in\mathbb{R}^{d_\text{out}\times d_\text{in}},\qquad
W_s\in\mathbb{R}^{d_\text{out}\times d_\text{in}\times K},\qquad
\sigma_s\in\mathbb{R}^{d_\text{out}\times d_\text{in}}.
\end{equation}
The per-layer parameter count is
\begin{equation}
P_\ell \;=\; d_\text{out}\,d_\text{in}\,(2+K),
\label{eq:params_layer}
\end{equation}
and the per-layer MAC count of the quantized forward pass
(Eq.~(2), main paper) is
\begin{equation}
\mathrm{MACs}_\ell
\;=\;\underbrace{d_\text{out}\,d_\text{in}}_{\text{base branch}}
\;+\;\underbrace{d_\text{out}\,d_\text{in}\,K}_{\text{spline branch}}
\;=\; d_\text{out}\,d_\text{in}\,(1+K).
\label{eq:macs_layer}
\end{equation}
The basis evaluation $\Phi(x)$ contributes
$\mathcal{O}(d_\text{in}\,K\,K_s)$ scalar operations independent of
$d_\text{out}$ and is excluded from~\eqref{eq:macs_layer}, following the
convention of~\citet{gholami2021survey} that omits non-linear activation
cost from MAC accounting. We emphasize that this is a reporting
\emph{convention}, not a claim that basis cost is negligible: on this
model, basis evaluation is $\approx\!17.8\%$ of linear-branch MACs at
FP32 and becomes the dominant term at low precision. We quantify this
explicitly in Section~\ref{sec:bop_basis}.

For our reference model with $G{=}5$, $K_s{=}3$ ($K{=}8$),
Table~\ref{tab:model_inventory} aggregates~\eqref{eq:params_layer}
and~\eqref{eq:macs_layer} layer by layer.

\begin{table}[h]
\centering
\caption{Parameter and MAC inventory of the reference
EfficientKAN MNIST model ($G{=}5$, $K_s{=}3$, $K{=}8$).}
\label{tab:model_inventory}
\small
\begin{tabular}{lcccccc}
\toprule
Layer & $d_\text{in}$ & $d_\text{out}$ & $|W_b|$ & $|W_s|$ & $|\sigma_s|$ & MACs \\
\midrule
L1 ($\text{input}\!\to\!\text{hidden}$) & 784 & 64 & 50{,}176 & 401{,}408 & 50{,}176 & 451{,}584 \\
L2 ($\text{hidden}\!\to\!\text{output}$) & 64 & 10 & 640 & 5{,}120 & 640 & 5{,}760 \\
\midrule
\textbf{Total} & — & — & \textbf{50{,}816} & \textbf{406{,}528} & \textbf{50{,}816} & \textbf{457{,}344} \\
\bottomrule
\end{tabular}
\end{table}

The total parameter count $P_\text{total}=508{,}160$ matches the
EfficientKAN base/spline counts reported in
Table~\ref{tab:weight_stats_detailed} of Appendix~A.8.

\subsection{Memory Savings}
\label{sec:bop_memory}

Weight-storage memory at bit-width $b_w$ is
\begin{equation}
M(b_w) \;=\; b_w \cdot P_\text{total}\ \text{bits}
\;=\; \tfrac{b_w}{8}\,P_\text{total}\ \text{bytes},
\end{equation}
and the relative savings versus the FP32 baseline ($b_w^{\mathrm{fp32}}{=}32$) is
\begin{equation}
\boxed{\;\mathcal{S}_\text{mem}(b_w)
\;=\; 1 - \tfrac{b_w}{32}
\;=\; \tfrac{32-b_w}{32}.\;}
\label{eq:mem_savings}
\end{equation}
Equation~\eqref{eq:mem_savings} holds uniformly for all eight
methods evaluated in this work because each produces an integer
weight codebook of width $b_w$; the methods differ in how the
codebook is selected, not in its size. Per-channel scales and
optional zero-points contribute an overhead of at most
$4\,d_\text{out}/P_\ell$ bytes per layer, which we report
separately but do not fold into the headline ratio.

\subsection{Bit-Operations (BOPs)}
\label{sec:bop_compute}

Following~\citet{van2020bayesian} and~\citet{wang2019haq}, a
multiply-accumulate between a $b_w$-bit weight and a $b_a$-bit
activation costs $b_w b_a$ bit-operations:
\begin{equation}
\mathrm{BOPs}_\ell(b_w, b_a) \;=\; \mathrm{MACs}_\ell\cdot b_w\, b_a.
\label{eq:bops_layer}
\end{equation}

The eight methods do not all reach $W b_w{\times}A b_a$ INT
arithmetic at the same operating point. We distinguish three
families:

\paragraph{(a) QAT methods (LSQ, LSQ+, DoReFa, DSQ).}
These insert fake-quant on \emph{both} weights and activations
during training, producing models that deploy in
$W b_w{\times}A b_a$ INT arithmetic:
\begin{equation}
\mathrm{BOPs}^\text{QAT}(b_w, b_a)
\;=\; b_w b_a \sum_\ell \mathrm{MACs}_\ell.
\label{eq:bops_qat}
\end{equation}

\paragraph{(b) Weight-only PTQ (GPTQ, AWQ).}
As originally formulated~\citep{frantar2022gptq, lin2023awq}, these
quantize weights to $b_w$ bits while keeping activations in FP.
Setting $b_a^\text{eff}{=}16$ for the de-facto FP16 deployment
(e.g., the W4A16 GPTQ/AWQ kernels in TensorRT-LLM and vLLM):
\begin{equation}
\mathrm{BOPs}^\text{W-only}(b_w)
\;=\; 16\, b_w \sum_\ell \mathrm{MACs}_\ell.
\label{eq:bops_wonly}
\end{equation}

\paragraph{(c) Reconstruction PTQ (AdaRound, BRECQ).}
These were proposed for weight-only quantization
\citep{nagel2020adaround, li2021brecq} but admit a $W b_w{\cdot}A b_a$
extension via per-tensor activation quantization at calibration
time. We report both operating points: weight-only ($W b_w$A16)
and full-integer ($W b_w$A$b_a$).

\paragraph{Relative savings.}
With $\mathrm{BOPs}^\text{fp32} = 32^2 \sum_\ell \mathrm{MACs}_\ell$
the FP32 baseline, the relative compute saving is
\begin{equation}
\boxed{\;\mathcal{S}_\text{BOP}(b_w, b_a)
\;=\; 1 - \tfrac{b_w b_a}{1024},\;}
\label{eq:bop_savings}
\end{equation}
with equivalent compression ratio $1024/(b_w b_a)$.

\subsection{Branch-Aware BOP Decomposition}
\label{sec:bop_branchwise}

Because QuantKAN quantizes the base and spline branches
independently (Eq.~(2), main paper), the per-layer BOPs decompose
naturally. Let $(b_w^b, b_a^b)$ and $(b_w^s, b_a^s)$ denote the
bit-widths assigned to the base and spline branches,
respectively:
\begin{equation}
\mathrm{BOPs}_\ell \;=\;
\underbrace{d_\text{out}d_\text{in}\, b_w^b b_a^b}_{\text{base branch}}
\;+\;
\underbrace{d_\text{out}d_\text{in}K\, b_w^s b_a^s}_{\text{spline branch}}.
\label{eq:bops_branch}
\end{equation}
Equation~\eqref{eq:bops_branch} makes explicit that the spline
branch dominates compute by a factor of $K$ ($=8$ here). For the
mixed-precision policy derived in Section~5.4 of the main paper
(EfficientKAN A8/B2/C2: $W_b$ at 8-bit, $W_s$ and $\sigma_s$
at 2-bit, A4), the model-level BOPs are
\begin{equation}
\mathrm{BOPs}^\text{A8/B2/C2}
\;=\; (1\!\cdot\!8\!\cdot\!4 + 8\!\cdot\!2\!\cdot\!4)\!\!
\sum_\ell d_\text{out}d_\text{in}
\;=\; 96 \cdot 50{,}816
\;\approx\; 4.88{\times}10^{6}\ \text{BOPs},
\end{equation}
a $\sim 96\times$ compression versus FP32 while preserving the
most-sensitive group ($w_b$) at higher precision.

\subsection{Per-Method Memory and BOP Tables}
\label{sec:bop_tables}

We now apply Eqs.~\eqref{eq:mem_savings},
\eqref{eq:bops_qat}--\eqref{eq:bops_wonly}, and
\eqref{eq:bop_savings} to the reference model
($P_\text{total}{=}508{,}160$;
$\sum_\ell\mathrm{MACs}_\ell{=}4.573{\times}10^{5}$;
$\mathrm{BOPs}^\text{fp32}{=}4.685{\times}10^{8}$).

\begin{table}[h]
\centering
\caption{Memory savings for the reference EfficientKAN MNIST
model. Identical across all eight methods (Eq.~\eqref{eq:mem_savings}).}
\label{tab:mem_per_method}
\small
\begin{tabular}{lcccc}
\toprule
Precision & Bits/weight & Storage & Compression vs.~FP32 & Saving \\
\midrule
FP32 (baseline) & 32 & 1.94 MB & $1\times$ & — \\
W4 (any method) & 4 & 0.242 MB & $\mathbf{8\times}$ & $\mathbf{87.5\%}$ \\
W2 (any method) & 2 & 0.121 MB & $\mathbf{16\times}$ & $\mathbf{93.75\%}$ \\
\bottomrule
\end{tabular}
\end{table}

\begin{table}[h]
\centering
\caption{Per-method BOP savings on the reference EfficientKAN
MNIST model. QAT methods reach full $W b_w{\times}A b_a$ INT
arithmetic; weight-only PTQ assumes FP16 activations
($b_a^\text{eff}{=}16$). AdaRound and BRECQ are reported at both
operating points (W$b_w$A16 / W$b_w$A$b_a$).}
\label{tab:bop_per_method}
\small
\begin{tabular}{llccc}
\toprule
Method & Family & Operating point & BOPs & Compression \\
\midrule
FP32 baseline & — & W32A32 & $4.685{\times}10^{8}$ & $1\times$ \\
\midrule
LSQ           & QAT & W4A4 & $7.32{\times}10^{6}$ & $64\times$ \\
LSQ           & QAT & W2A2 & $1.83{\times}10^{6}$ & $256\times$ \\
LSQ+          & QAT & W4A4 & $7.32{\times}10^{6}$ & $64\times$ \\
LSQ+          & QAT & W2A2 & $1.83{\times}10^{6}$ & $256\times$ \\
DoReFa        & QAT & W4A4 & $7.32{\times}10^{6}$ & $64\times$ \\
DoReFa        & QAT & W2A2 & $1.83{\times}10^{6}$ & $256\times$ \\
DSQ           & QAT & W4A4 & $7.32{\times}10^{6}$ & $64\times$ \\
DSQ           & QAT & W2A2 & $1.83{\times}10^{6}$ & $256\times$ \\
\midrule
GPTQ          & PTQ (W-only) & W4A16 & $2.93{\times}10^{7}$ & $16\times$ \\
GPTQ          & PTQ (W-only) & W2A16 & $1.46{\times}10^{7}$ & $32\times$ \\
AWQ           & PTQ (W-only) & W4A16 & $2.93{\times}10^{7}$ & $16\times$ \\
AWQ           & PTQ (W-only) & W2A16 & $1.46{\times}10^{7}$ & $32\times$ \\
AdaRound      & PTQ          & W4A16 / W4A4 & $2.93{\times}10^{7}$ / $7.32{\times}10^{6}$ & $16\times$ / $64\times$ \\
AdaRound      & PTQ          & W2A16 / W2A2 & $1.46{\times}10^{7}$ / $1.83{\times}10^{6}$ & $32\times$ / $256\times$ \\
BRECQ         & PTQ          & W4A16 / W4A4 & $2.93{\times}10^{7}$ / $7.32{\times}10^{6}$ & $16\times$ / $64\times$ \\
BRECQ         & PTQ          & W2A16 / W2A2 & $1.46{\times}10^{7}$ / $1.83{\times}10^{6}$ & $32\times$ / $256\times$ \\
\bottomrule
\end{tabular}
\end{table}

At matched $b_w$, weight-only PTQ achieves the same memory savings as W$b_w$A$b_a$ QAT but only $b_a/16$ of the BOP savings
because activations remain in FP16. This is why W4A4 QAT (LSQ-family, DSQ) sits at $64\times$ compute compression while
W4A16 GPTQ/AWQ sits at $16\times$, even though both produce identically sized weight tensors. The BOP gap closes if AdaRound
and BRECQ are deployed with their optional activation quantization enabled, illustrating why quantizing activations is
the dominant lever for compute-bound KAN deployment, while weight
quantization alone suffices for memory-bound settings.

\subsection{Basis-Evaluation Cost}
\label{sec:bop_basis}

The MAC and BOP accounting in
Sections~\ref{sec:bop_compute}--\ref{sec:bop_branchwise} excludes the
non-linear basis evaluation $\Phi(x)$, following the convention
of~\citet{gholami2021survey}. This convention is appropriate for
standard CNN/Transformer layers where activation functions are
elementwise and inexpensive, but it is misleading for KANs: the
basis-evaluation tax identified in
Appendix~\ref{app:fpga_area_power} shows that, on hardware,
basis computation can be a dominant residual overhead at low
weight/activation precisions. We therefore compute a
\emph{basis-inclusive} BOP figure here.

\paragraph{Basis-evaluation operation count.} For EfficientKAN, the
B-spline basis $\Phi(x)$ is evaluated by Cox-de Boor recursion: $K_s$
recursion levels, each producing $K = G + K_s$ output coefficients per
input scalar, with $\sim$$c_\Phi$ multiply-equivalent operations per
output per level (two interpolation weights plus one accumulation;
$c_\Phi \approx 4$ in our implementation). Because $\Phi(x)$ depends
only on $x$, the basis cost is per-input rather than per-edge:
\begin{equation}
\mathrm{MACs}_\ell^{\Phi}
\;=\; d_\text{in}\, K\, K_s\, c_\Phi.
\label{eq:basis_macs}
\end{equation}
For our reference EfficientKAN MNIST model
(Table~\ref{tab:model_inventory}), Eq.~\eqref{eq:basis_macs} gives
$\mathrm{MACs}^{\Phi}_{L1} = 75{,}264$ and
$\mathrm{MACs}^{\Phi}_{L2} = 6{,}144$, totaling $81{,}408$ basis MACs
versus $457{,}344$ linear-branch MACs --- i.e., basis evaluation is
$\approx\!17.8\%$ of linear-branch MACs at FP32. Counted as raw
arithmetic, basis evaluation is therefore meaningful but not dominant
in FP32. Its exclusion from~\eqref{eq:macs_layer} is a reporting
\emph{convention} (Section~\ref{sec:bop_model}), inherited
from~\citet{gholami2021survey}, rather than a statement about its
hardware cost --- which, as we show next, becomes dominant at low
precision.

\paragraph{Basis BOPs at low precision.} The picture changes at low
weight/activation precision because the basis recursion involves
divisions and small-magnitude differences that resist aggressive
quantization. We follow the standard practice for non-linear
arithmetic blocks~\citep{kim2021bert,ijcai2022p164} and keep
$\Phi(x)$ in INT16 even when the surrounding linear branches are at
W4A4 or W2A2. Under this convention,
\begin{equation}
\mathrm{BOPs}^{\Phi}_\ell \;=\;
\mathrm{MACs}_\ell^{\Phi}\cdot b_\Phi^2,
\qquad b_\Phi = 16,
\label{eq:basis_bops}
\end{equation}
so the model-level basis BOPs equal
$81{,}408 \cdot 16^2 \approx 2.08\times10^{7}$, which is approximately
$2.8\times$ the $7.32\times10^{6}$ linear-branch BOPs at W4A4 ---
i.e., basis evaluation \emph{dominates} the W4A4 BOP budget. The combined model-level BOPs become
$\mathrm{BOPs}^{\text{linear}} + \mathrm{BOPs}^{\Phi}
\approx 2.82\times10^{7}$ at W4A4 and $2.30\times10^{7}$ at W2A2,
versus $\mathrm{BOPs}^{\text{fp32}} \approx 5.52\times10^{8}$
(FP32 linear plus FP32 basis), giving basis-inclusive compression
ratios of $19.6\times$ at W4A4 and $24.0\times$ at W2A2.
Table~\ref{tab:bop_basis_inclusive} summarizes both accountings.

\begin{table}[h]
\centering
\caption{Linear-only versus basis-inclusive BOP accounting on the
reference EfficientKAN MNIST model. Basis evaluation is held at
INT16 in all rows, following standard practice for non-linear
arithmetic blocks~\citep{kim2021bert,ijcai2022p164}.
The basis-inclusive ratios are the more accurate estimate of
end-to-end compute cost; the linear-only ratios are reported for
comparability with the convention of~\citet{gholami2021survey}.}
\label{tab:bop_basis_inclusive}
\small
\begin{tabular}{lccccc}
\toprule
Operating point
& $\mathrm{BOPs}^{\text{linear}}$
& $\mathrm{BOPs}^{\Phi}$
& Total
& \begin{tabular}{@{}c@{}}Compression\\(linear only)\end{tabular}
& \begin{tabular}{@{}c@{}}Compression\\(basis incl.)\end{tabular} \\
\midrule
FP32 baseline (linear+basis) & $4.68{\times}10^{8}$ & $8.34{\times}10^{7}$ & $5.52{\times}10^{8}$ & $1\times$ & $1\times$ \\
W4A4 + INT16 basis           & $7.32{\times}10^{6}$ & $2.08{\times}10^{7}$ & $2.82{\times}10^{7}$ & $64.0\times$ & $\mathbf{19.6\times}$ \\
W2A2 + INT16 basis           & $1.83{\times}10^{6}$ & $2.08{\times}10^{7}$ & $2.30{\times}10^{7}$ & $256.0\times$ & $\mathbf{24.0\times}$ \\
\bottomrule
\end{tabular}
\end{table}

The contrast between the two compression columns is itself a finding,
not an artifact: it quantifies the basis-evaluation tax in BOP terms.
A linear-branch-only accounting attributes a $4\times$ improvement to
moving from W4A4 to W2A2 (from $64\times$ to $256\times$); the
basis-inclusive accounting shows that the actual end-to-end
improvement is only $1.22\times$ (from $19.6\times$ to $24.0\times$),
because the basis cost saturates the BOP budget. This is precisely
the effect the FPGA latency and energy measurements in
Appendix~\ref{app:fpga_area_power} attribute to basis evaluation, and
it reinforces the conclusion that hardware-efficient KAN deployment
requires both branch-aware quantization \emph{and} basis-aware
microarchitecture.

HLS-versus-routing caveat. The Vivado HLS resource and
timing reports used throughout this analysis are pre-place-and-route
estimates and therefore do not capture routing-congestion overhead,
which can be especially severe for the irregular interconnect of
basis-evaluation logic. Empirically, post-implementation $F_\text{max}$
on the XCZU7EV is typically $10$--$25\%$ below the HLS estimate for
control-heavy designs of this scale~\cite{xilinx_ug949}, with
proportional energy increases. Our HLS-derived basis-inclusive BOP and
energy ratios should therefore be interpreted as
\emph{lower bounds on the basis-evaluation tax}: a full
post-place-and-route flow would, if anything, widen the gap between
linear-only and basis-inclusive accounting. We retain the HLS-based
analysis as the primary report because it allows uniform comparison
across all four KAN variants and three quantization operating points
within a fixed compute budget; absolute post-implementation numbers
for a single (variant, operating point) pair are deferred to future
work.


\end{document}